\documentclass{article}
\PassOptionsToPackage{numbers}{natbib}
\usepackage[preprint]{neurips_2025}
\usepackage[utf8]{inputenc}
\usepackage[T1]{fontenc}    
\usepackage{hyperref}      
\usepackage{url}         
\usepackage{booktabs}      
\usepackage{amsfonts}    
\usepackage{nicefrac}       
\usepackage{microtype}       

\usepackage{amsmath}
\usepackage{amssymb}
\usepackage{graphicx}
\usepackage{float} 
\usepackage{subfigure}
\usepackage{subcaption}  
\usepackage[export]{adjustbox}
\usepackage{xcolor}  
\usepackage{multirow}
\usepackage[toc,page]{appendix}
\usepackage{natbib} 

\title{Diffusion Sampling Path Tells More: An Efficient Plug-and-Play Strategy for Sample Filtering}

\author{Sixian Wang\\
  School of Science and Engineering\\
  The Chinese University of Hong Kong, Shenzhen\\
  \texttt{} \\
  \And
  Zhiwei Tang \\
  DAMO Academy, Alibaba Group\\
  \And
  Tsung-Hui Chang \\
  School of Artificial Intelligence \\
  The Chinese University of Hong Kong, Shenzhen \\
 \\ 
  {\tt\normalsize { \href{https://github.com/WSX20003/CFG-Rejection.git}{Official Implementation}}}
}

\begin{document}
\maketitle
\vspace{-0.3cm}
\begin{abstract}
\vspace{-0.3cm}
Diffusion models often exhibit inconsistent sample quality due to stochastic variations inherent in their sampling trajectories. Although training-based fine-tuning (e.g. DDPO \citep{black2023training}) and inference-time alignment techniques\citep{tang2024inference} aim to improve sample fidelity, they typically necessitate full denoising processes and external reward signals. This incurs substantial computational costs, hindering their broader applicability. In this work, we unveil an intriguing phenomenon: a previously unobserved yet exploitable link between sample quality and characteristics of the denoising trajectory during classifier-free guidance (CFG). Specifically, we identify a strong correlation between high-density regions of the sample distribution and the Accumulated Score Differences (ASD)—the cumulative divergence between conditional and unconditional scores. Leveraging this insight, we introduce CFG-Rejection, an efficient, plug-and-play strategy that filters low-quality samples at an early stage of the denoising process, crucially without requiring external reward signals or model retraining. Importantly, our approach necessitates no modifications to model architectures or sampling schedules and maintains full compatibility with existing diffusion frameworks. We validate the effectiveness of CFG-Rejection in image generation through extensive experiments, demonstrating marked improvements on human preference scores (HPSv2, PickScore) and challenging benchmarks (GenEval, DPG-Bench). We anticipate that CFG-Rejection will offer significant advantages for diverse generative modalities beyond images, paving the way for more efficient and reliable high-quality sample generation.
\end{abstract}
\vspace{-0.3cm}
\section{Introduction}
\vspace{-0.1cm}
Denoising diffusion models~\cite{ho2020ndm,song2021ddim,song2019generative} have emerged as a powerful generative framework across modalities including image~\cite{sdxl,peebles2023scalable,dhariwal2021diffusion}, video~\cite{wang2025wan}, and 3D generation~\cite{hollein2024viewdiff,voleti2024sv3d}, leveraging iterative noise refinement through the reverse diffusion process~\cite{song2021sde} to model complex data distributions.

Despite these remarkable performances, a key challenge remains: the stochastic nature of the sampling process often results in inconsistent output quality. For users of diffusion-based image generators, the experience resembles a game of chance: submit a prompt and hope that the generated image aligns with their envisioned result. In practice, this often requires multiple regeneration attempts with different random seeds, leading to a trial-and-error workflow that is both time-consuming and computationally inefficient.

\vspace{-0.3cm}
\begin{figure}[H]
  \centering
  \includegraphics[width=14cm]{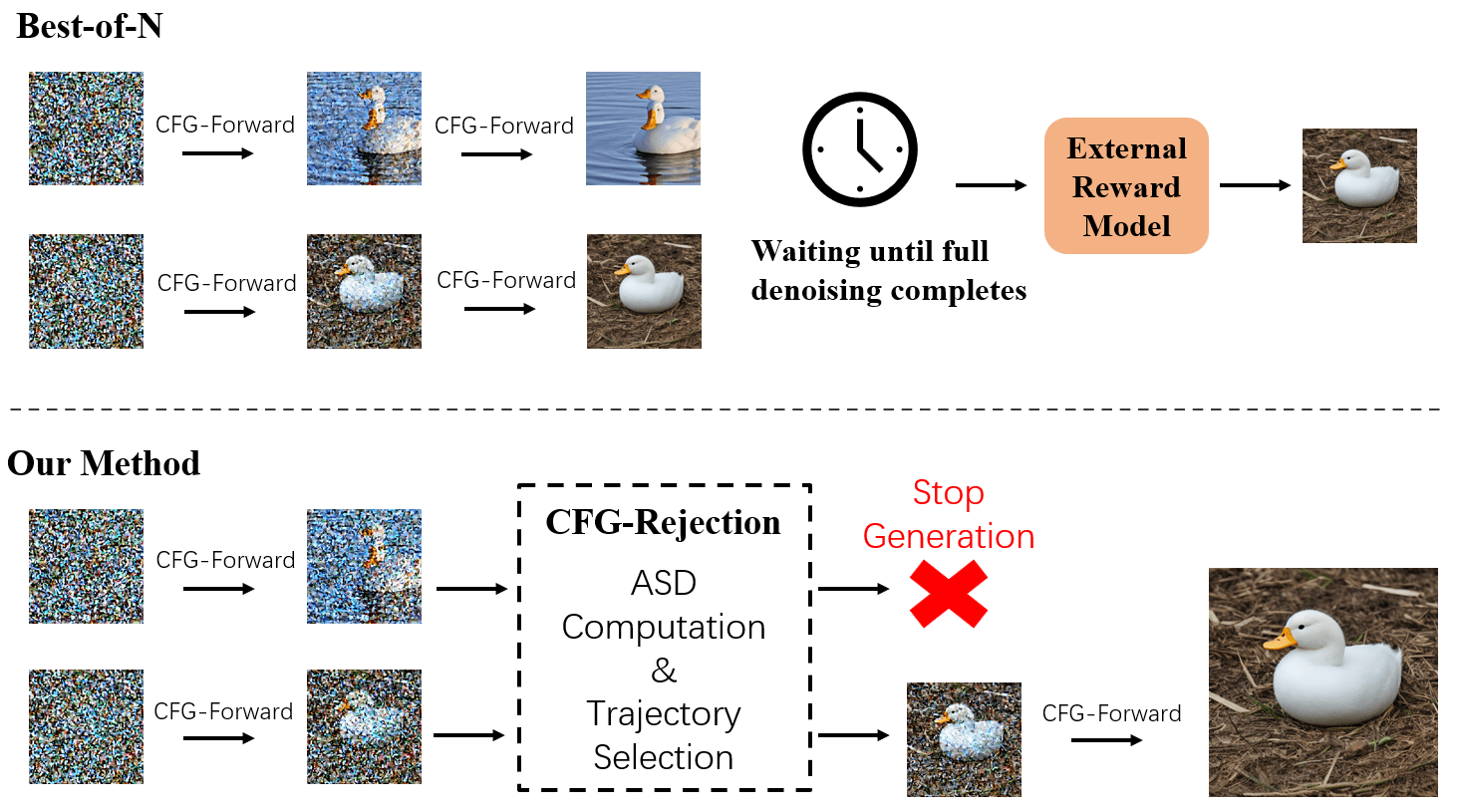}
  \caption{Illustration of filtering framework. Best-of-N completes all denoising steps, using an external reward model to select the high-quality image, while our method halts low-quality generations early with the intrinsic information in the sampling path.}
  \label{method} 
\end{figure}
\vspace{-0.3cm}

Several strategies have emerged to mitigate this output variability. One prominent avenue involves training-based methodologies, which fine-tune pre-trained diffusion models using task-specific or implicit rewards~\cite{simsar2024loraclr,wallace2024diffusion,dong2023raft,fan2023dpok,black2023training}. For instance, \cite{black2023training} demonstrated the integration of reinforcement learning objectives to guide generation towards desired properties. However, these approaches typically incur substantial computational overhead associated with reward model construction, data relabeling, and extensive fine-tuning.

Alternatively, inference-time alignment strategies focus on optimizing the sampling process itself, often by manipulating initial noise vectors, to produce higher-fidelity samples~\cite{tang2024inference,wallace2023end,ma2025inference,ben2024d}. For example, techniques like Best-of-N sampling~\cite{ma2025inference} leverage external verifiers to identify optimal noise trajectories. Others employ first-order optimization along the sampling path to steer the trajectory toward higher quality outputs~\cite{tang2024inference}. However, these methods grapple with two critical limitations that hinder broader adoption:

\begin{itemize}
\item \textbf{Dependence on External Reward Signals.} Existing approaches rely on external reward models trained on limited data and older architectures (e.g., PickScore~\cite{kirstain2023pick} with SD1.5 and SDXL images), which may not generalize well to new models or nuanced downstream tasks such as text rendering.
\item \textbf{Inefficient Pixel-Space Evaluation.} These reward models typically operate post hoc, evaluating fully generated images in pixel space. As a result, they require complete denoising before any quality signal is available, leading to significant inference overhead.
\end{itemize}

Our work centers on advancing inference-time alignment, aiming to efficiently identify superior samples from the diffusion model's distribution. To overcome these challenges, we introduce \textbf{CFG-Rejection}, an efficient, plug-and-play strategy that \textbf{filters low-quality samples at an early stage of the denoising process}, crucially \textbf{without requiring external reward signals or model retraining}. As an inference-time alignment technique, CFG-Rejection requires no modifications to model architectures or sampling schedules and maintains full compatibility with existing diffusion frameworks.

Our approach is predicated on a key insight: a previously unobserved yet exploitable link between sample quality and the characteristics of the denoising trajectory during classifier-free guidance (CFG)~\cite{ho2022classifier}. Specifically, we identify that the \textbf{Accumulated Score Difference (ASD)}---defined as the cumulative divergence between conditional and unconditional score predictions during CFG---strongly correlates with a sample's likelihood of originating from high-density regions of the data manifold. Empirically, we find that trajectories exhibiting high ASD are more prone to yield high-fidelity samples, whereas low-ASD paths often culminate in outliers or aesthetically unpleasing results. Leveraging this insight, CFG-Rejection delivers its advantages through two core mechanisms:

\begin{itemize}
\item \textbf{Reward-Free Evaluation}: Unlike prior methods reliant on external evaluators, CFG-Rejection leverages intrinsic signals already present within the classifier-free guidance mechanism in the latent space, enabling fully self-contained and cost-free sample quality assessment.
    \item \textbf{Early-Stage Filtering}: By monitoring ASD in the initial steps of the denoising process, our method can prematurely terminate low-quality sampling trajectories, significantly reducing computational expenditure.
\end{itemize}

To summarize, our contributions are delineated as follows:
\begin{itemize}
    \item We are the first to reveal and empirically validate a strong correlation between final sample quality and the Accumulated Score Differences (ASD) within classifier-free guidance---an intrinsic, previously underexplored signal within the diffusion sampling path.
    \item We propose CFG-Rejection, a novel, plug-and-play filtering strategy that leverages ASD for early trajectory pruning. This enhances sample quality and efficiency without requiring model retraining or reliance on external, score-based post-selection. By design, CFG-Rejection is versatile and readily applicable across diverse diffusion models and downstream tasks.
    \item We demonstrate the efficacy of CFG-Rejection through extensive experiments on ImageNet~\cite{deng2009imagenet} and challenging benchmarks such as GenEval~\cite{ghosh2023geneval} and DPG-Bench~\cite{hu2024ella}. Our method shows consistent improvements across human preference metrics including PickScore~\cite{kirstain2023pick}, Aesthetic Score~\cite{schuhmann2022laion}, and HPSv2~\cite{wu2023human}, underscoring its potential as a standard 'zero-cost' technique for evaluating sample quality in diffusion models.
\end{itemize}

\begin{figure}[H]
\centering  
{
    \begin{minipage}[b]{\linewidth}
        \centering
        \includegraphics[width=\linewidth]{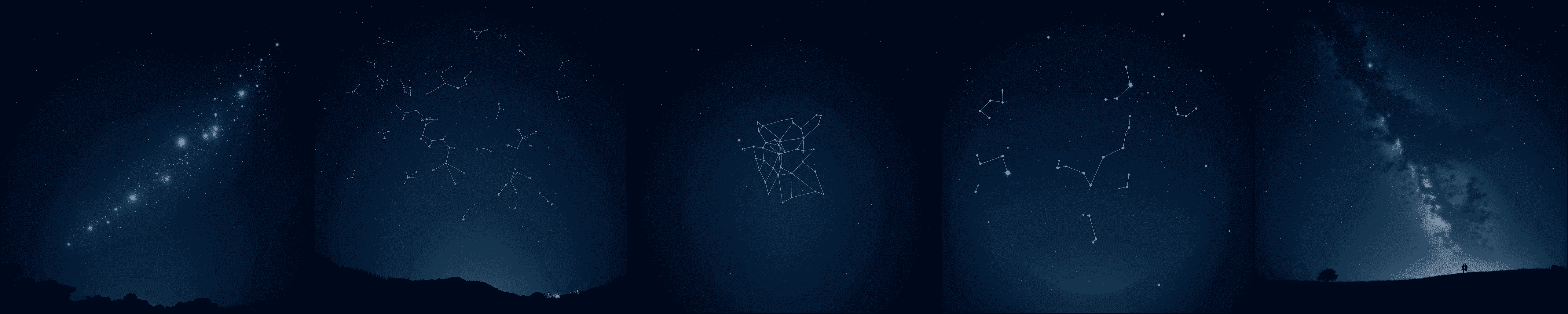}
    \end{minipage}
}%
{
    \begin{minipage}[b]{\linewidth}
        \centering
        \includegraphics[width=\linewidth]{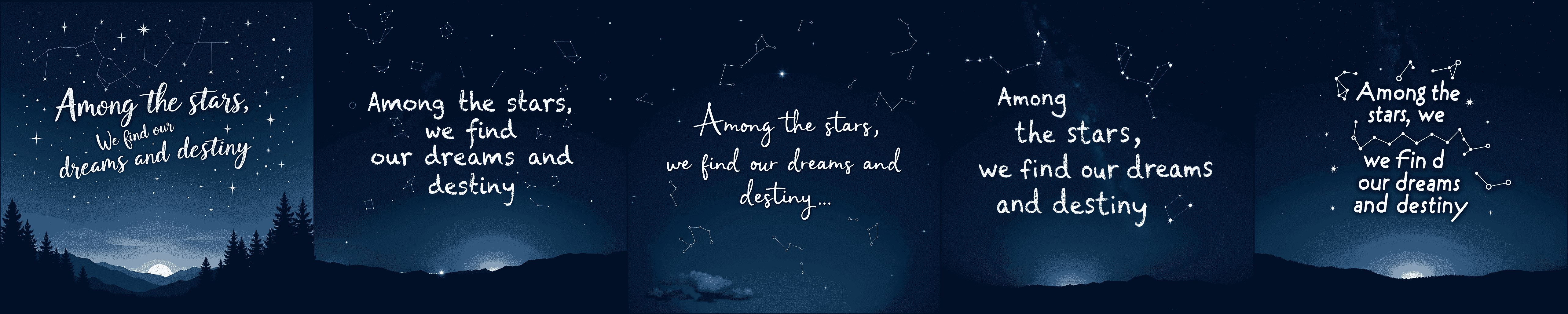}
    \end{minipage}
}%
  \caption{The qualitative comparison of filtering results demonstrates the effectiveness of our method in the text alignment of complex prompts. Prompt: "A night sky with constellations forming the words 'Among the stars, we find our dreams and destiny'". Low-ASD images (top row) exhibit completely missing strokes, while high-ASD samples (bottom row) ensure textual requirements.}
  \label{flux}
\end{figure}
\vspace{-0.3cm}

Finally, in Figure \ref{flux}, we provide a visual demonstration to give an early glimpse into the remarkable effectiveness of the ASD metrics in distinguishing the distribution of good and bad samples.

\vspace{-0.2cm}
\section{Background}
\vspace{-0.1cm}
\subsection{Diffusion Model}
\vspace{-0.1cm}
Denoising diffusion models generate samples by reversing a forward noising process~\cite{ho2020ndm}, where data is gradually corrupted by Gaussian noise:
\begin{equation}
p(\mathbf{x};\sigma) = p_{\text{data}}(\mathbf{x}) \ast \mathcal{N}(\mathbf{x};\mathbf{0}, \sigma^2\mathbf{I}),
\end{equation}
in which \( \ast \) denotes convolution and \(\sigma \in [0, \sigma_{\max}]\) controls the noise scale. As \(\sigma\) increases, the data distribution becomes increasingly Gaussian, and eventually approaches \(\mathcal{N}(\mathbf{x};\mathbf{0}, \sigma_{\max}^{2}\mathbf{I}) \) at the maximum noise level.

Sampling is performed by solving a reverse-time Ordinary Differential Equation (ODE)~\cite{song2021sde,karras2022elucidating}:
\begin{equation}
    \mathrm{d}\mathbf{x}_\sigma = -\sigma \nabla_{\mathbf{x}_\sigma} \log p(\mathbf{x}_\sigma;\sigma) \mathrm{d}\sigma.
    \label{eq:ode}
\end{equation}
In practice, the score function is approximated by a neural network \( S_\theta(\mathbf{x};\sigma) \), and samples are obtained via discretized integration from \(\sigma_{T}=\sigma_{max}\) to \(\sigma_{0}=0\).

\vspace{-0.1cm}
\subsection{Classifier-free guidance}
\vspace{-0.1cm}
To condition the generation of inputs such as text prompts \(\mathbf{c}\), classifier-free guidance (CFG)~\cite{ho2022classifier} uses a linear combination of conditional and unconditional score estimates:
\begin{equation}
S_w(\mathbf{x}; \sigma, \mathbf{c}) 
= w S_\theta(\mathbf{x}; \sigma, \mathbf{c}) 
+ (1-w) S_\theta(\mathbf{x}; \sigma, \emptyset),
\label{eq:weighted_model}
\end{equation}
where \(\omega \geq 0\) controls the guidance strength. This corresponds to modifying the score function as:
\begin{equation}
\nabla_{\mathbf{x}} \log p_w(\mathbf{x} \mid \sigma,\mathbf{c}) 
= \nabla_{\mathbf{x}} \log p(\mathbf{x} \mid \sigma,\mathbf{c}) 
+ (w-1) \nabla_{\mathbf{x}} \log \frac{p(\mathbf{x} \mid \sigma,\mathbf{c})}{p(\mathbf{x} \mid \sigma,\emptyset)}.
\label{rewrite_score}
\end{equation}
The guidance term \(\nabla_{\mathbf{x}} \log \frac{p(\mathbf{x} \mid \sigma,\mathbf{c})}{p(\mathbf{x} \mid \sigma,\emptyset)}\) encourages alignment with the conditional distribution while suppressing low-likelihood modes~\cite{dhariwal2021diffusion}. Recent work~\cite{karras2024guiding} further interprets this as a form of adaptive truncation, directing underfit trajectories to better regions of the sample space. Adjusting the guidance weight \(\omega\), one can control the trade-off between fidelity and diversity.

\vspace{-0.1cm}
\subsection{Inference-Time Alignment}
\vspace{-0.1cm}
Beyond classifier-free guidance (CFG), many inference-time alignment methods leverage external reward signals (e.g. PickScore~\cite{kirstain2023pick}) to guide generation by evaluating the final outputs. While effective, these approaches share two key limitations: (1) they require full denoising for all candidates, and (2) they rely on external reward models applied post-generation, introducing substantial computational cost and potential domain mismatch.

These methods typically frame alignment as a reward maximization problem, and can be grouped into three categories:

\begin{itemize}
\item \textbf{Direct noise optimization:}
Methods like DNO~\cite{tang2024inference} and RENO~\cite{eyring2024reno} perform gradient-based optimization in the latent space to adjust noise vectors and maximize final reward scores. This involves iterative forward and reverse sampling, which is computationally expensive.
\item \textbf{Random search based on reward:}
Best-of-N sampling~\cite{ma2025inference} generates multiple candidates and selects the best based on a reward model acting as a final-stage verifier. This results in significant waste, as all samples must be fully denoised before evaluation.
\item \textbf{Reward-guided generation:}
Methods like Loss-Guided Diffusion (LGD)~\cite{song2023loss} integrates reward gradients into the denoising process, treating the reward model as a dynamic signal to steer generation. While enabling online guidance, it introduces instability and remains dependent on external supervision.
\end{itemize}

Despite differences in implementation, these approaches share common drawbacks: all require full denoising and depend on external reward models to evaluate final outcomes. In contrast, our proposed CFG-Rejection method sidesteps these limitations by leveraging intrinsic signals available during early denoising. This enables lightweight, reward-free filtering, achieving efficient alignment with negligible overhead.

\vspace{-0.2cm}
\section{Our Method}
\vspace{-0.1cm}
We begin by presenting key empirical findings showing that Accumulated Score Differences (ASD) correlate strongly with sample density. Based on this, we derive CFG-Rejection to improve sample fidelity through efficient filtering.

\vspace{-0.1cm}
\subsection{Unveiling the information in ASD}
\begin{figure}[htbp]
\centering  
\subfigure[Samples with CFG]{   
\begin{minipage}{0.4\textwidth}
\centering   
\includegraphics[width=0.9\linewidth]{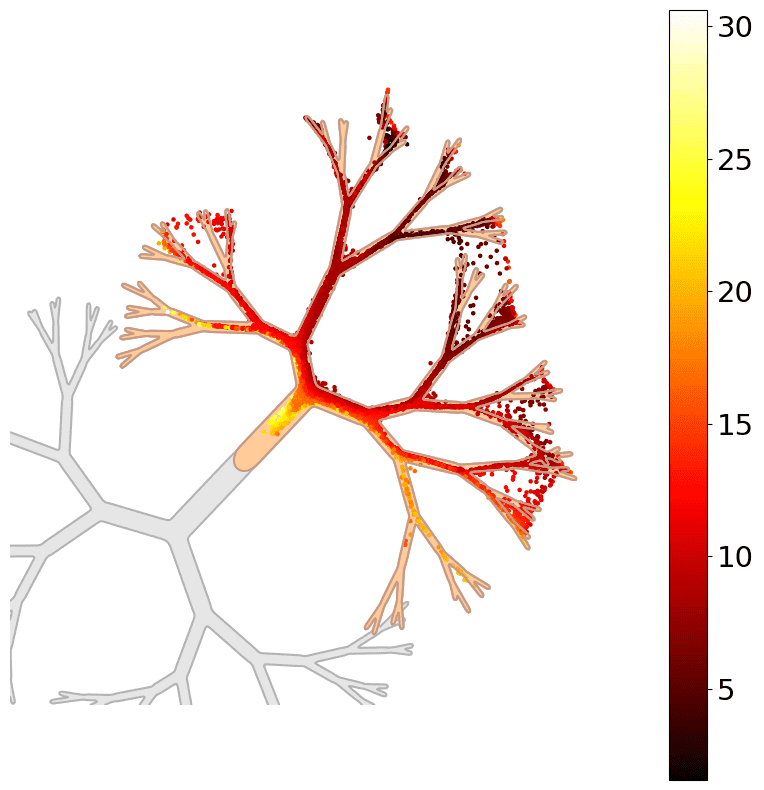}  
\label{toy_images}
\end{minipage}
}
\subfigure[Positive relationship]{
\begin{minipage}{0.4\textwidth}
\centering   
\includegraphics[width=1.1\linewidth]{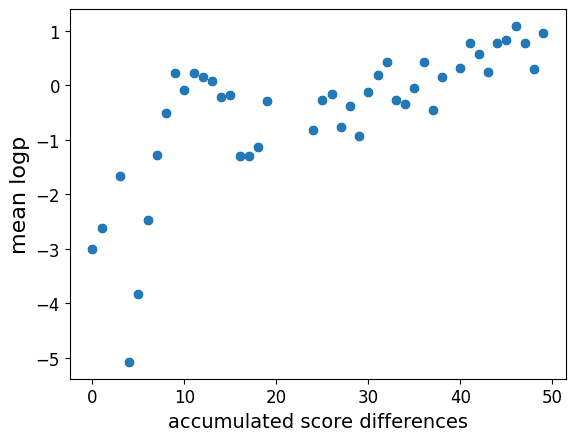}
\label{toy_relation}
\end{minipage}
}
\label{toy examples}
\caption{A fractal-like 2D distribution with two classes (gray and orange). (\textbf{a}) Samples generated with CFG (\(\omega=2\)) are color-coded by ASD. High-ASD samples concentrate in the dense trunk. (\textbf{b}) A log-linear trend emerges between local density and ASD, indicating that large score differences align with high-likelihood regions.}
\end{figure}
\vspace{-0.2cm}

Motivated by classifier-free guidance (CFG), we investigate how guidance magnitudes influence sample fidelity. Specifically, we track the cumulative score differences between conditional and unconditional models during denoising, which we refer to as Accumulated Score Differences (ASD), and examine the resulting sample positions in a controlled setup. We adopt a 2D toy distribution from~\cite{karras2024guiding}, designed to mimic properties of real data manifolds such as low local dimensionality and high anisotropy~\cite{brown2022verifying,pope2021intrinsic}. The complete setup details are provided in the Appendix~\ref{toy_app}.

As shown in Fig.~\ref{toy_images}, this distribution features a high-density central trunk and sparse peripheral branches, with two classes marked in gray and orange. Samples are color-coded by their ASD values. We observe that high-ASD samples concentrate in the dense central region with strong class consistency, while low-ASD samples tend to appear in sparse regions with weaker semantic alignment. Degenerate samples, those that deviate significantly from the conditioning, often have near-zero ASD. These outliers correspond to broken images in practical text-to-image generations.

This geometric pattern is further quantified in Fig.~\ref{toy_relation}, which reveals a strong positive correlation between the local sample density and the accumulated score differences. These findings motivate a latent-space filtering principle: by computing ASD during generation, we can identify and prune low-likelihood trajectories while preserving high-quality ones.

\vspace{-0.1cm}
\subsection{CFG-Rejection: A versatile sample selection approach}
\vspace{-0.1cm}
We formalize our filtering strategy by measuring the influence of conditional signals during denoising. At each denoising step \(t\), we define the \textit{score differences} as:
\begin{equation}
\mathcal{G}_t(c) = \| S_\theta(\mathbf{x}_t;\sigma_t,\mathbf{c}) - S_\theta(\mathbf{x}_t;\sigma_t,\emptyset) \|_2.
\label{nv_gap}
\end{equation}
This term reflects how much conditioning alters the denoiser’s prediction, with larger differences indicating stronger alignment with the prompt.

\vspace{-0.1cm}
\paragraph{Accumulated Score Differences} To capture this alignment over time, we define the \textit{Accumulated Score Differences} (ASD) as the sum of squared score differences across all denoising steps:
\begin{equation}
\mathcal{E}_{T}(c) = \sum_{t=1}^T \mathcal{G}_t(c)^2.
\end{equation}
ASD acts as a global indicator of conditional influence throughout the generation. Unlike instantaneous metrics, it captures the temporal consistency of alignment, allowing us to distinguish trajectories that align with the conditional distribution from those that deviate into low-density or semantically irrelevant regions. Empirical observations on toy data reveal a strong positive correlation between ASD and sample density region. These findings are further supported by large-scale experiments on the ImageNet dataset, where ASD proves predictive of prompt consistency and perceptual quality.

\vspace{-0.1cm}
\paragraph{CFG-Rejection} A naive approach would evaluate \(\mathcal{E}_{T}(c)\) post hoc and retain the top k samples, which incurs full denoising costs for all candidates. Observing that \(\mathcal{G}_t(c)\) typically decreases in later steps as the generative process becomes increasingly deterministic, we propose CFG-Rejection strategy based on the partial accumulation of score differences. Formally, for a hyperparameter \(\tau \in [1,T]\) defining the cutoff step: 
\begin{equation}
\mathcal{E}_{\tau:T}(c) = \sum_{t=T-\tau}^T \mathcal{G}_t(c)^2,
\end{equation}
samples with \(\mathcal{E}_{\tau:T}(c) < \gamma\) (where \(\gamma\) is a percentile-based threshold) are discarded without completing full denoising, ensuring more efficient filtering of low-quality candidates. 

The proposed CFG-Rejection framework consists of three key steps:
1) \textbf{Score difference tracking}: record the instantaneous \(\mathcal{G}_t(c)\) at each denoising step. 2) \textbf{Partial differences accumulation}: compute the cumulative score differences \(\mathcal{E}_{\tau:T}(c)\) from a predefined step \(\tau\). 3) \textbf{Sample selection}: discard low-potential trajectories based on a threshold \(\gamma\). Our method integrates seamlessly into existing generation pipelines with minimal computational overhead. By enabling early-stage filtering through an intrinsic, self-contained metric, it substantially reduces the cost associated with full denoising and reliance on external reward models.

\vspace{-0.2cm}
\section{Experiments}
\vspace{-0.1cm}
We conduct extensive experiments to evaluate the effectiveness of CFG-Rejection across architectures, datasets, and tasks. Our study is guided by two key questions: 1) Does the correlation between \(\mathcal{E}_{T}(c)\) and sample quality persist in modern diffusion paradigms? 2) How does the early termination threshold \(\tau \in \{1,\dots,T\}\) affect the trade-off between quality and compute? To answer (1), we evaluate three diffusion models (EDM2~\cite{karras2024analyzing}, SDv1.5, SDXL~\cite{sdxl}) on three benchmarks (ImageNet~\cite{deng2009imagenet}, GenEval~\cite{ghosh2023geneval}, DPG-Bench~\cite{hu2024ella}), and additionally test on the FLUX model for text rendering tasks. For (2), we analyze the impact of varying \(\tau\) using both human preference metrics (PickScore~\cite{kirstain2023pick}, Aesthetic Score~\cite{schuhmann2022laion}, HPSv2~\cite{wu2023human}) and automated quality indicators.

\vspace{-0.1cm}
\subsection{ImageNet Benchmark Analysis}
\vspace{-0.2cm}
\begin{figure}[H]
\centering
\subfigure[Golden retriever]{  
\begin{minipage}{0.3\textwidth}
\centering
\includegraphics[width=\linewidth]{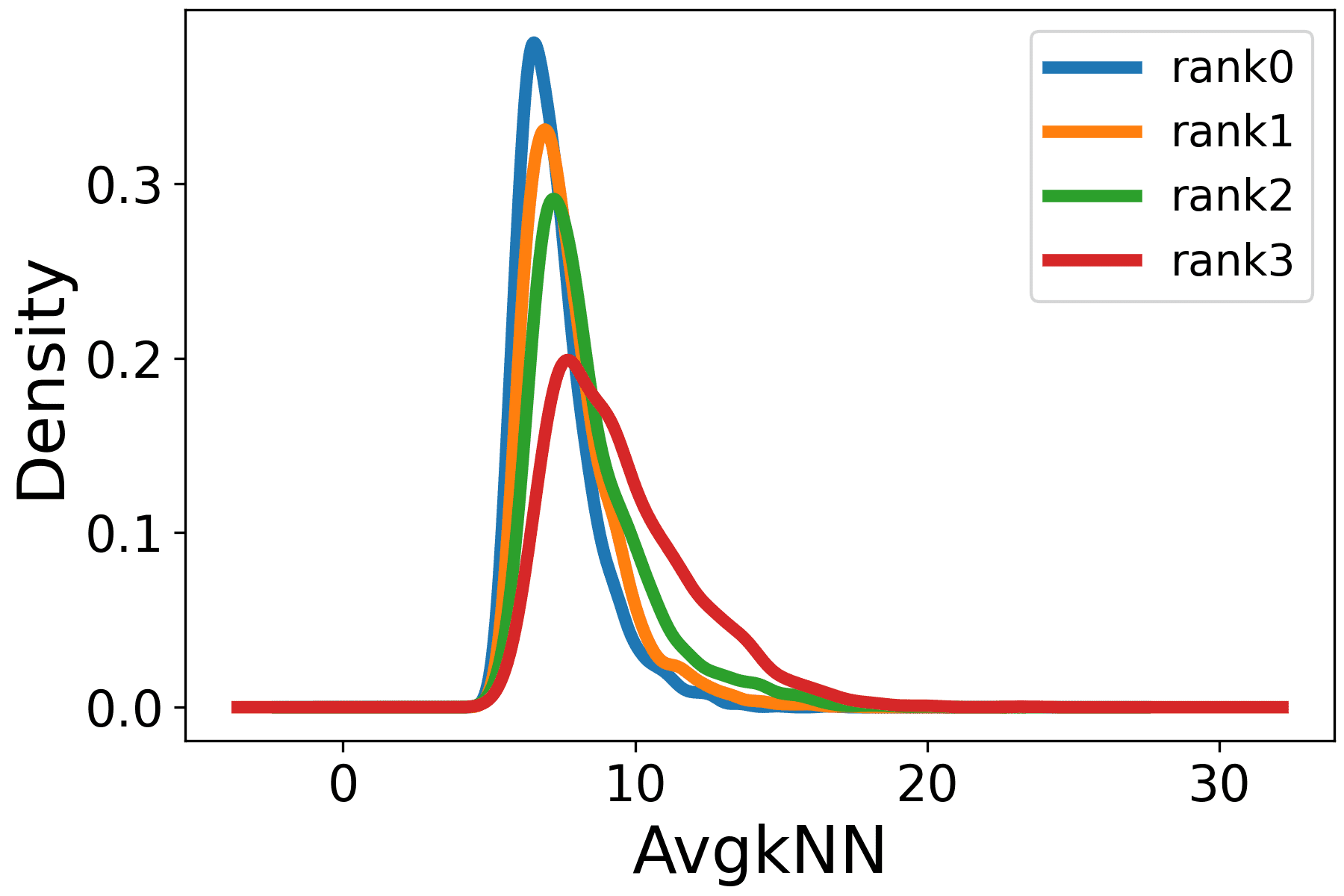} 
\includegraphics[width=\linewidth]{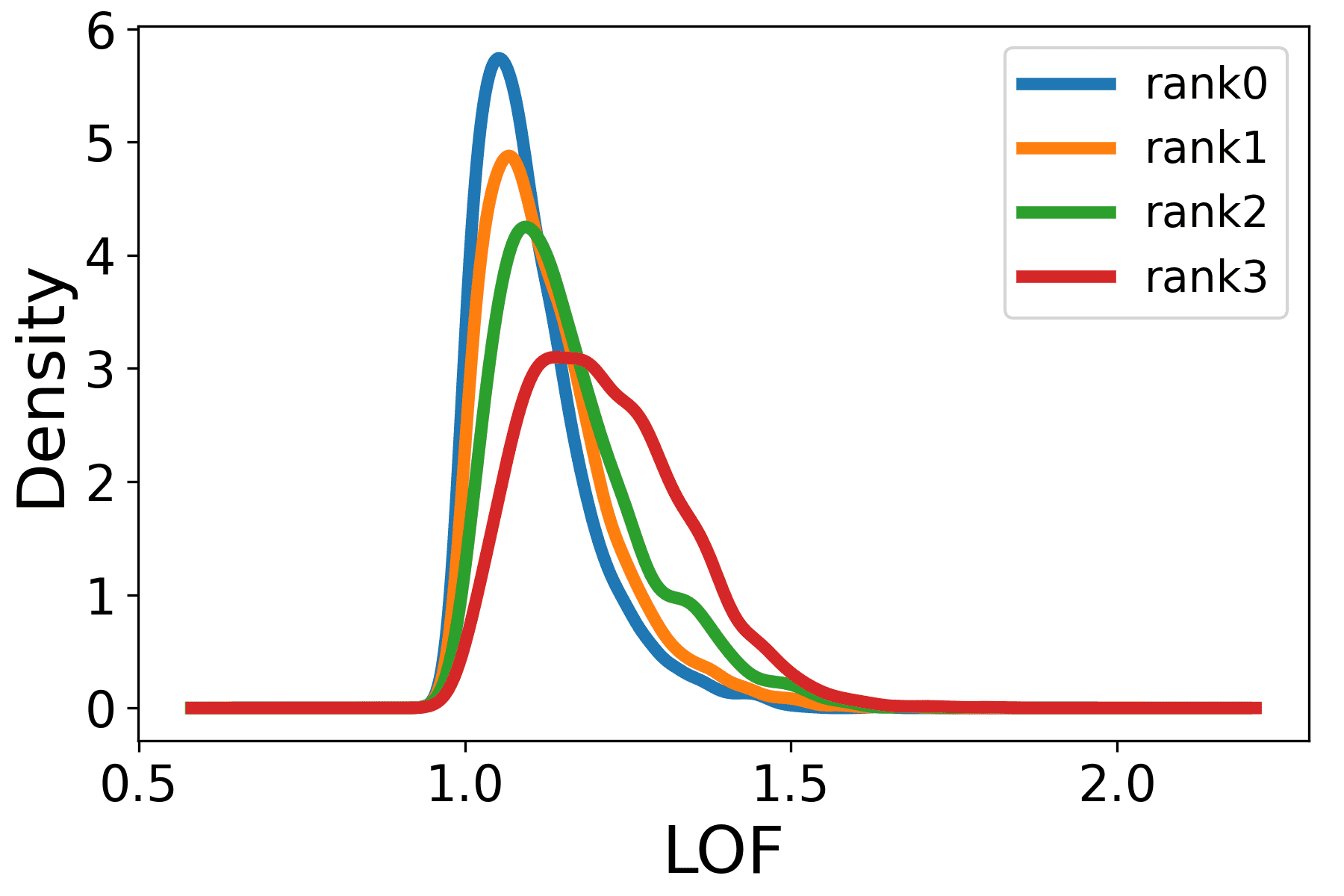} 
\label{207_density}
\end{minipage}
}
\subfigure[Giant panda]{  
\begin{minipage}{0.3\textwidth}
\centering
\includegraphics[width=\linewidth]{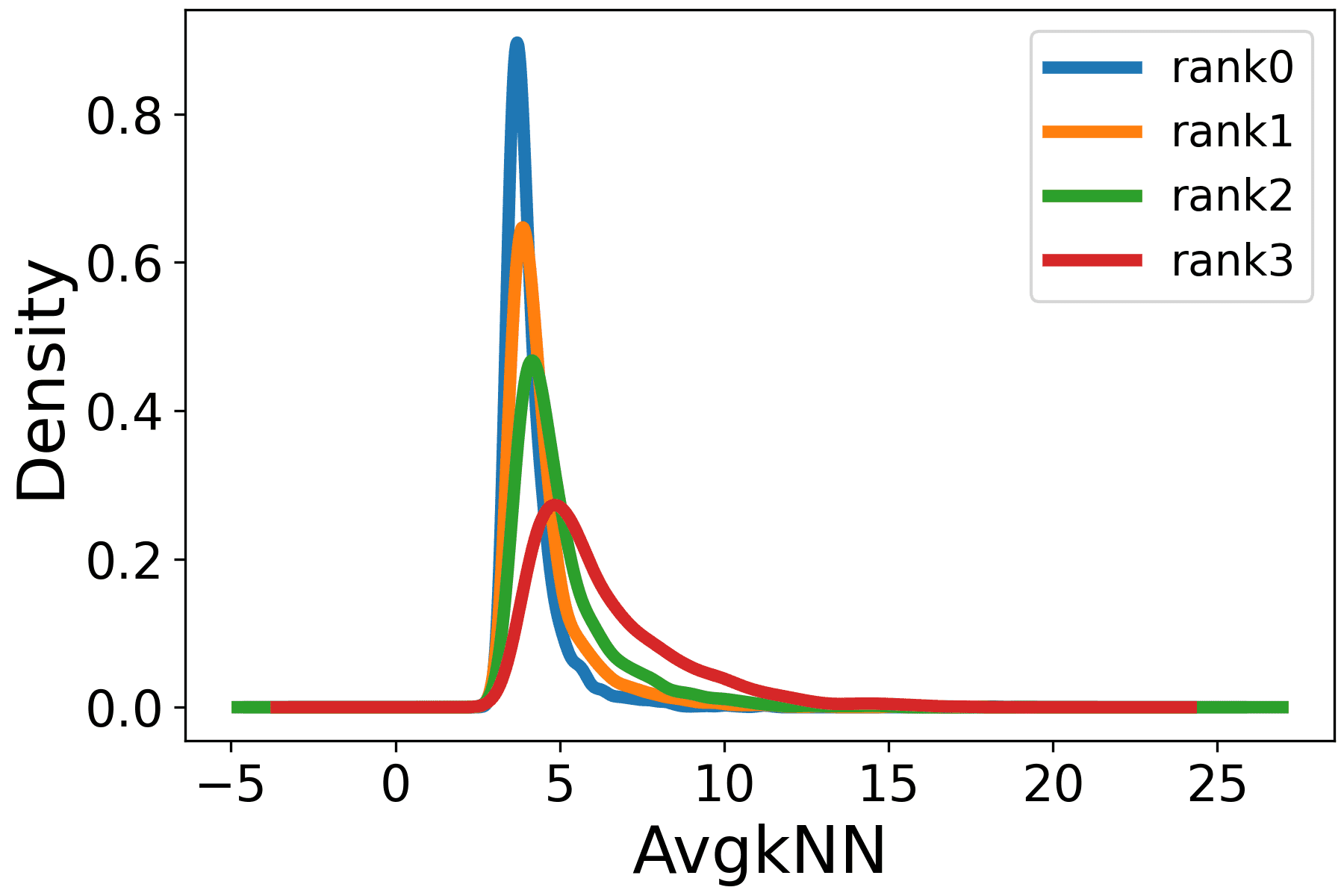}
\includegraphics[width=\linewidth]{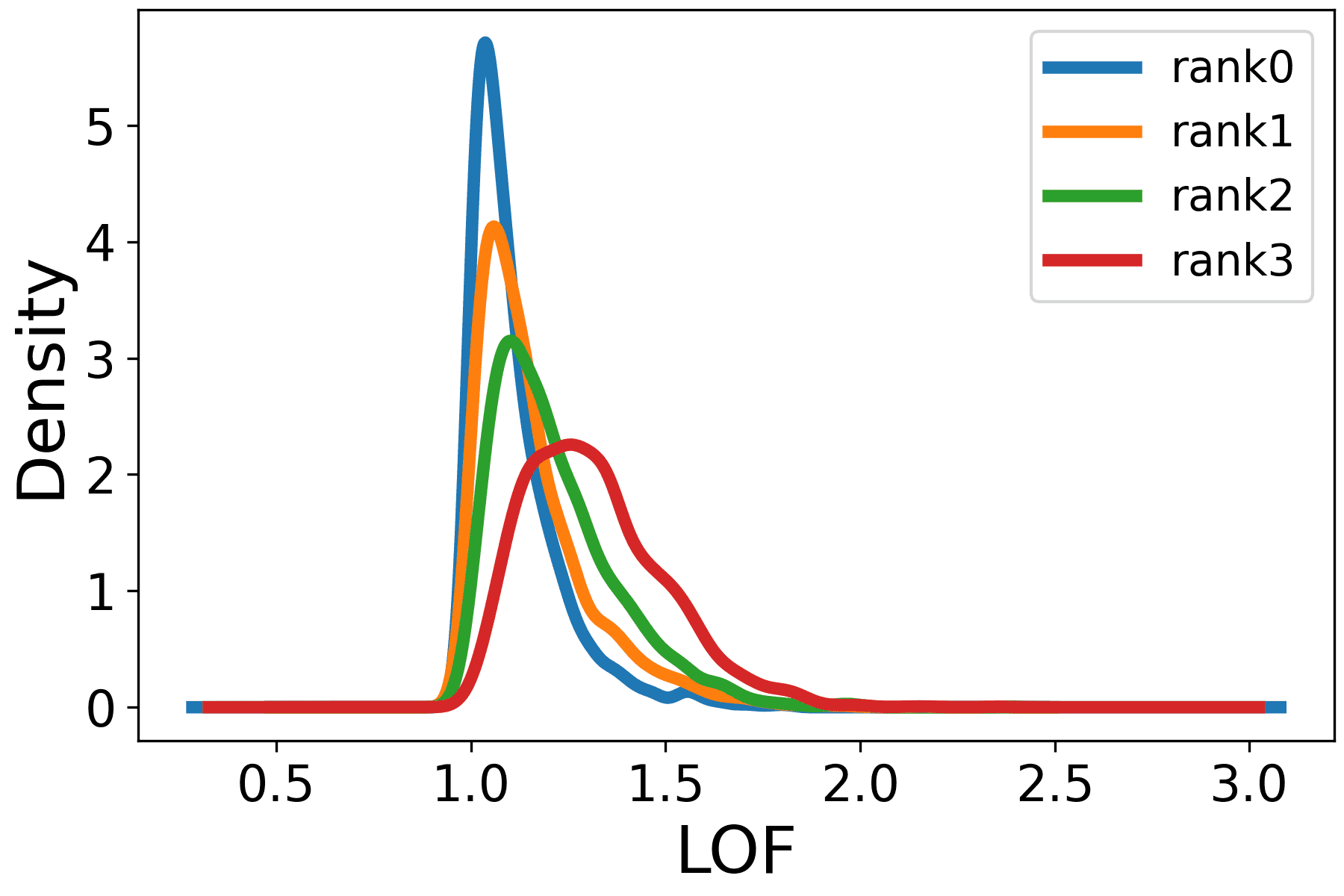}
\label{388_density}
\end{minipage}
}
\subfigure[Strawberry]{  
\begin{minipage}{0.3\textwidth}
\centering
\includegraphics[width=\linewidth]{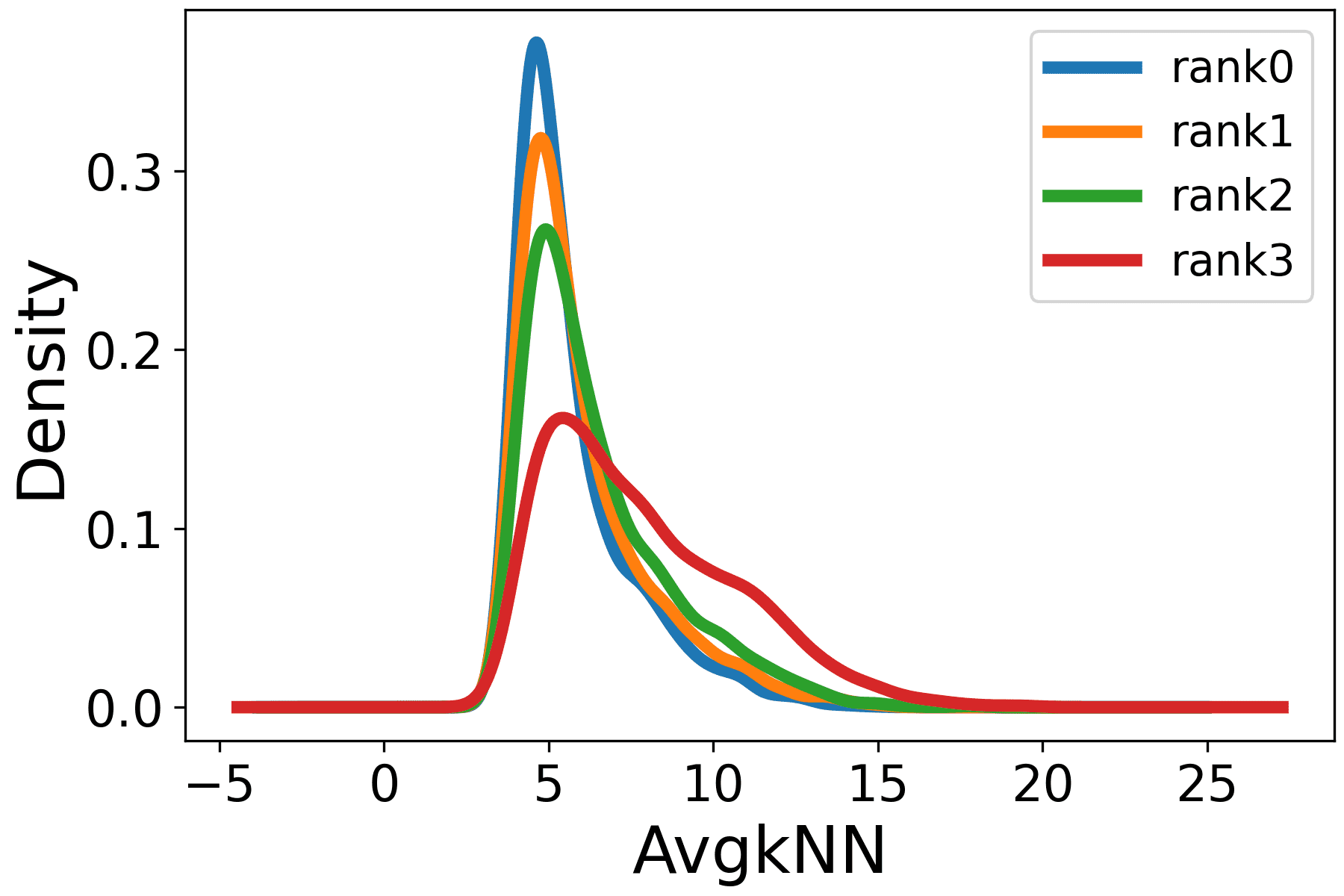}
\includegraphics[width=\linewidth]{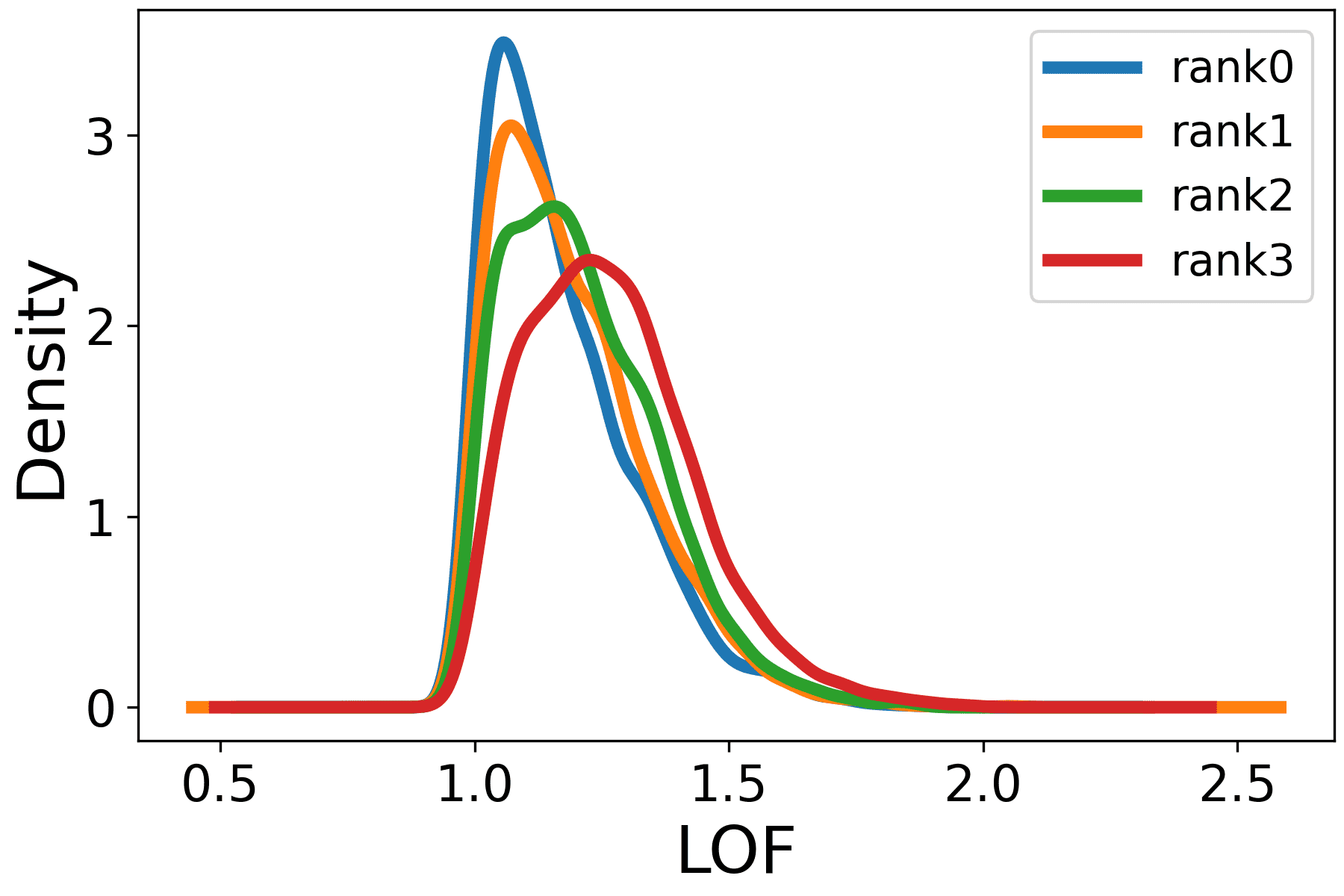}
\label{949_density}
\end{minipage}
}
\caption{Density estimation curves for samples with varying accumulated score differences. The top and bottom rows display the AvgkNN and LOF density profiles, respectively, across three distinct labels. As the ASD decreases from rank 0 (highest) to rank 3 (lowest), we observe a systematic shift of samples from high-density to low-density regions.} 
\label{ImageNet_density}
\end{figure}
\vspace{-0.2cm}

We evaluate CFG-Rejection on EDM2-S~\cite{karras2024analyzing} with Heun's sampler (32 steps), generating 10k samples in 50 random ImageNet classes. To test whether \(\mathcal{E}_{T}(c)\) correlates with sample likelihood, we use two density estimators: Average k-Nearest Neighbor (AvgkNN) and Local Outlier Factor (LOF)\cite{breunig2000lof}, where higher scores indicate lower manifold likelihood\cite{sehwag2022generating}. 

As shown in Fig.~\ref{ImageNet_density}, samples with lower ASD shift toward higher density scores, suggesting that they occupy low-likelihood regions. This aligns with our findings on the toy example, further confirming the positive correlation between ASD and sample density regions. Additional density curves for other labels can be found in Appendix~\ref{Ima_density}.

\begin{figure}[H]
\centering
\subfigure[Golden retriever]{  
\begin{minipage}{0.313\textwidth}
\centering
\includegraphics[width=\linewidth]{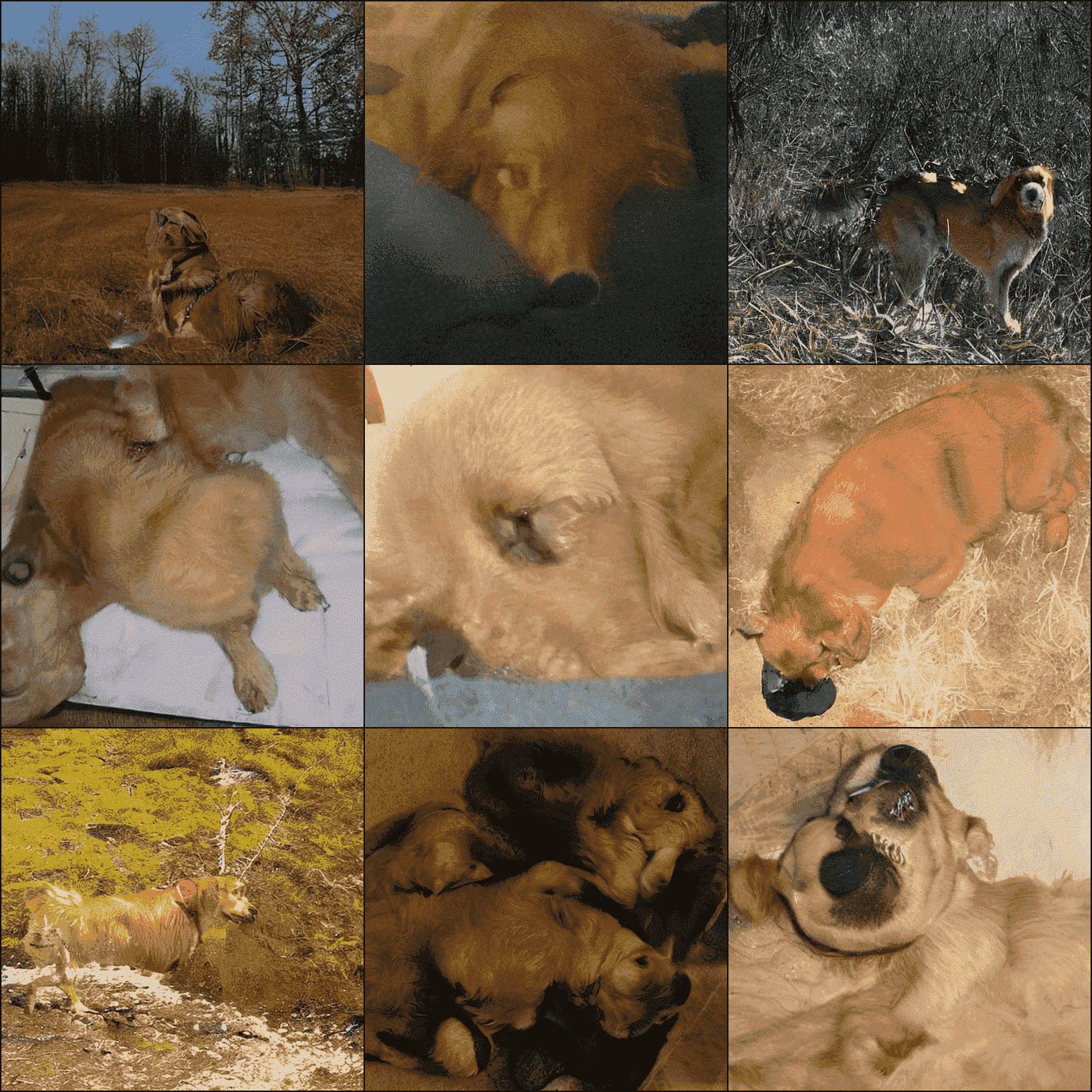} 
\includegraphics[width=\linewidth]{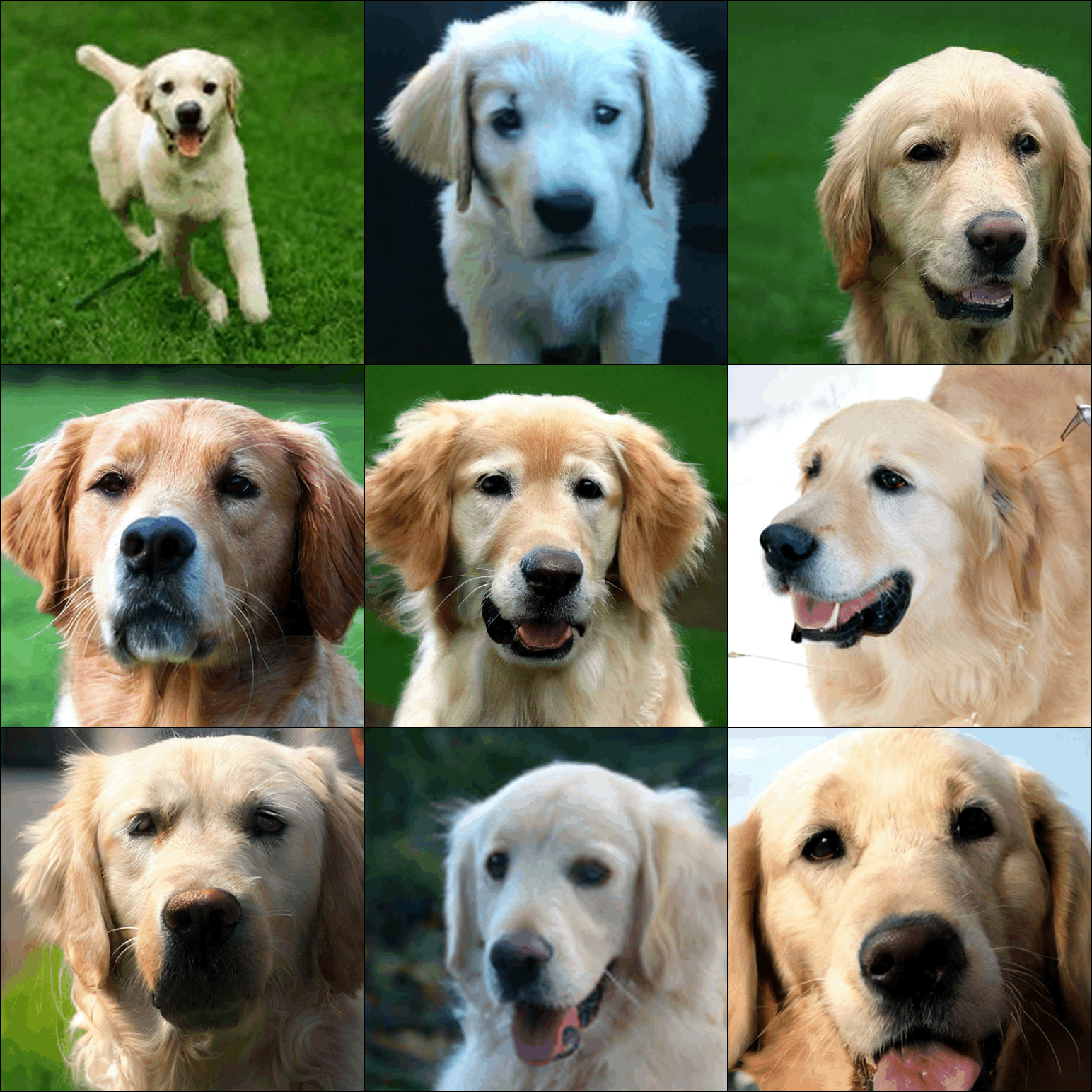} 
\label{207_img}
\end{minipage}
}
\subfigure[Giant panda]{  
\begin{minipage}{0.313\textwidth}
\centering
\includegraphics[width=\linewidth]{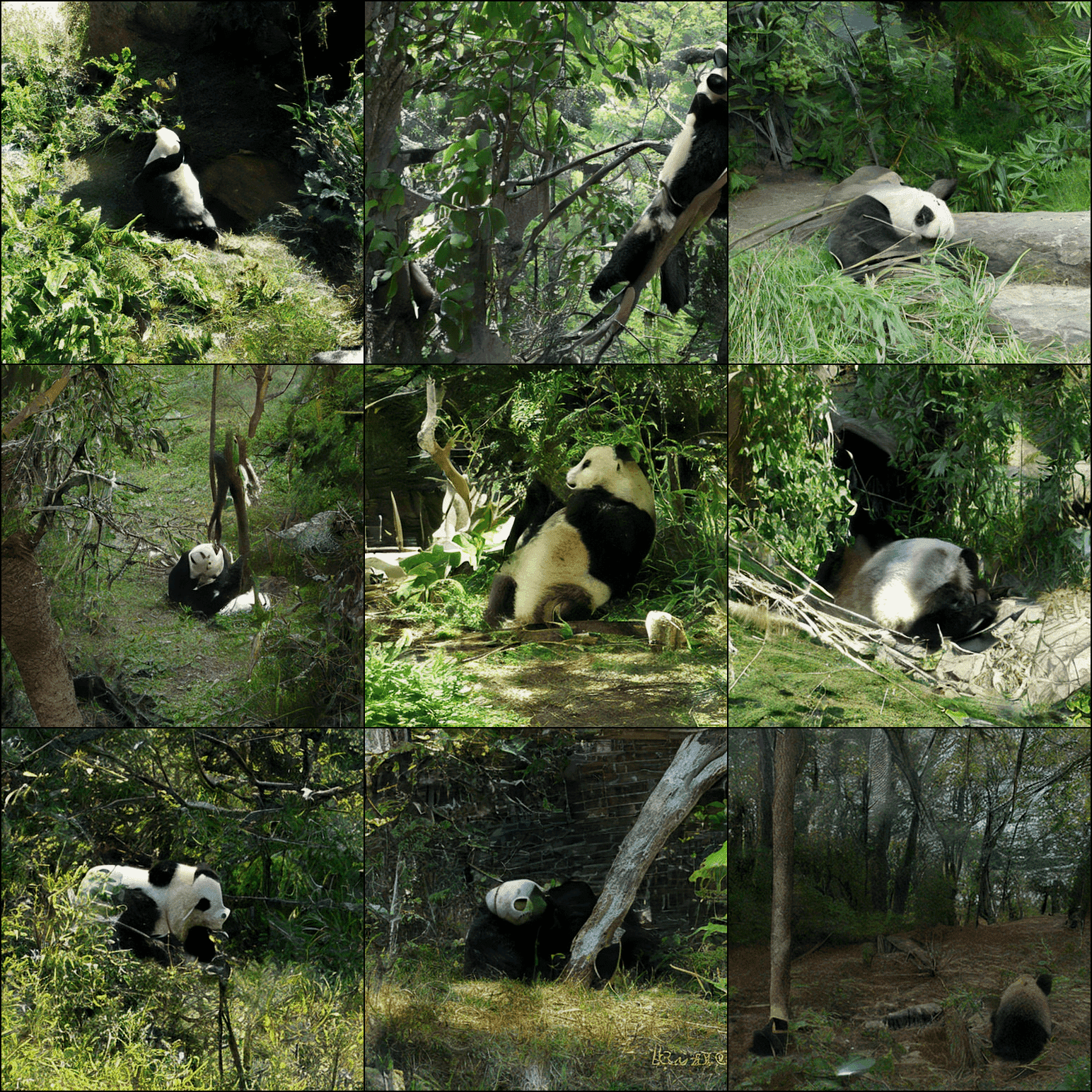}
\includegraphics[width=\linewidth]{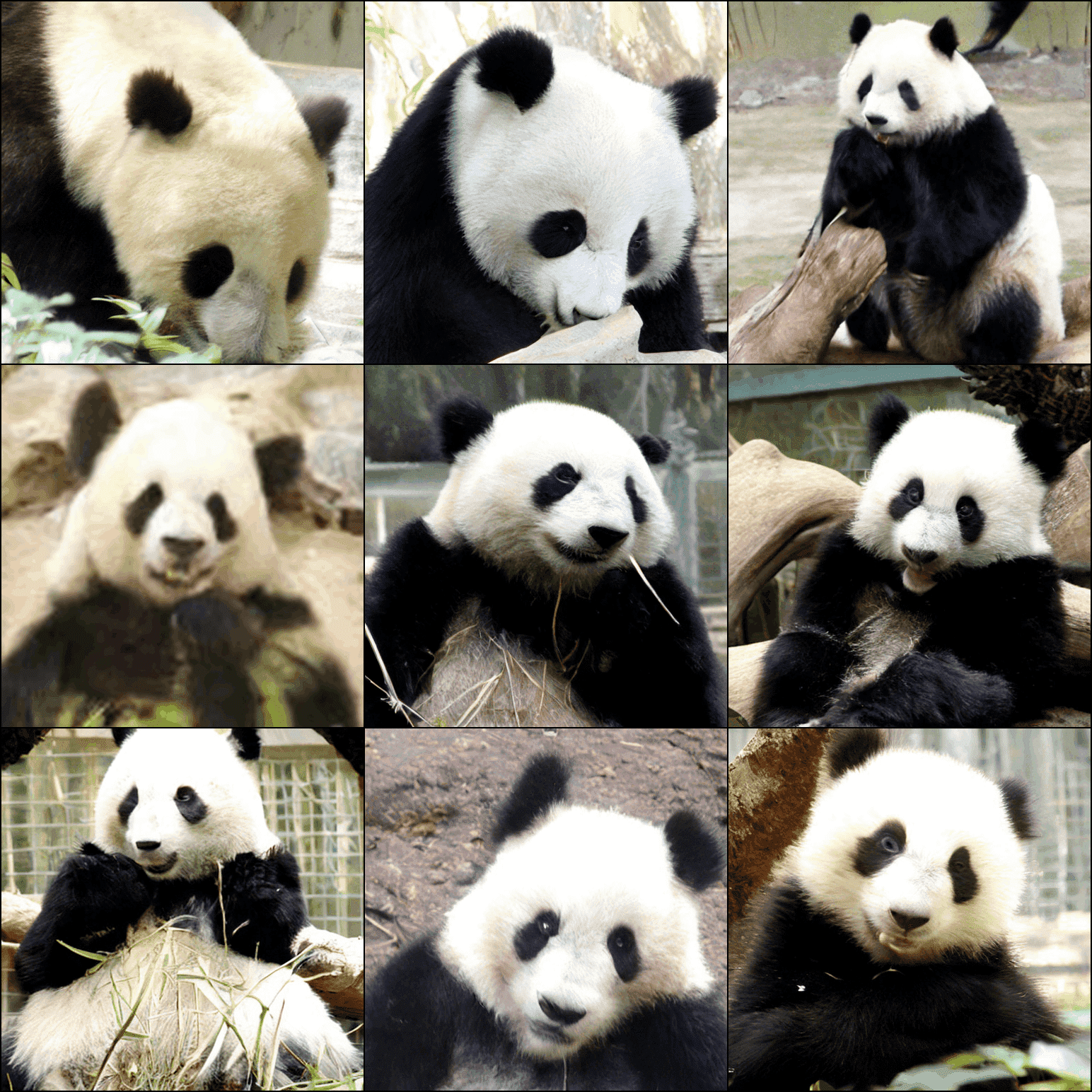}
\label{388_img}
\end{minipage}
}
\subfigure[Strawberry]{  
\begin{minipage}{0.313\textwidth}
\centering
\includegraphics[width=\linewidth]{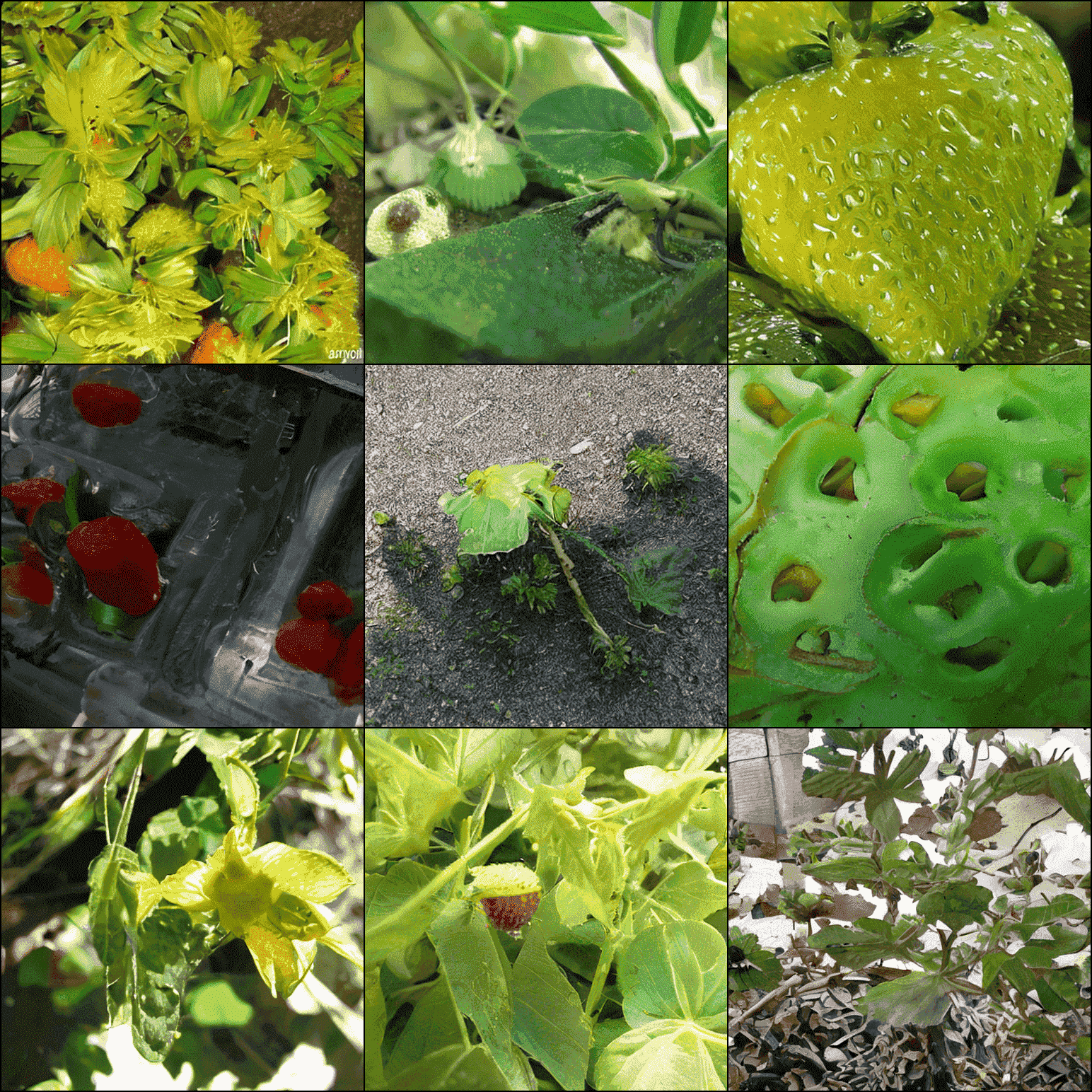}
\includegraphics[width=\linewidth]{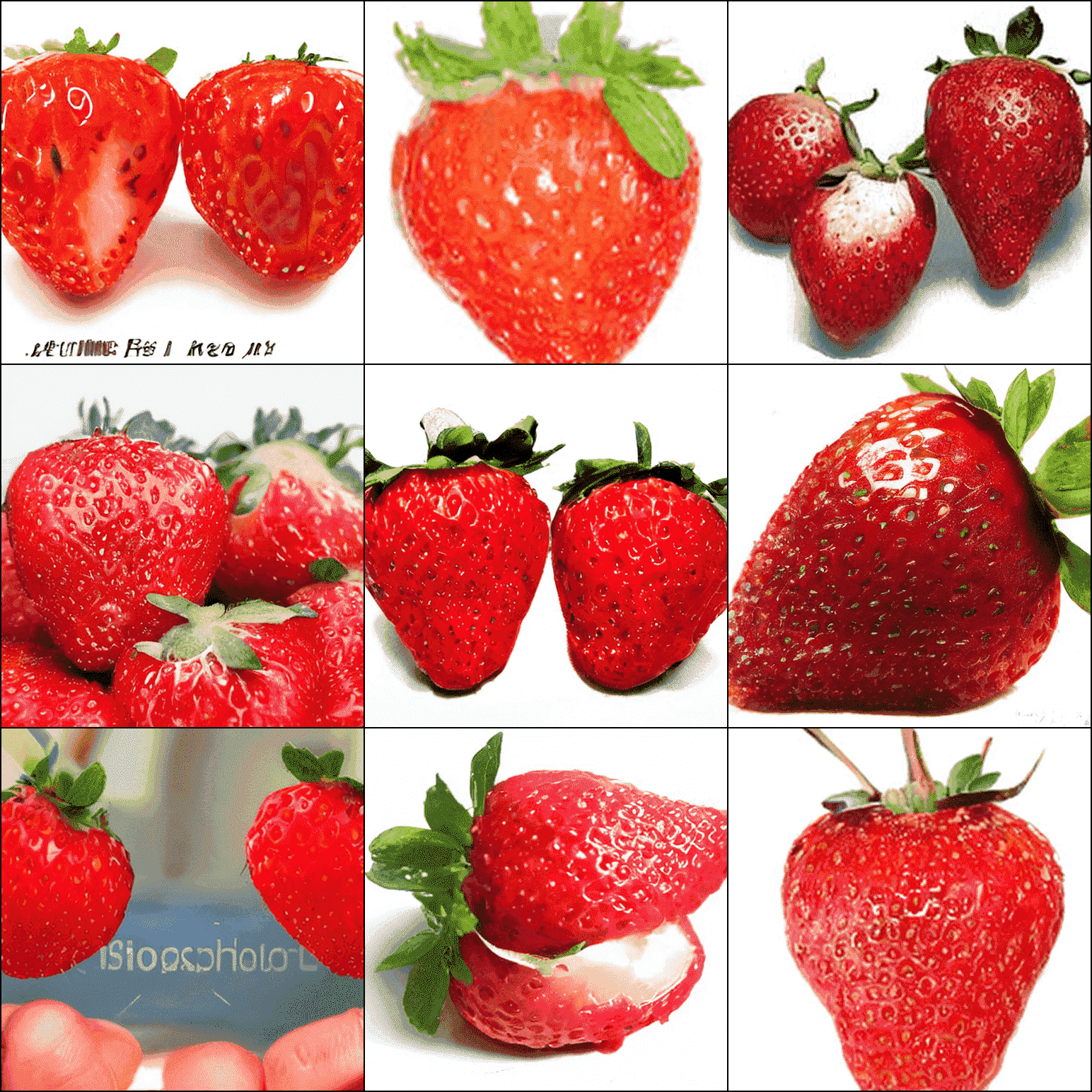}
\label{949_img}
\end{minipage}
}
\caption{Qualitative comparison on the ImageNet dataset. (Top) Baseline samples with the lowest \(\mathcal{E}_{T}(c)\) exhibit artifacts and misalignment. (Bottom) Our selected samples with the highest \(\mathcal{E}_{T}(c)\) show better fidelity and prompt adherence.} 
\label{ImageNet}
\end{figure}

\vspace{-0.2cm}
\paragraph{Qualitative comparison} 
We compare samples with extreme \(\mathcal{E}_{T}(c)\) values (Fig.\ref{ImageNet}). Samples with high-ASD exhibit better semantic alignment and visual coherence, while samples with low-ASD often show structural failures or class mismatches. For instance, in the giant panda class, first-row samples frequently suffer from misaligned limbs and inconsistent panda features, whereas second-row samples consistently reflect key attributes such as the iconic black-and-white fur and natural panda posture. More examples are provided in Appendix\ref{Ima_qua}.

\vspace{-0.4cm}

\vspace{-0.1cm}
\paragraph{Quantitative comparison}
We evaluate all generated samples using three human preference metrics: PickScore~\cite{kirstain2023pick}, Aesthetic Score~\cite{schuhmann2022laion}, and HPSv2~\cite{wu2023human}. To assess the impact of our filtering strategy, we compare the average scores of all samples with those of the top 10\% selected by \(\mathcal{E}_{T}(c)\). As shown in Table~\ref{quan_imageNet}, filtering consistently improves quality across all metrics. A threshold of \(\tau=10\) is sufficient to select high-quality samples, with performance plateauing at \(\tau=20\), indicating that later score differences contribute marginally to perceived quality.

\vspace{-0.3cm}
\begin{table}[H]
  \centering
  \small
  \caption{The quantitative results on ImageNet dataset}
  \vspace{1em} 
  \label{quan_imageNet}
  \begin{tabular}{@{} l  *{7}{c}} 
    \toprule
     Metric & Full set & Top 10\% & \(\tau=5\) & \(\tau=10\) & \(\tau=15\) & \(\tau=20\) & \(\tau=25\) \\
     \midrule
     PickScore\(\uparrow\) & 20.38 & \textbf{20.61} & 20.49 & 20.52 & 20.60 & \textbf{20.61} & \textbf{20.61} \\
     AES\(\uparrow\) & 4.82 & \textbf{4.86} & 4.80 & 4.82 & 4.85 & \textbf{4.86} & \textbf{4.86} \\
     HPS v2\(\uparrow\)  & 26.13 & \textbf{26.57} & 26.26 & 26.31 & 26.48 & 26.55 & \textbf{26.57} \\
    \bottomrule
  \end{tabular}
\end{table}
\vspace{-0.3cm}

\begin{figure}[H]
\centering  
\subfigure[Pick Score comparison]{   
\begin{minipage}{0.4\textwidth}
\centering   
\includegraphics[width=\linewidth]{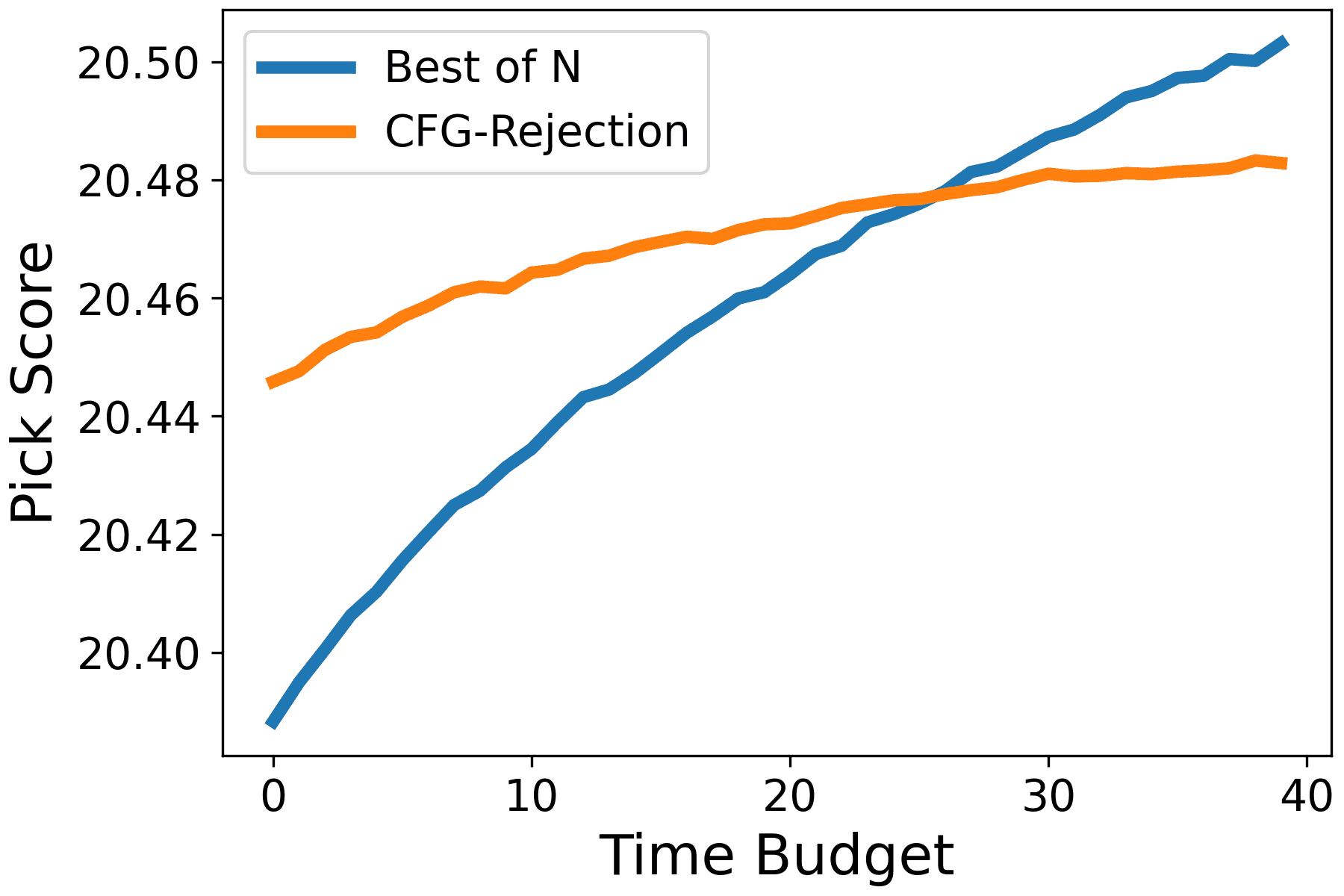}  
\label{pick}
\end{minipage}
}
\subfigure[HPSv2 Score comparison]{
\begin{minipage}{0.4\textwidth}
\centering   
\includegraphics[width=\linewidth]{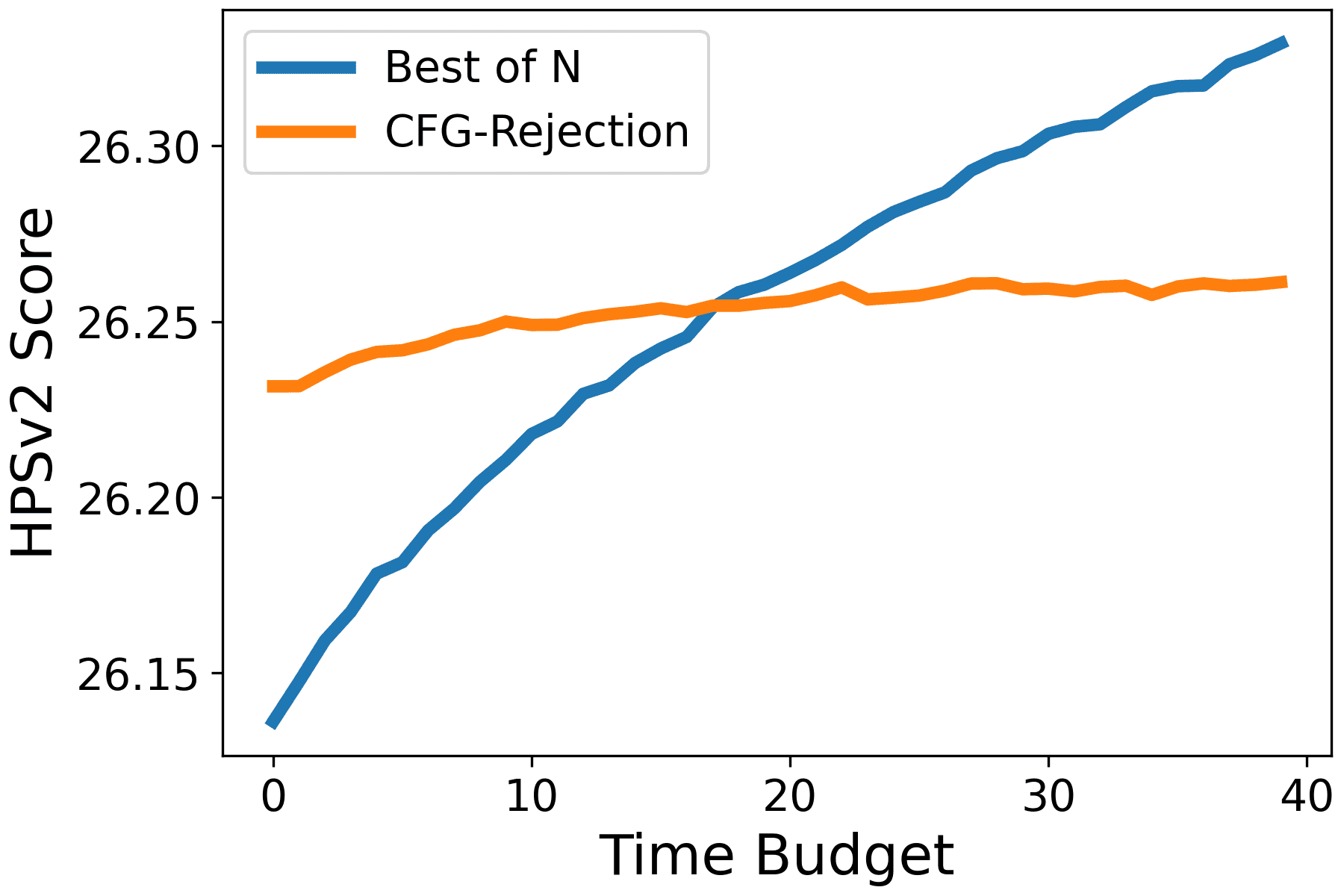}
\label{hpv2}
\end{minipage}
}
\caption{Performance comparison under limited inference budget. CFG-Rejection outperforms Best of N method under constrained computational budgets in practical usage.}
\label{compare_baseline}
\end{figure}

We further compare CFG-Rejection (\(\tau=5\)) with the Best-of-N strategy under identical time constraints (Appendix~\ref{Ima_baseline}). As shown in Fig.~\ref{compare_baseline}, Best-of-N improves gradually but lags initially due to the overhead of full denoising. In contrast, CFG-Rejection enables early pruning of low-quality paths, achieving higher scores under limited compute. This highlights its practical efficiency for time-sensitive generation.

\vspace{-0.3cm}

\vspace{-0.2cm}
\subsection{Large scale verifications}
\vspace{-0.1cm}
We further evaluate our method on two challenging benchmarks: GenEval~\cite{ghosh2023geneval} and DPG-Bench~\cite{hu2024ella}, using large-scale diffusion models: SDv1.5 and SDXL. For each model, we test two guidance scales and fix the number of inference steps to 50. Detailed configurations are provided in Appendix~\ref{bench_app}.

\vspace{-0.1cm}
\paragraph{GenEval Benchmark}
\label{paper_gen}
GenEval offers a fine-grained, instance-level evaluation of compositionality in text-to-image generation across 553 prompts and six compositional categories. To assess both efficiency and performance, we conduct two filtering schemes: (1) 4/20 sampling with varying \(\tau\) to examine compute savings, and (2) 4/50 full ablations to estimate performance ceilings.

\begin{table}[H]
  \centering
  \small
  \caption{The quantitative results on GenEval. Model: SDv1.5}
  \vspace{1em} 
  \label{sd1.5_gen}
  \begin{tabular}{@{} l l *{6}{c} c @{}}
    \toprule
    & Method& Single Obj. & Two Obj. & Counting & Colors & Position & Color Attri. & Overall\(\uparrow\)\\
    \midrule
    \multirow{6}{*}{guidance = 5} 
    & random & 0.9761 & 0.3540 & 0.3359 & 0.7301 & 0.0375 & 0.0579 & 0.4152 \\
    & 4 from 20 & 0.9797 & 0.4912 & 0.3235 & 0.7580 & 0.0500 & \textbf{0.1025} & 0.4508 \\
    & 4 from 50 & 0.9812 & \textbf{0.5177} & 0.3406 & 0.7367 & \textbf{0.0575} & 0.09 & 0.4540 \\
    & \(\tau=10\)  & \textbf{0.9891} & 0.3801 & \textbf{0.4110} & 0.7846 & 0.0388 & 0.0638 & 0.4445 \\
    & \(\tau=20\)  & 0.9782 & 0.4670 & 0.3750 & \textbf{0.7873} & 0.0438 & 0.0963 & \textbf{0.4594} \\
    & \(\tau=30\) & 0.9797 & 0.4988 & 0.3516 & 0.7686 & 0.0525 & \textbf{0.1025}  & 0.4589 \\  
    \midrule
    \multirow{6}{*}{guidance = 9} 
    & random & 0.9823 & 0.3725 & 0.3578 & 0.7713 & 0.0408 & 0.0688 & 0.4322 \\
    & 4 from 20 & \textbf{0.9954} & 0.5013 & 0.3328 & 0.8006 & 0.0525 & 0.1025 & 0.4642 \\
    & 4 from 50 & 0.9938 & \textbf{0.5455} & 0.3375 & 0.7952 & 0.055 & 0.11 & 0.4728 \\
    & \(\tau=10\)  & 0.9891 & 0.3662 & 0.3875 & 0.8006 & 0.0375 & 0.0975 & 0.4464 \\
    & \(\tau=20\) & 0.9906 & 0.50 & \textbf{0.4094} & \textbf{0.8112} & 0.0425 & 0.1175 & \textbf{0.4785} \\
    & \(\tau=30\) & \textbf{0.9954} & 0.5051 & 0.3563 & 0.8085 & \textbf{0.0575} & \textbf{0.1225} & 0.4742 \\  
    \bottomrule
  \end{tabular}
\end{table}

As shown in Table~\ref{sd1.5_gen}, filtering via ASD yields consistent improvements across all categories and guidance scales (\(\omega=5/9\)). Gains are more pronounced for SDv1.5 than SDXL (see Appendix~\ref{gen_sdxl}, Table~\ref{sdxl_gen}), probably due to the inherent sample concentration of SDXL in high-density regions, which limits the benefit of density-based filtering. Notably, our method achieves a \textbf{+46\%} gain on the two-object relationship prompts, highlighting the enhanced compositional reasoning. Furthermore, a threshold of \(\tau=10\) proves sufficient for high-quality selection while substantially reducing the compute.

\vspace{-0.1cm}
\paragraph{DPG-Bench Evaluation} 
\label{paper_dpg}
DPG-Bench provides 1,065 linguistically complex prompts to assess semantic alignment. We adopt the same filtering strategies as in GenEval. Table~\ref{sd1.5_dpg} shows consistent improvements across most categories, with the exception of the Global dimension. Again, SDv1.5 shows larger gains than SDXL (Appendix~\ref{dpg_sdxl}, Table~\ref{sdxl_dpg}), suggesting greater compatibility between our approach and diversity-promoting architectures. Importantly, early-stage filtering maintains strong performance across architectures and guidance settings.
 
\begin{table}[H]
  \centering
  \small
  \caption{The quantitative results on DPG-bench. Model: SDv1.5}
  \vspace{1em} 
  \label{sd1.5_dpg}
  \begin{tabular}{@{} l l *{5}{c} c @{}} 
    \toprule
    & Method& Global & Entity & Attribute & Relation & Other. & Overall\(\uparrow\)\\
    \midrule
    \multirow{5}{*}{guidance = 5} 
    & random & \textbf{82.02} & 72.16 & 71.63 & 80.1 & 55.4 & 62.45  \\
    & 4 from 20 & 79.03 & 73.87 & 72.82 & 82.0 & \textbf{59} & 64.48  \\
    & 4 from 50 & 78.72 & \textbf{74.51} & \textbf{73.11} & \textbf{82.61}  & 58 & \textbf{65.15}  \\
    & \(\tau=10\)  & 80.09 & 72.06 & 72.67 & 79.12 & 56.4 & 62.8  \\
    & \(\tau=20\)  & 81.31 &73.22 & 72.63 & 80.51 & 57.8 & 64.14  \\
    & \(\tau=30\) & 79.94 & 73.27 &72.65 &81.24 & 58.4 & 64.19  \\   
    \midrule
    \multirow{5}{*}{guidance = 9} 
    & random & \textbf{82.88} & 71.71 & 72.54 & 78.92 & 55.2 & 62.64  \\
    & 4 from 20 & 81.31 & 73.60 & 74.3 & 80.24 & 61 & 64.87  \\
    & 4 from 50 & 82.07 & 73.3 & \textbf{74.42} &\textbf{81.37}  & 61.6 & 64.99  \\
    & \(\tau=10\)  & 81.61 & 71.61 & 71.92 & 78.4 & 57.6 & 62.74  \\
    & \(\tau=20\)  & 82.53 & 73.53 & 73.04 & 78.98 & \textbf{62.4} & 65.33  \\
    & \(\tau=30\) & 81.01 & \textbf{74.02} & 73.82 & 79.82 & 61.85 & \textbf{65.58}  \\  
    \bottomrule
  \end{tabular}
\end{table}
\vspace{-0.2cm}

\subsection{Visual text generation}

We demonstrate the applicability of our method to the state-of-the-art Flux diffusion architecture on a set of challenging visual text rendering tasks, which typically yield low success rates in standard generation. For each prompt, we generate 100 images and apply our filtering strategy to select top and bottom ranked outputs for qualitative comparison. As illustrated in Fig.\ref{flux}, high-ASD samples consistently render complete and accurate text, while low-ASD ones often fail to generate legible content. Detailed settings and additional examples are provided in Appendix\ref{flux_app}.

\vspace{-0.3cm}
\section{Discussion and Conclusion}
\vspace{-0.1cm}
\paragraph{Discussion}
\label{dis_limit}
The primary limitation of our work lies in the theoretical understanding of the filtering mechanism. While we observe a strong empirical correlation between accumulated score differences and sample density, the theoretical foundations remain underexplored. What structural conditions guarantee this correlation? When does early filtering preserve semantic fidelity? Addressing these questions may uncover deeper principles behind guidance effectiveness.

Looking ahead, our approach opens several promising avenues. First, incorporating directional signals (e.g. cosine similarity) could yield a more nuanced measure of alignment and further enhance filtering precision. Second, the impact of different ODE solvers (e.g., Euler vs. Heun) and noise schedules on the effectiveness of ASD-based filtering remains largely unexplored; systematically studying these factors could further improve robustness and generalization. Third, extending CFG-Rejection to other generative domains, such as 3D synthesis or audio generation, holds the potential to generalize early-stage filtering across modalities. Lastly, given its orthogonality to reward-based alignment, our method could be combined with existing inference-time optimization techniques to achieve both efficiency and controllability in generation.

\vspace{-0.1cm}
\paragraph{Conclusion}
To the best of our knowledge, this work is the first to formally identify the positive correlation between accumulated score differences and sampling density regions. Drawing insights from the 2D fractal example, we propose the CFG-Rejection method that enhances text alignment and sample quality without external reward models. We validate these observations on the ImageNet dataset through density curves using LOF and AvgkNN. Extensive experiments with EDM2-S, SDXL and SD1.5, coupled with human preference ratings and challenging benchmarks, demonstrate the effectiveness and compatibility of our method. Qualitative comparisons on visual text application underscore the rich, yet underutilized, signals for quality control present in the early stages of latent spaces. We believe CFG-Rejection offers a “free lunch” in sample quality control and can serve as a default enhancement in diffusion-based generation pipelines, paving the way for future methods to more deeply leverage intrinsic signals during inference.

\bibliographystyle{unsrt}
\bibliography{reference} 

\newpage
\appendix

\section{Illustration on toy example}
\label{toy_app}
In this section, we detail the construction of the 2D toy example and the procedure to quantify the geometric relationship between accumulated score differences (ASD) and sample density. We further verify that this relationship remains stable across varying guidance strengths \(\omega\).

We follow the implementation of~\cite{karras2024guiding} for both dataset construction and diffusion model training. Each class \(\mathbf{c}\) is modeled as a Gaussian mixture \(\mathcal{M}_{\mathbf{c}} = \big( \{\phi_i\}, \{\mathbf{\mu}_i\}, \{\mathbf{\Sigma}_i\} \big)\), where \(\phi_i\), \(\mathbf{\mu}_i\) and \(\mathbf{\Sigma}_i\) denote the mixture weight, mean, and \(2{\times}2\) covariance matrix of component \(i\), respectively. This formulation allows for closed-form computation of ground-truth scores and densities:

\vspace{-0.1cm}
\begin{align}
  p_\text{data}(x | c) &\,=\, \sum_{i \in \mathcal{M}_{\mathbf{c}}} \!\phi_i \,\mathcal{N}(x; \mathbf{\mu}_i, \mathbf{\Sigma}_i)
  \text{,}\\
  \mathcal{N}(x; \mathbf{\mu}, \mathbf{\Sigma}) &\,=\, \frac{1}{\sqrt{(2 \pi)^2 \det(\mathbf{\Sigma})}} \exp \bigg( \!-\frac{1}{2} (x - \mathbf{\mu})^\top \mathbf{\Sigma}^{-1} (x - \mathbf{\mu}) \bigg)
  \text{.}
\end{align}

To construct a tree-like structure, we follow the open-source implementation provided in~\cite{karras2024guiding}. Specifically, we begin with a main branch and recursively subdivide it into finer branches. Each branch is represented by 8 anisotropic Gaussian components, and the subdivision is repeated 6 times. All settings, including initialization, branching rules, and Gaussian configurations, strictly follow the open code at \url{https://github.com/NVlabs/edm2} given by~\cite{karras2024guiding}.

To track the accumulated score differences (ASD), we take advantage of the fact that this is a 2D toy example, which allows us to compute the score difference in closed form using the \(\ell_2\)-norm. Specifically, we directly evaluate \(\mathcal{G}_t(c) = \| S_\theta(\mathbf{x}_t;\sigma_t,\mathbf{c}) - S_\theta(\mathbf{x}_t;\sigma_t,\emptyset) \|_2\) following Equation~\ref{nv_gap} at each timestep and accumulate it across the entire denoising trajectory. Note that the model used here differs from our pre-defined \(S_\theta(\mathbf{x}_t;\sigma_t,\mathbf{c})\); thus, we multiply the raw model outputs by \(\sigma_t\) to obtain the correct scaled scores for computing ASD. All results are based on 32-step inference using the default Heun’s second-order solver.

\begin{figure}[H]
\centering  
\subfigure[Samples with CFG]{   
\begin{minipage}{0.45\textwidth}
\centering   
\includegraphics[width=0.9\linewidth]{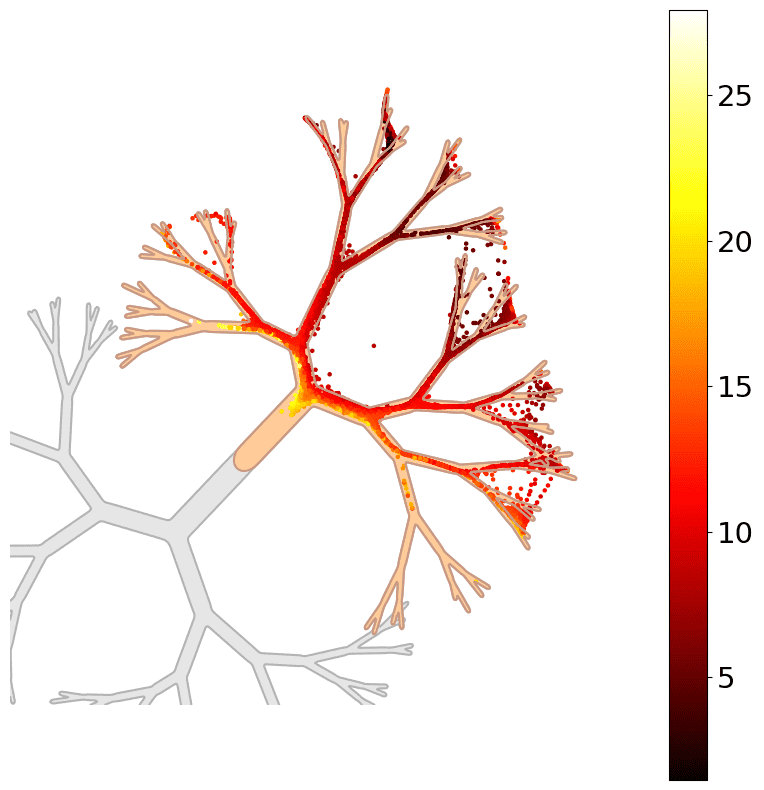}  
\label{toy_images app1}
\end{minipage}
}
\subfigure[Positive relationship]{
\begin{minipage}{0.45\textwidth}
\centering   
\includegraphics[width=1.1\linewidth]{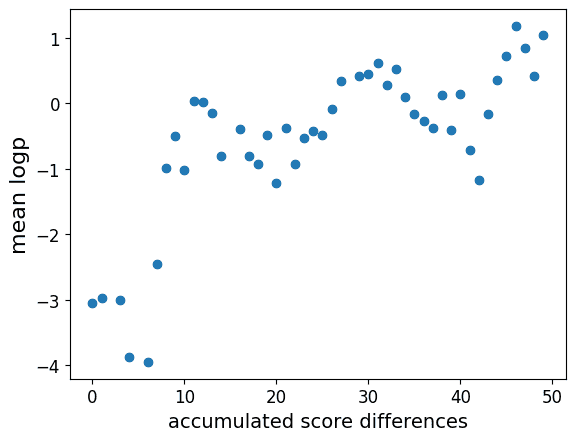}
\label{toy_relation app1}
\end{minipage}
}
\caption{Illustration on toy example with \(\omega=2.5\).}
\label{toy examples app1}
\end{figure}
\begin{figure}[H]
\centering  
\subfigure[Samples with CFG]{   
\begin{minipage}{0.45\textwidth}
\centering   
\includegraphics[width=0.9\linewidth]{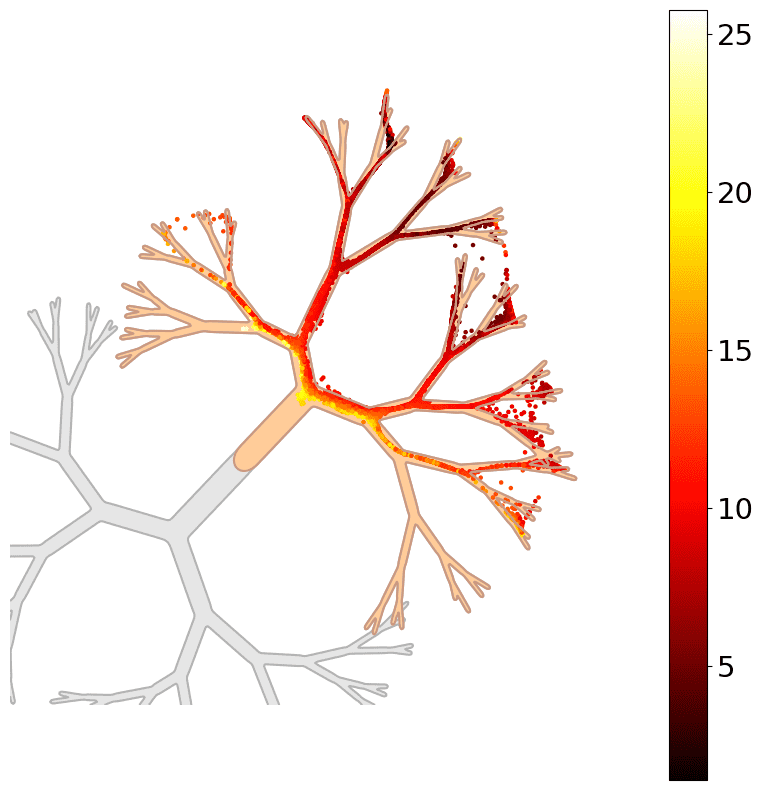}  
\label{toy_images app2}
\end{minipage}
}
\subfigure[Positive relationship]{
\begin{minipage}{0.45\textwidth}
\centering   
\includegraphics[width=1.1\linewidth]{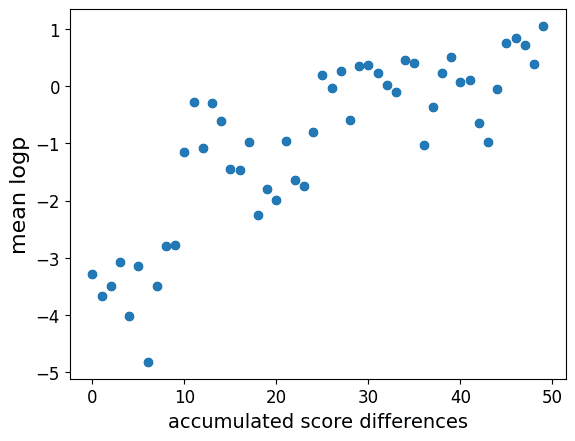}
\label{toy_relation app2}
\end{minipage}
}
\caption{Illustration on toy example with \(\omega=3\).}
\label{toy examples app2}
\end{figure}
\begin{figure}[H]
\centering  
\subfigure[Samples with CFG]{   
\begin{minipage}{0.45\textwidth}
\centering   
\includegraphics[width=0.9\linewidth]{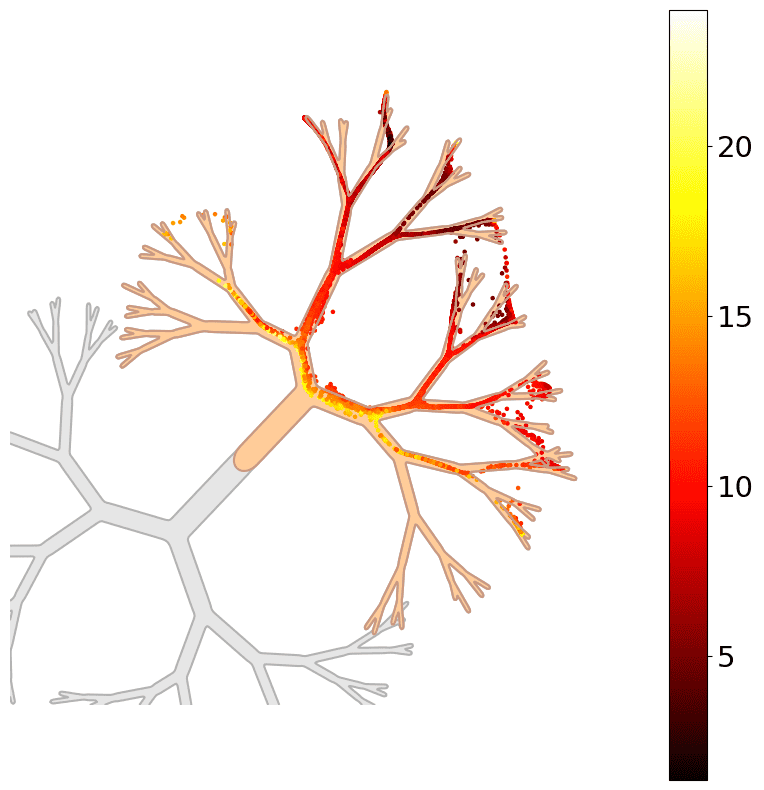}  
\label{toy_images app3}
\end{minipage}
}
\subfigure[Positive relationship]{
\begin{minipage}{0.45\textwidth}
\centering   
\includegraphics[width=1.1\linewidth]{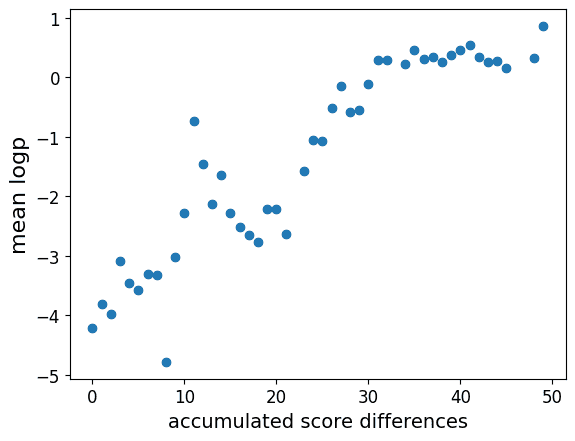}
\label{toy_relation app3}
\end{minipage}
}
\caption{Illustration on toy example with \(\omega=3.5\).}
\label{toy examples app3}
\end{figure}

\vspace{-0.1cm}
Figures~\ref{toy examples app1}–\ref{toy examples app3} present observations under various \(\omega\). Empirically, we find the relationship between ASD and sample density to be largely invariant to the guidance strength. Samples with high ASD values (\(\mathcal{E}_{T}(c) > \gamma\)) predominantly concentrate within the high-density trunk regions, while samples with smaller ASD (\(\mathcal{E}_{T}(c) < \gamma\)) primarily populate the low-density branches, with extreme cases (\(\mathcal{E}_{T}(c)\approx0\)) corresponding to degenerate outputs that violate label constraints.

For a quantitative analysis, we divide ASD into 50 bins and plot the mean log-probability density of samples in each bin. All figures reveal a consistent log-linear relationship between local density and ASD across different \(\omega\), supporting the use of ASD as a self-supervised signal for filtering.

\section{More results on ImageNet}
\label{Img_app}
We adopt the pre-trained EDM2-S model provided by~\cite{karras2024guiding}, using default inference settings with 32 denoising steps and Heun’s second-order sampler. The classifier-free guidance weight is set to \(\omega=1.4\). ImageNet samples are generated at \(512{\times}512\), with denoising carried out in the latent space of shape \(4{\times}64{\times}64\).

To compute our metric \(\mathcal{G}_t(c)\), we calculate the matrix two-norm of the score difference at each step and average it over the channel dimension. Unlike the 2D toy setting, there is no need to scale the output by \(\sigma_t\) here. All experiments are conducted on a single RTX 4090 GPU.

The following subsections provide additional results: Subsection~\ref{Ima_density} presents extended density visualizations, Subsection~\ref{Ima_qua} includes more qualitative comparisons, and Section~\ref{Ima_baseline} reports quantitative results across baseline methods.

\subsection{Manifold density analysis}
\label{Ima_density}
\begin{figure}[htbp]
\centering
\subfigure[Crib]{  
\begin{minipage}{0.31\textwidth}
\centering
\includegraphics[width=\linewidth]{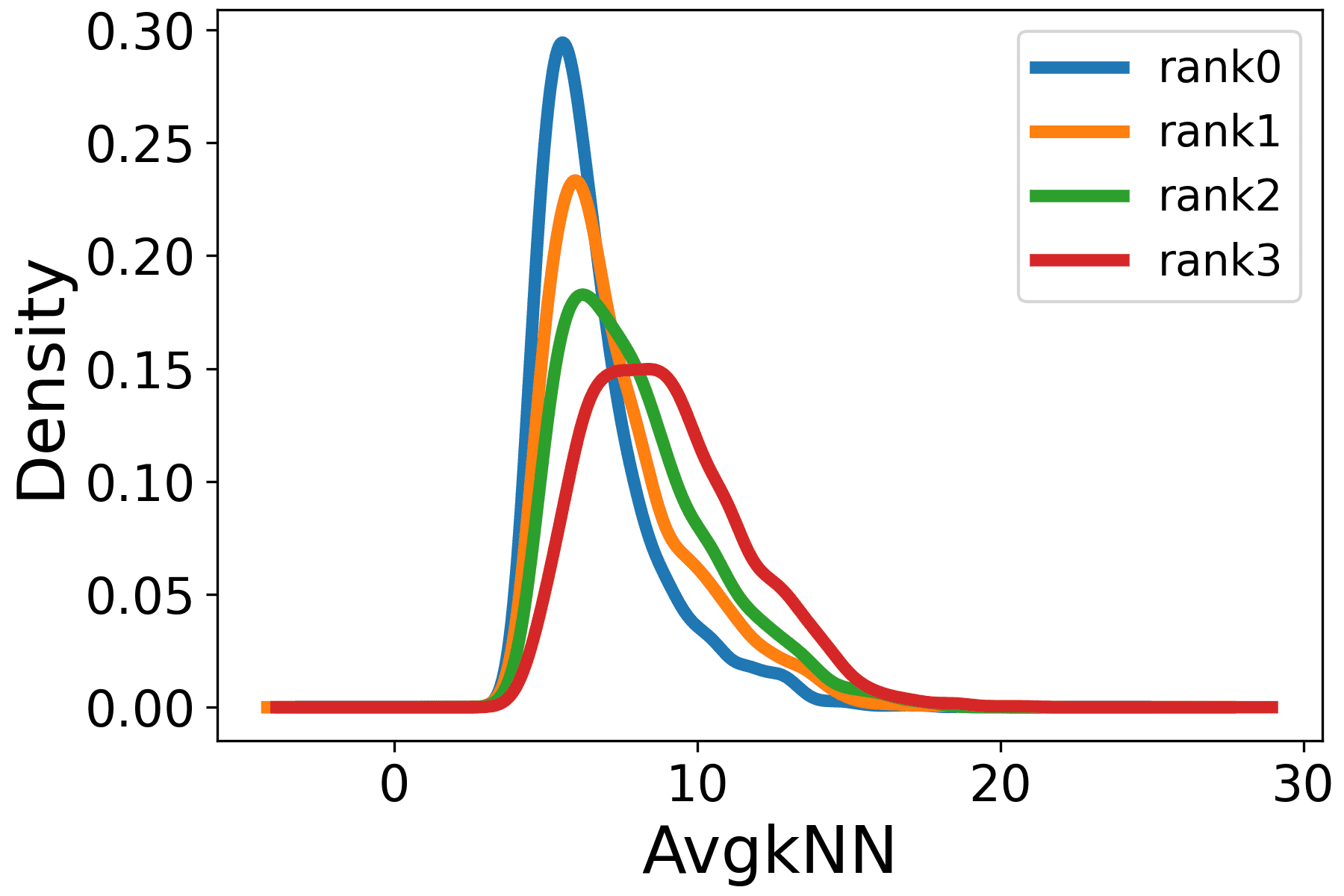}
\includegraphics[width=\linewidth]{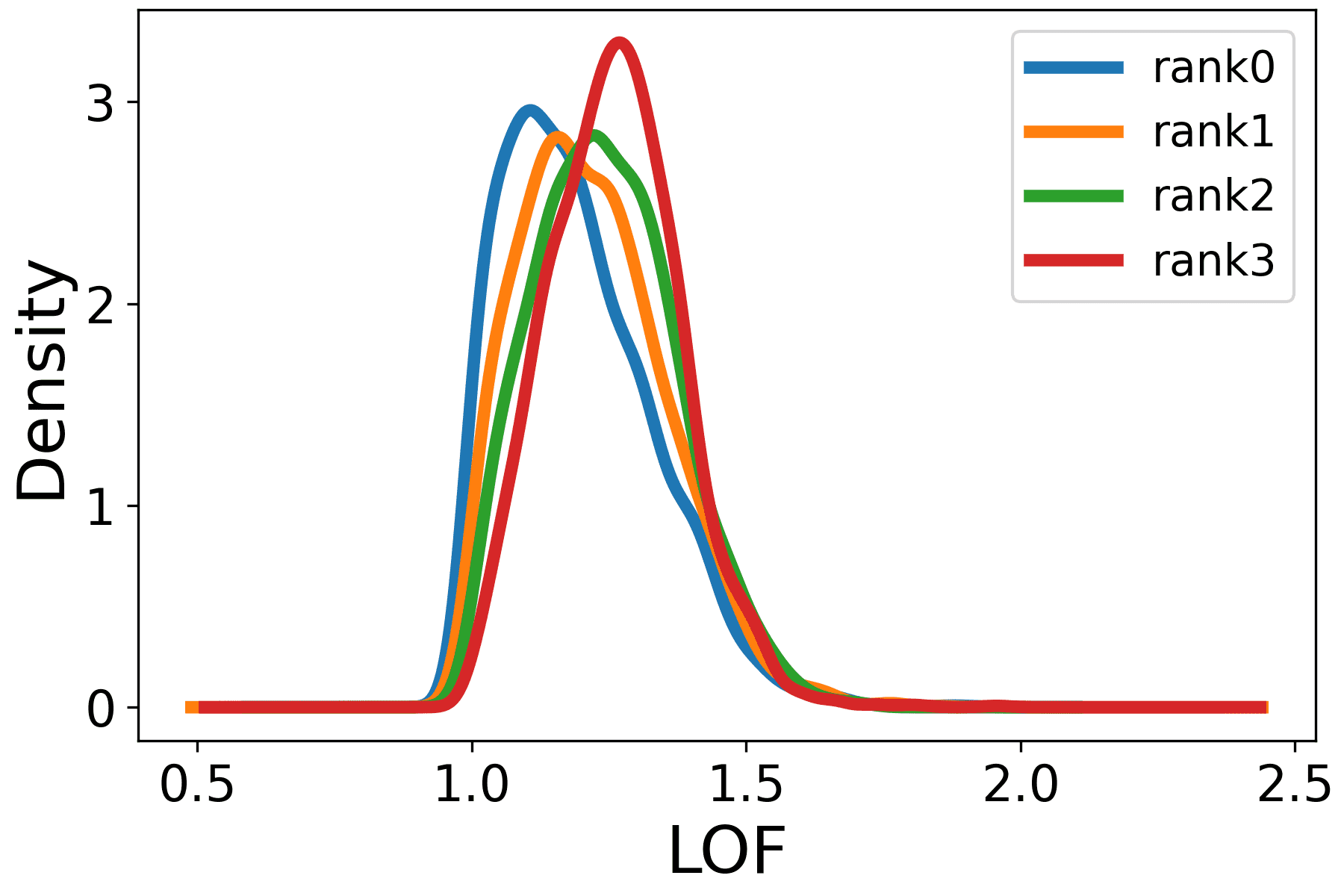} 
\label{520_density}
\end{minipage}
}
\subfigure[Fountain]{  
\begin{minipage}{0.31\textwidth}
\centering
\includegraphics[width=\linewidth]{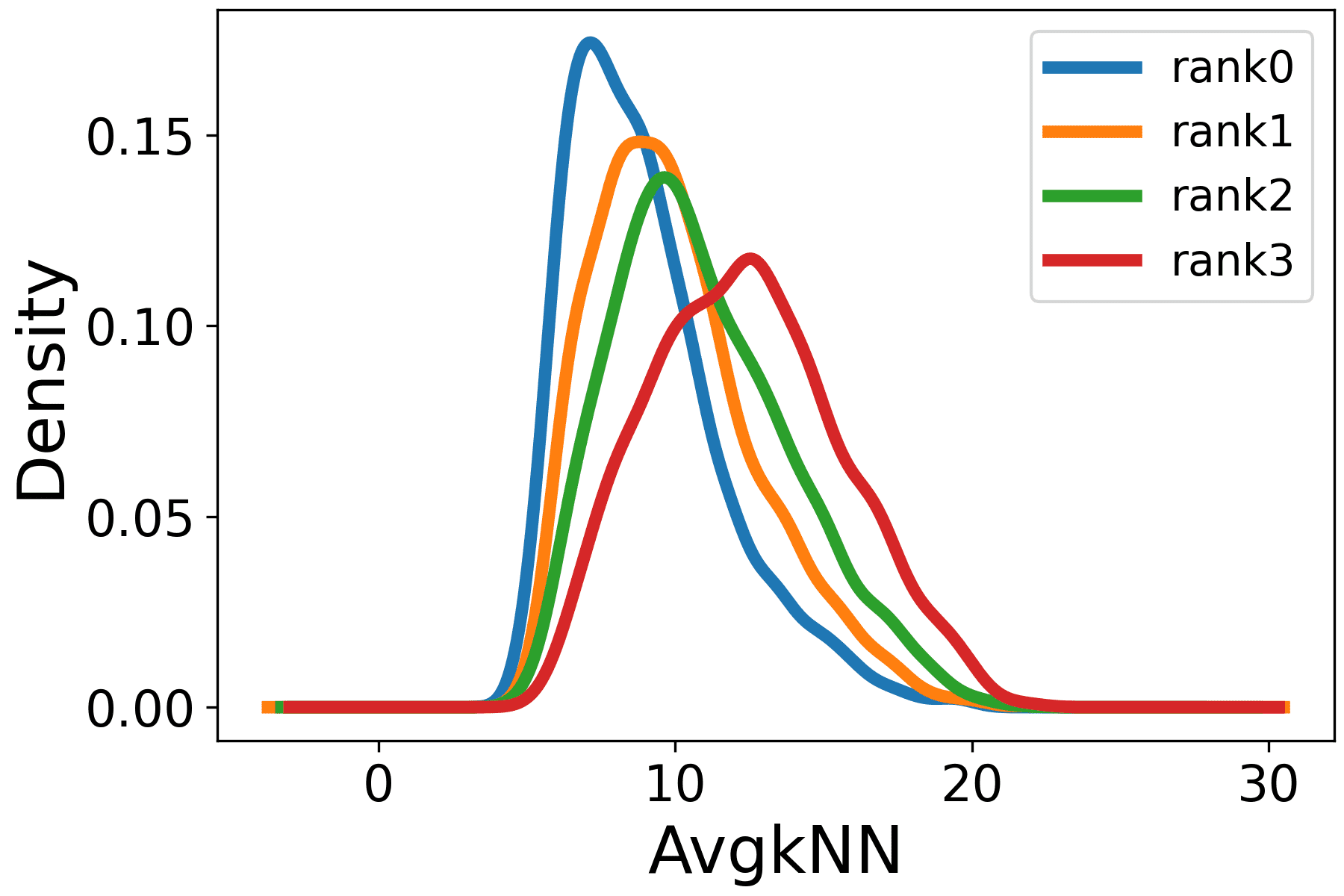}
\includegraphics[width=\linewidth]{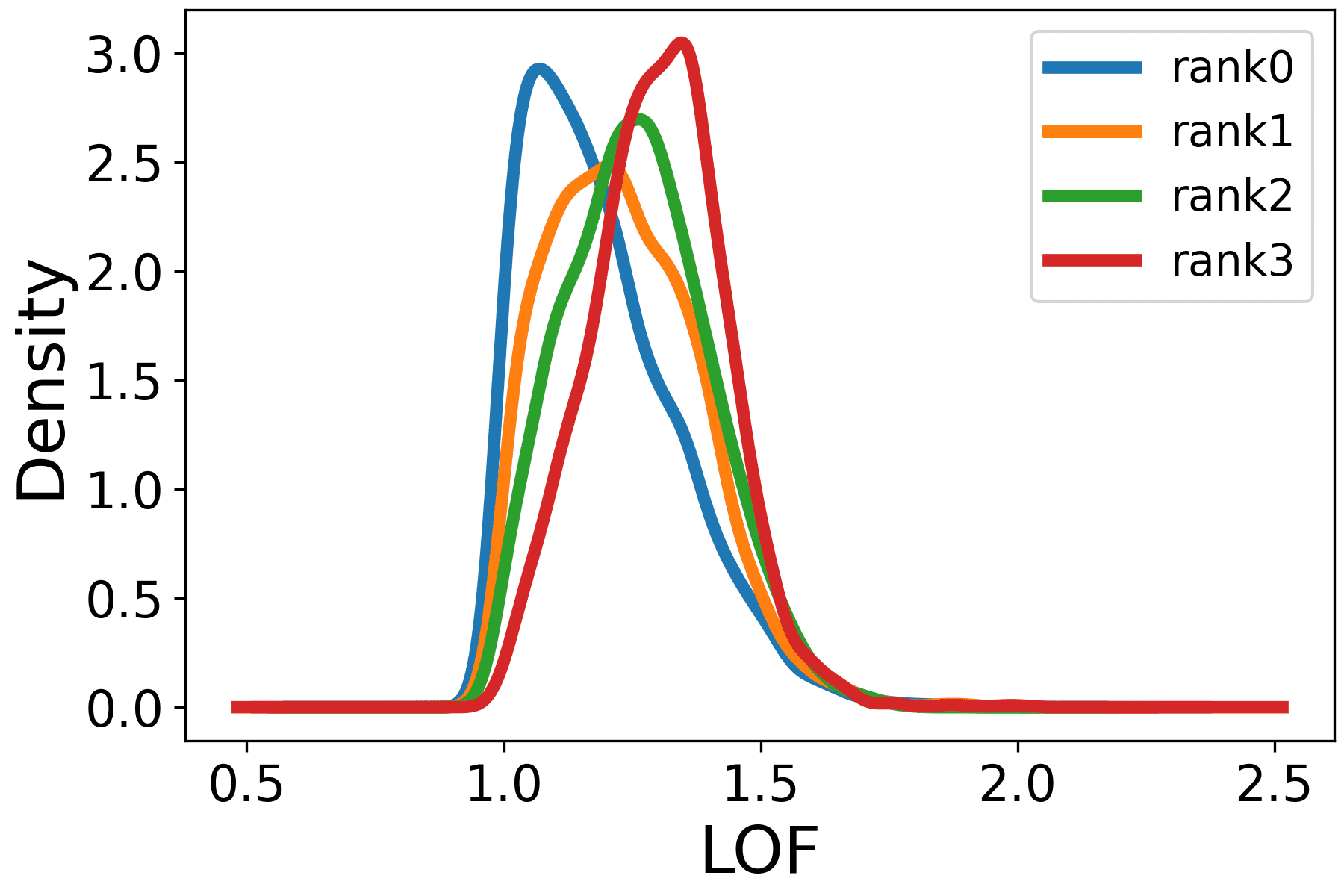}
\label{562_density}
\end{minipage}
}
\subfigure[Parachute]{  
\begin{minipage}{0.31\textwidth}
\centering
\includegraphics[width=\linewidth]{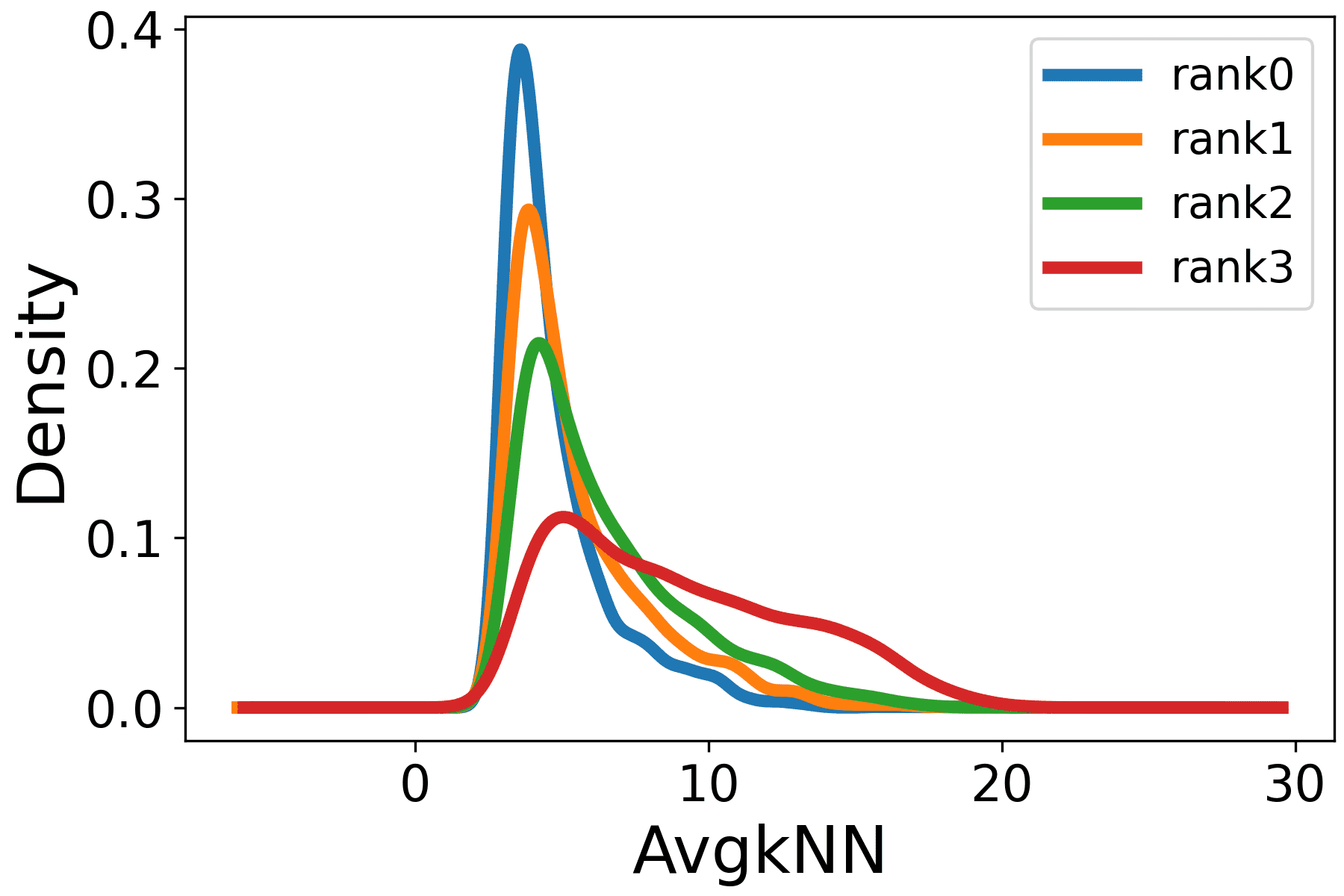}
\includegraphics[width=\linewidth]{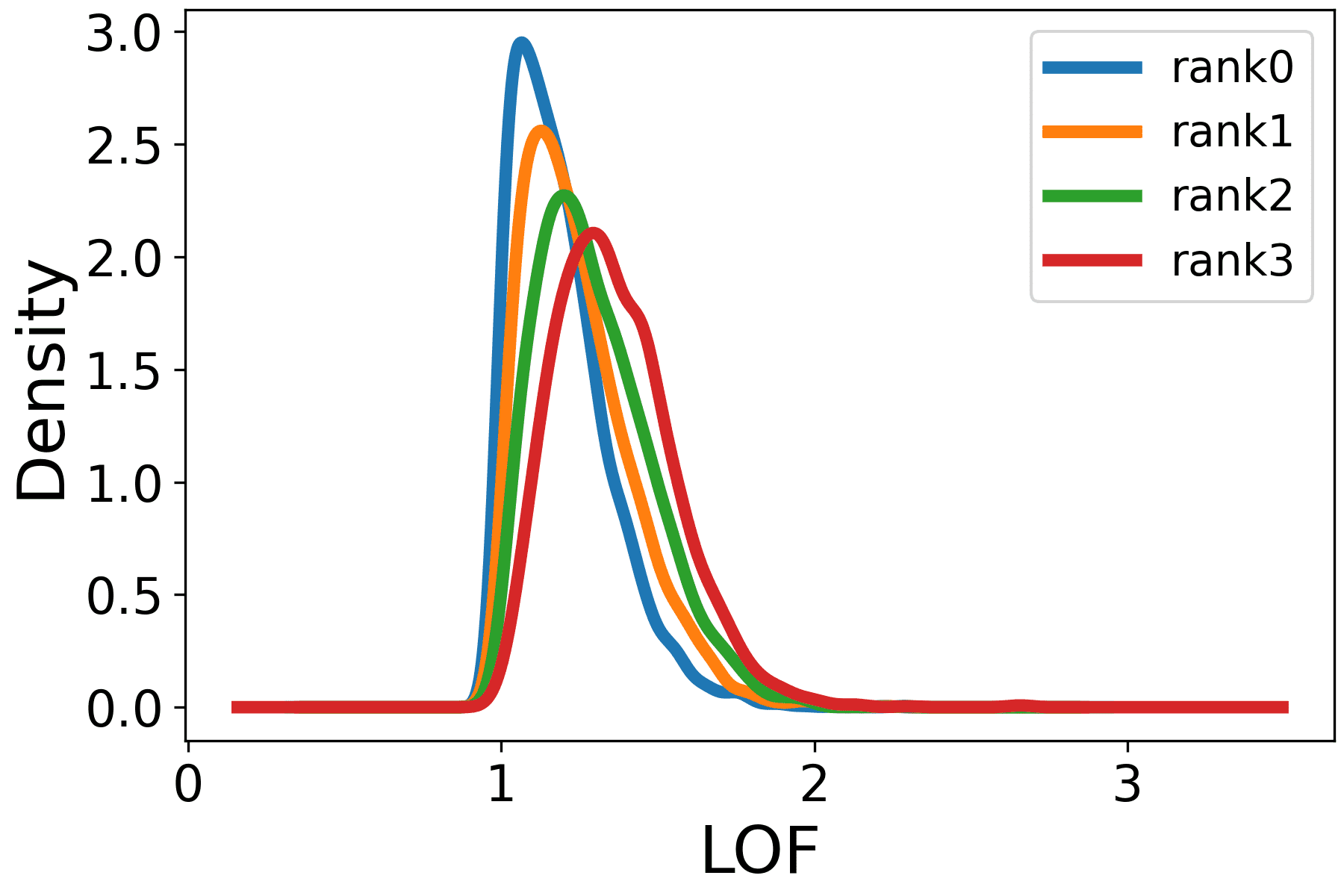}
\label{701_density}
\end{minipage}
}
\caption{Density estimation curves for generated samples from the classes Crib, Fountain, and Parachute.} 
\label{ImageNet_density app2}
\end{figure}

\begin{figure}[htbp]
\centering
\subfigure[Bulbul]{  
\begin{minipage}{0.31\textwidth}
\centering
\includegraphics[width=\linewidth]{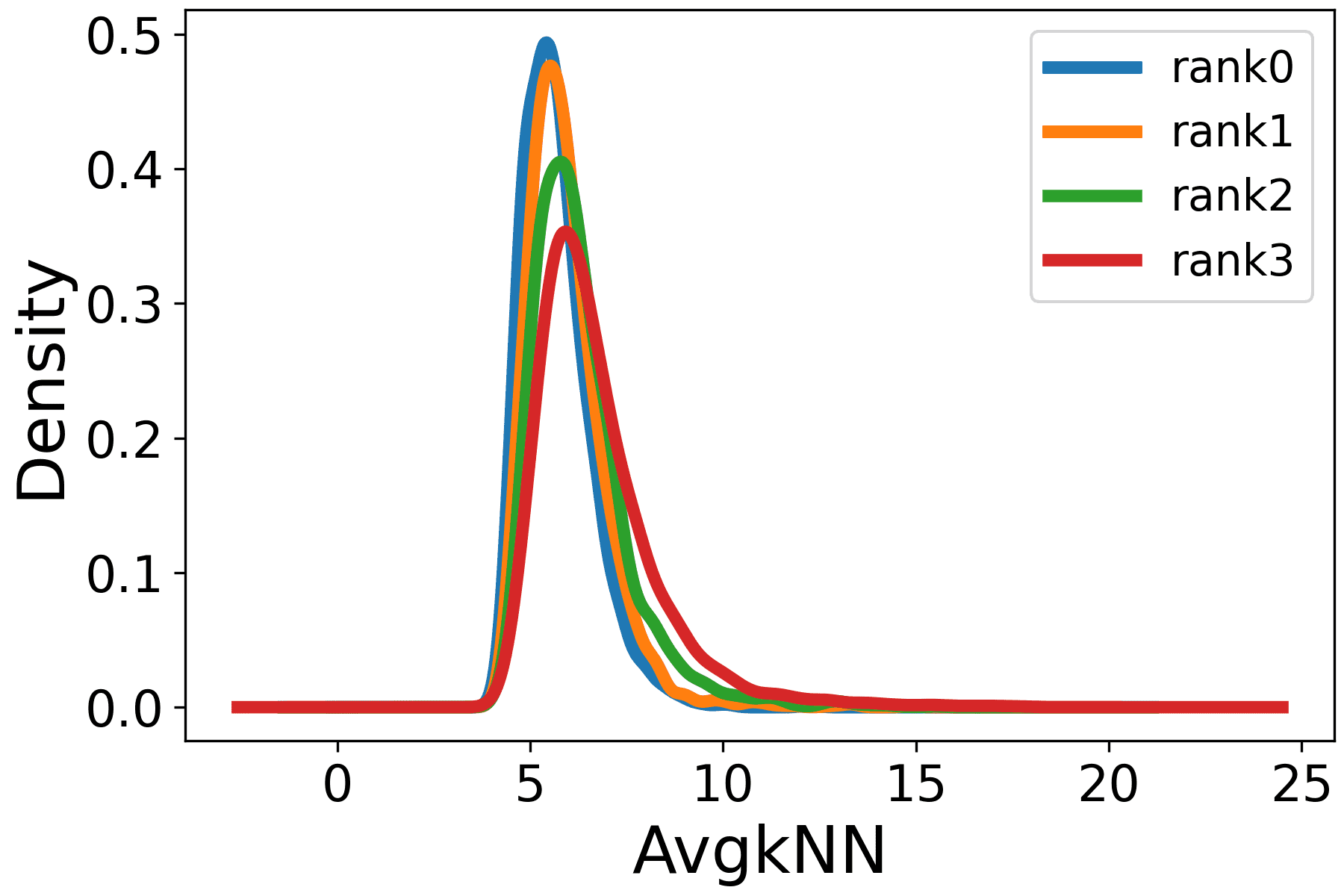} 
\includegraphics[width=\linewidth]{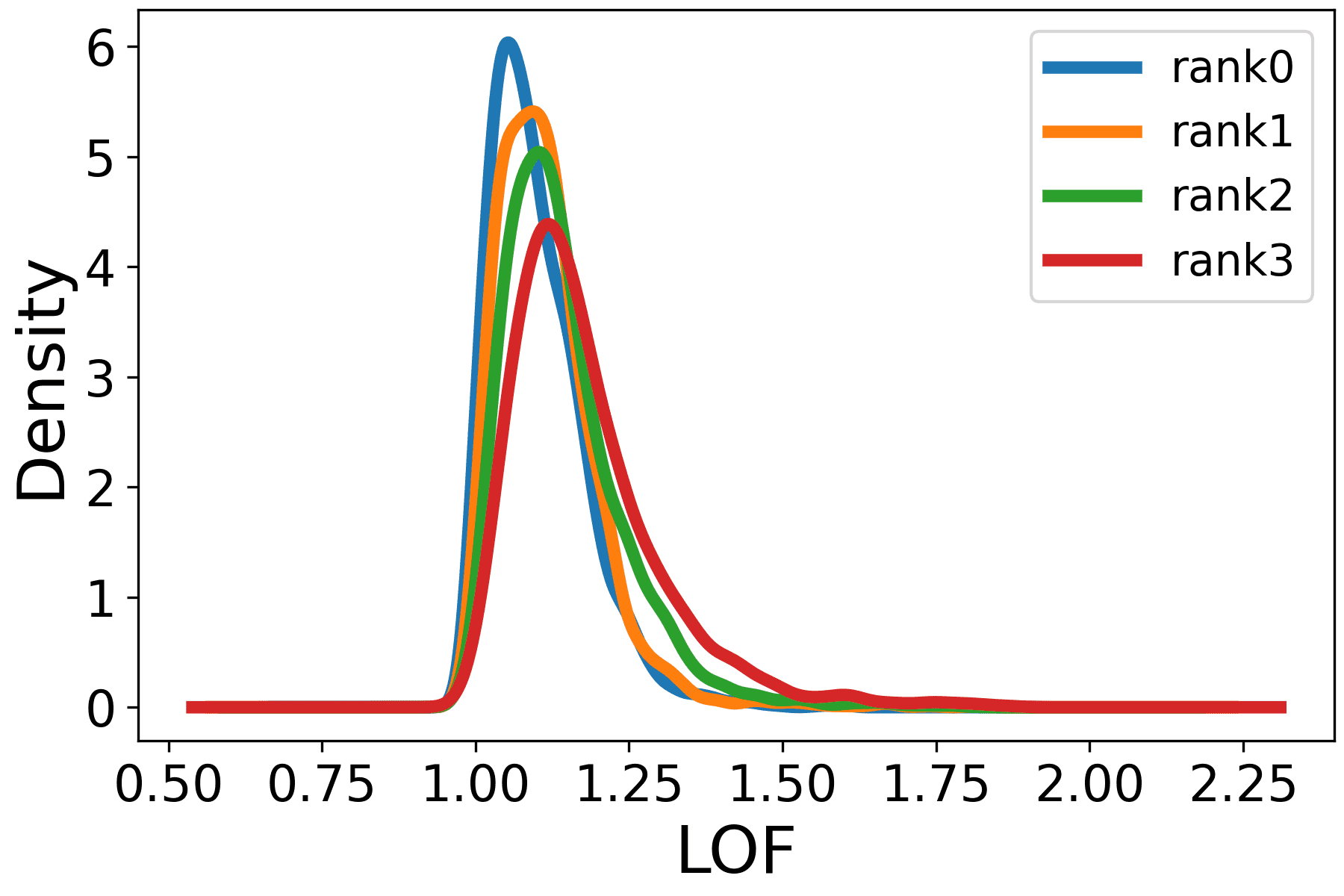} 
\label{16_density}
\end{minipage}
}
\subfigure[Goldfish]{  
\begin{minipage}{0.31\textwidth}
\centering
\includegraphics[width=\linewidth]{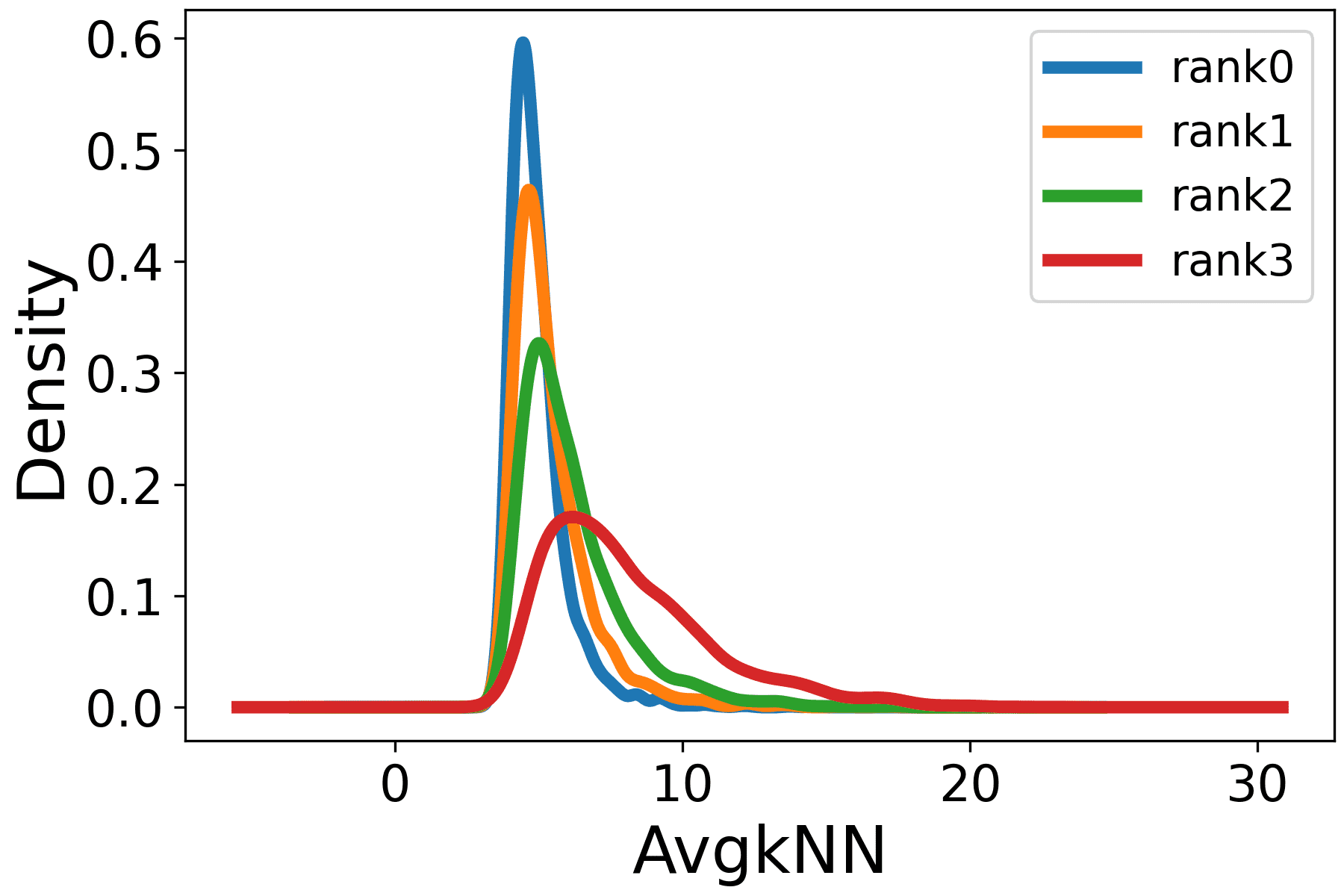}
\includegraphics[width=\linewidth]{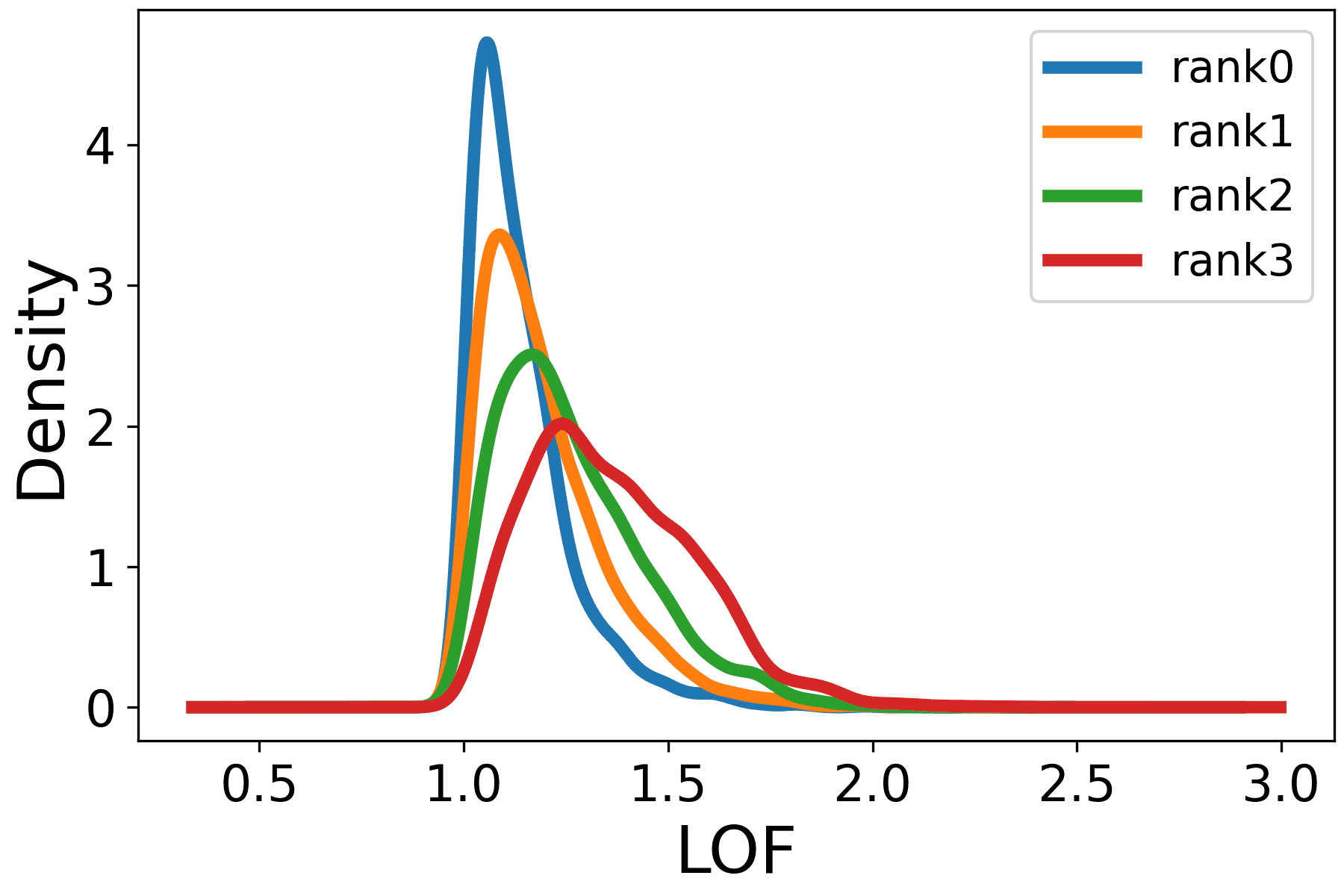}
\label{1_density}
\end{minipage}
}
\subfigure[Viaduct]{  
\begin{minipage}{0.31\textwidth}
\centering
\includegraphics[width=\linewidth]{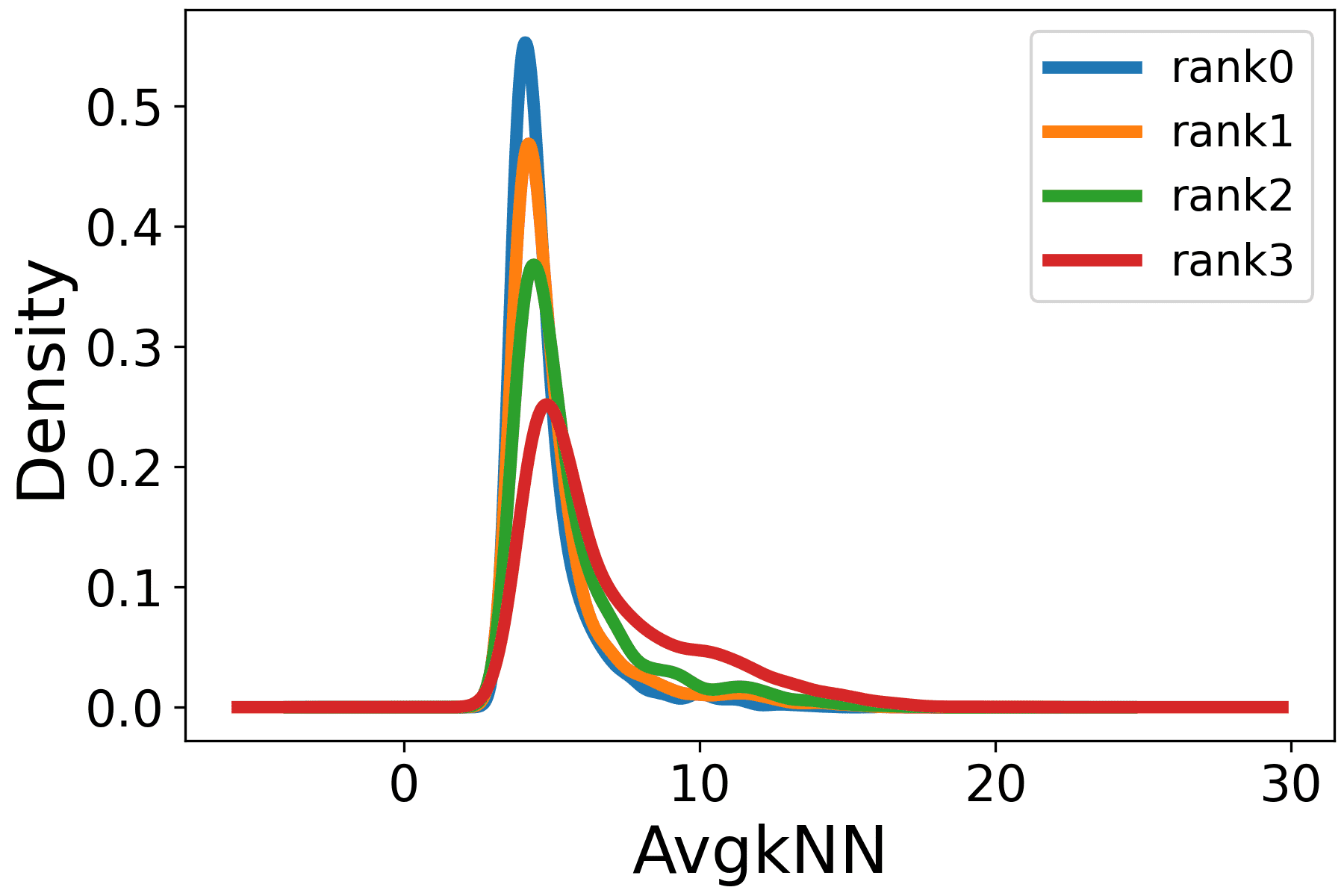}
\includegraphics[width=\linewidth]{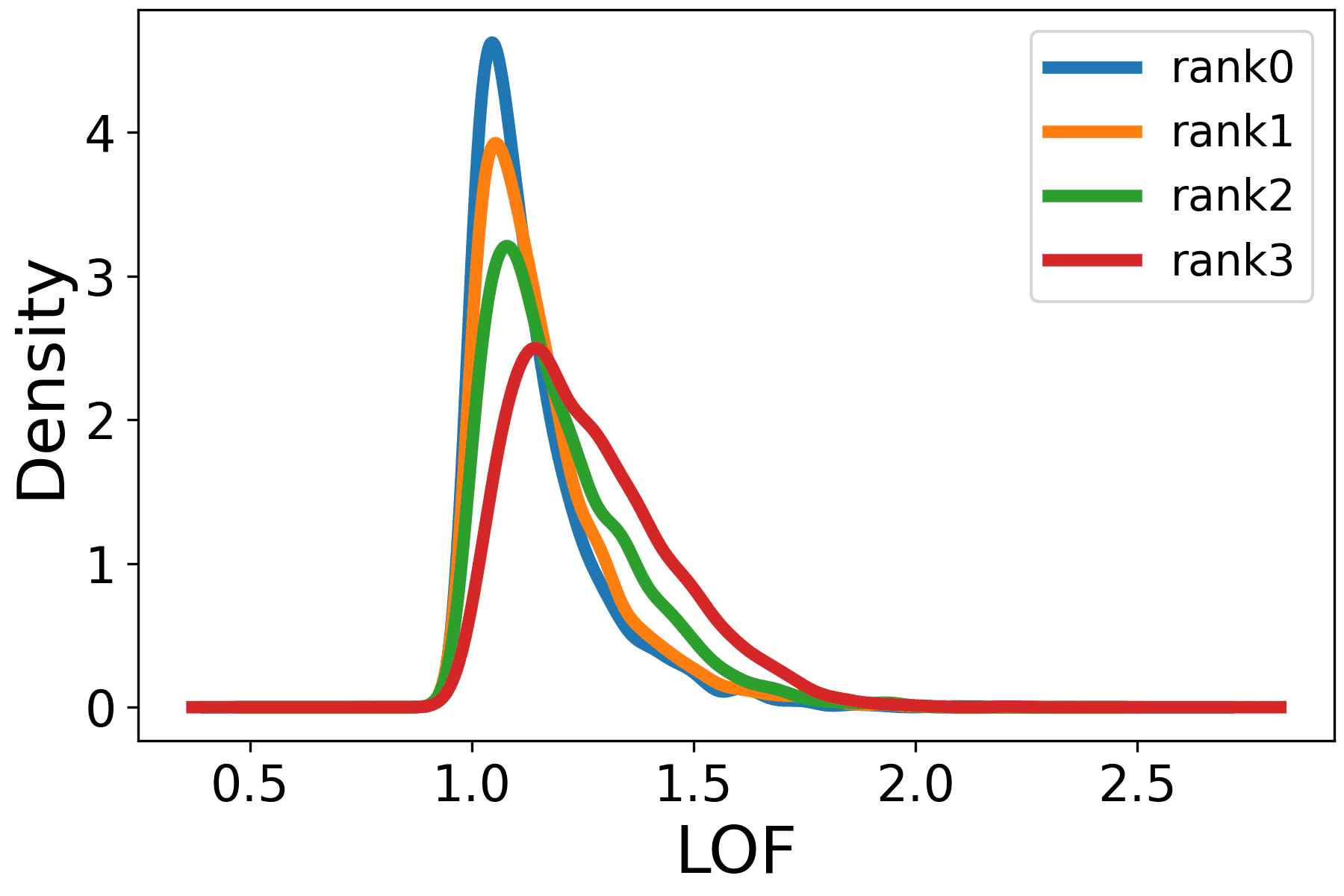}
\label{888_density}
\end{minipage}
}
\caption{Density estimation curves for generated samples from the classes Bulbul, Goldfish, and Viaduct.} 
\label{ImageNet_density app3}
\end{figure}

\begin{figure}[htbp]
\centering
\subfigure[Water tower]{  
\begin{minipage}{0.31\textwidth}
\centering
\includegraphics[width=\linewidth]{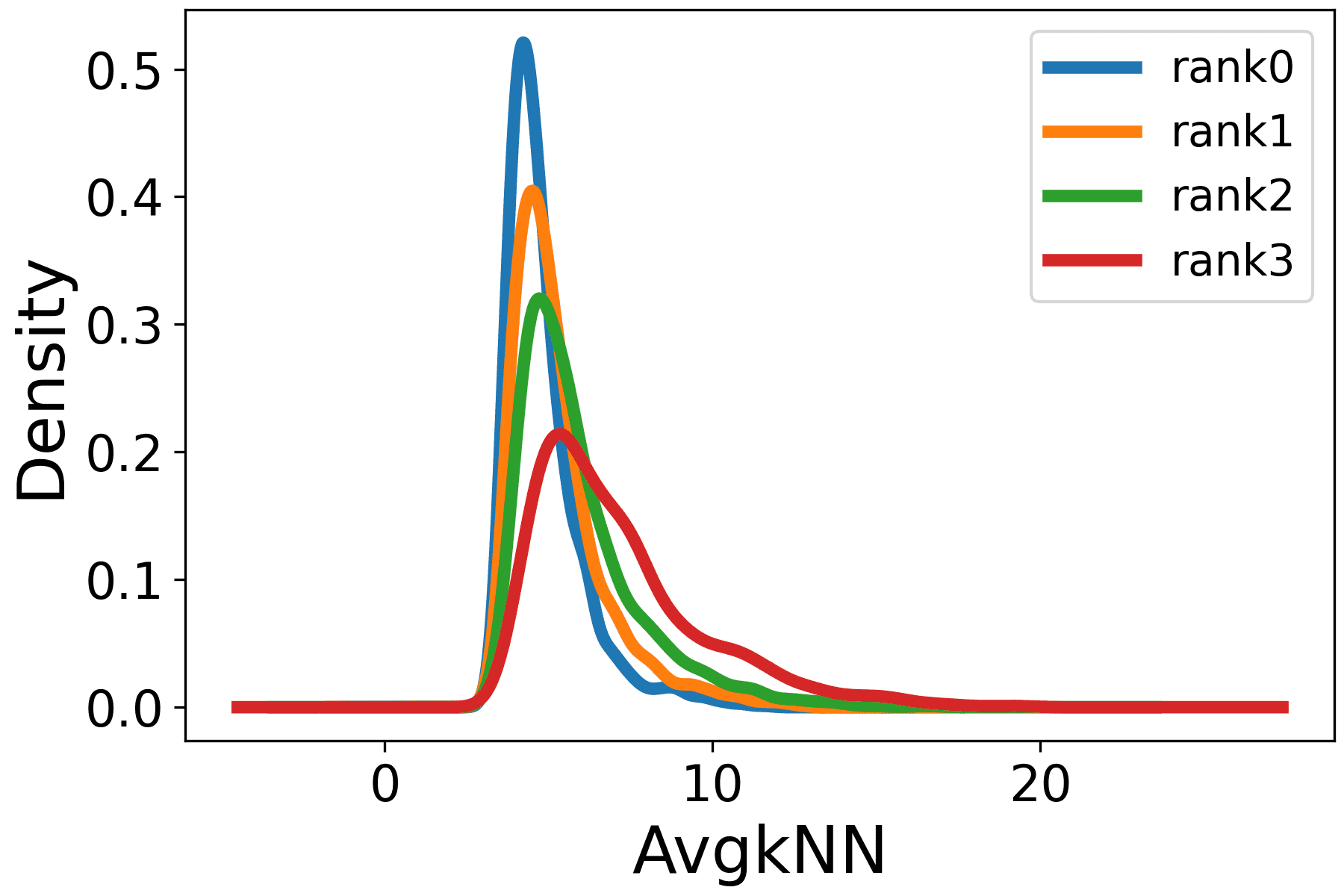} 
\includegraphics[width=\linewidth]{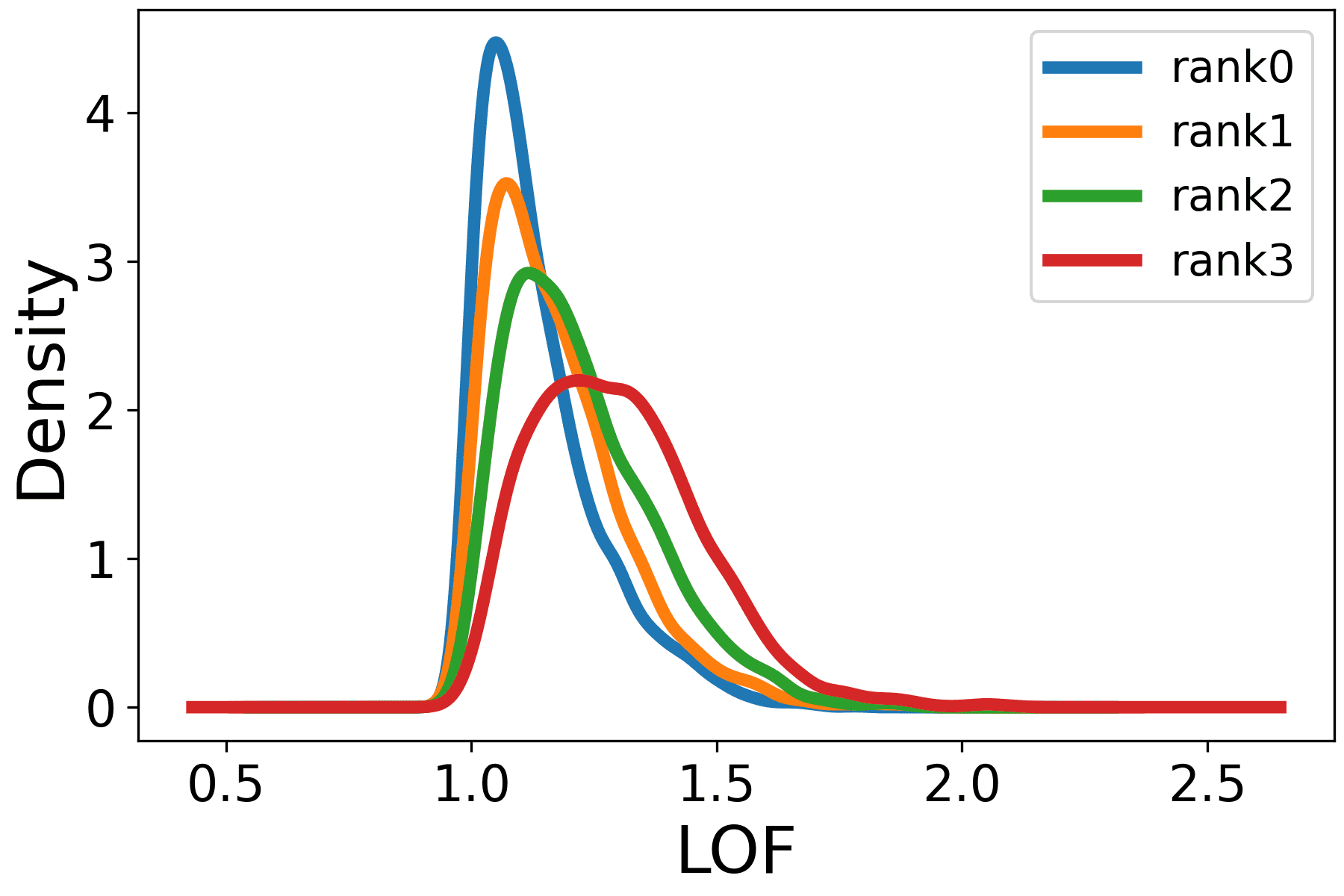} 
\label{900_density}
\end{minipage}
}
\subfigure[Snow leopard]{  
\begin{minipage}{0.31\textwidth}
\centering
\includegraphics[width=\linewidth]{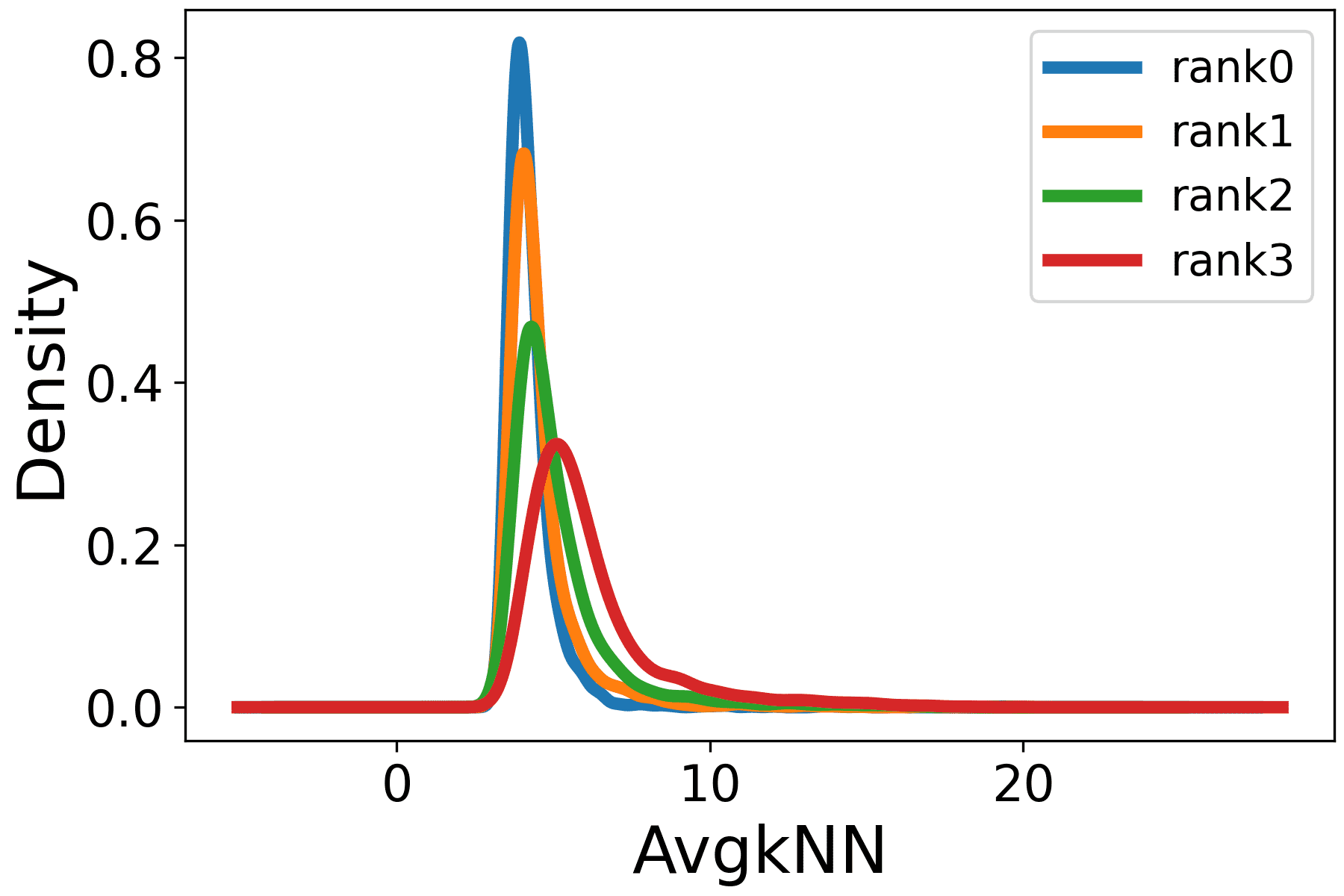}
\includegraphics[width=\linewidth]{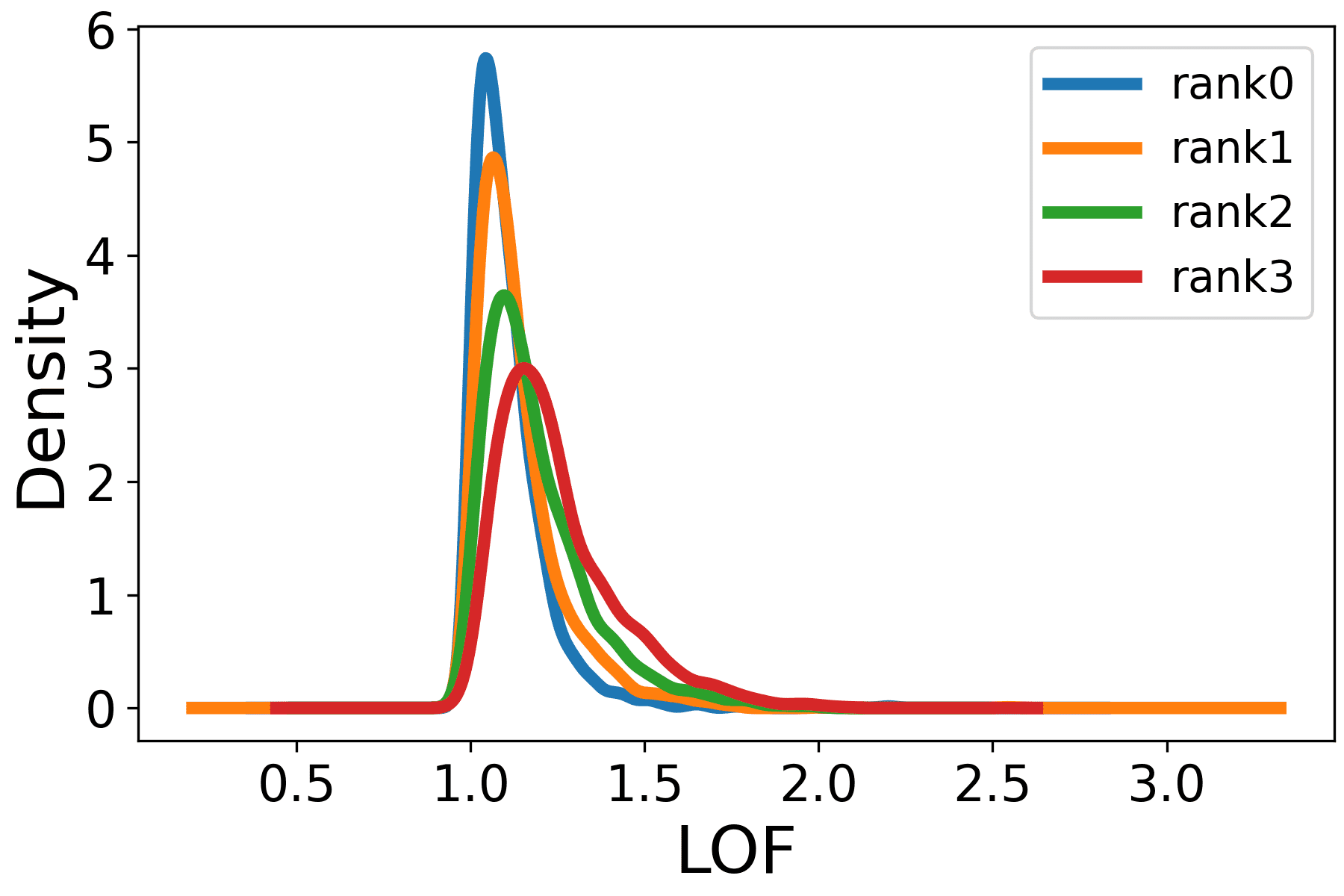}
\label{289_density}
\end{minipage}
}
\subfigure[Head cabbage]{  
\begin{minipage}{0.31\textwidth}
\centering
\includegraphics[width=\linewidth]{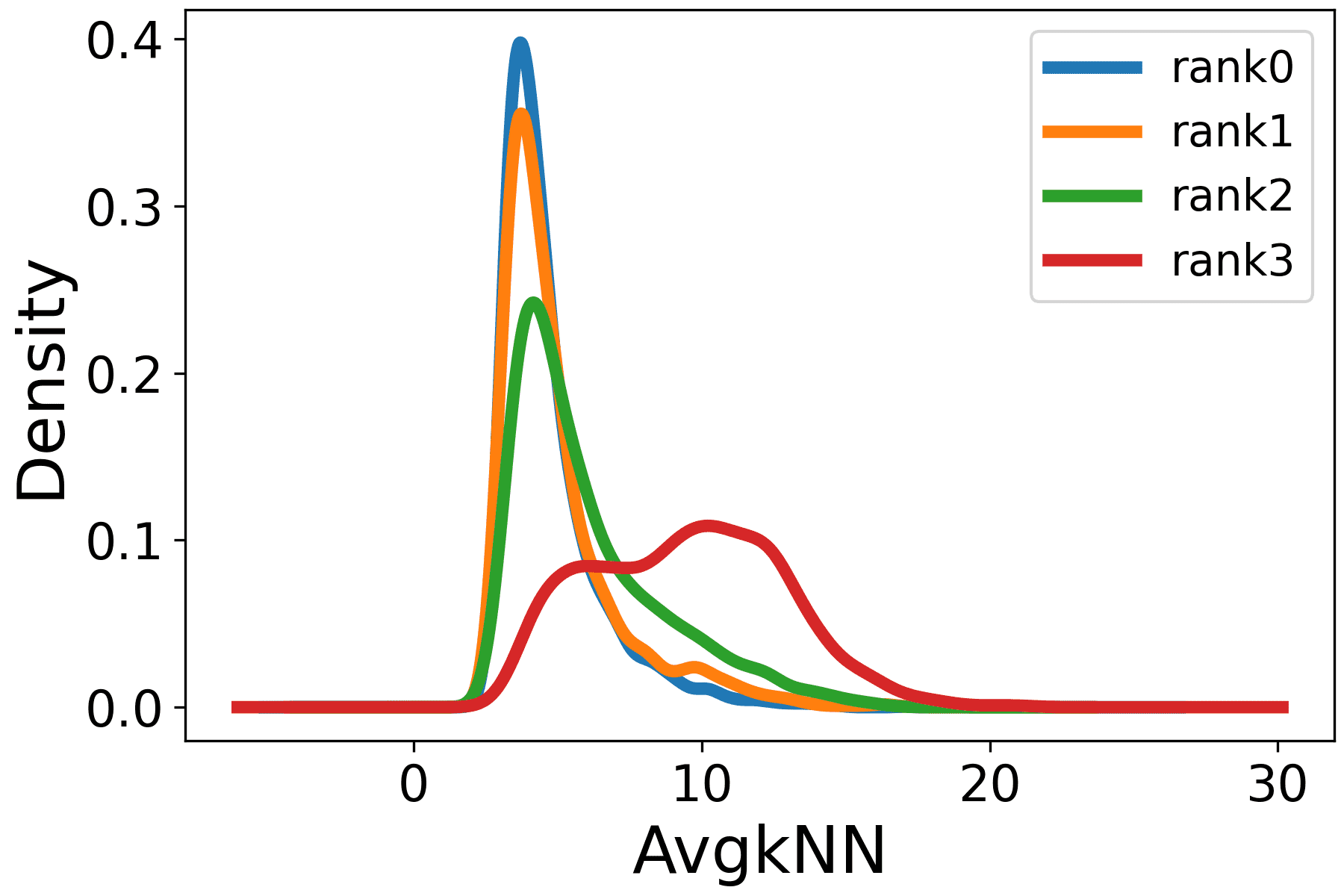}
\includegraphics[width=\linewidth]{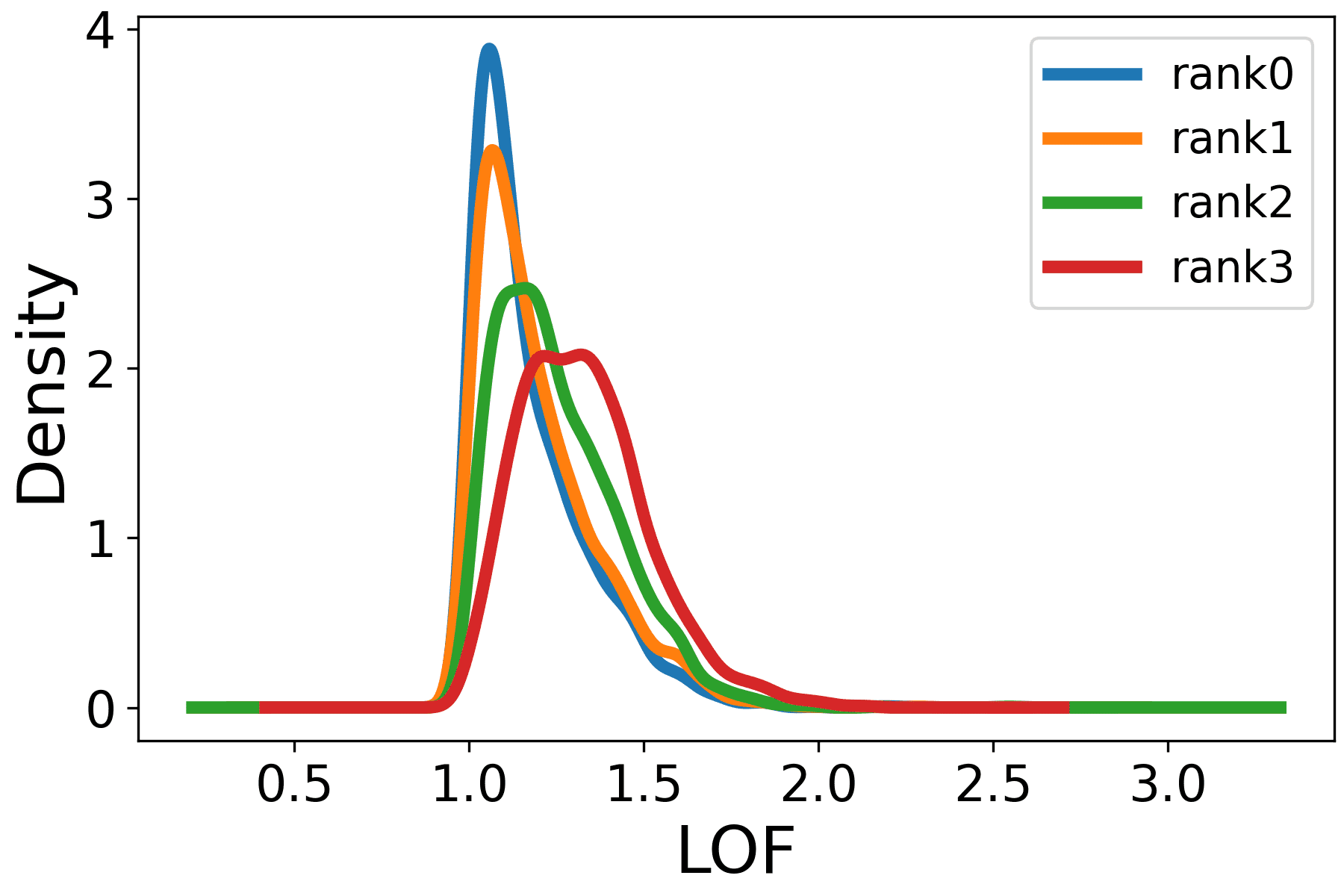}
\label{936_density}
\end{minipage}
}
\caption{Density estimation curves for generated samples from the classes Water tower, Snow leopard, and Head cabbage. As the accumulated score differences decrease from rank 0 (highest) to rank 3 (lowest), we observe a systematic shift of samples from high-density to low-density regions.} 
\label{ImageNet_density app4}
\end{figure}

Figures~\ref{ImageNet_density app2} to~\ref{ImageNet_density app4} show additional density plots across various class labels. Consistent with our observations in the toy setting, samples with lower accumulated score differences \(\mathcal{E}_{T}(c)\) tend to exhibit higher AvgkNN and LOF scores, indicating a stronger presence in low-likelihood regions of the data manifold. These results provide further empirical support for the positive correlation between ASD and sample density, reinforcing the foundation of our filtering approach.

\vspace{-0.1cm}
\vspace{-0.1cm}

\vspace{-0.1cm}
\vspace{-0.1cm}

\vspace{-0.1cm}
\vspace{-0.1cm}

\subsection{Qualitative comparison}
\label{Ima_qua}
\begin{figure}[htbp]
\centering
\includegraphics[width=0.9\linewidth]{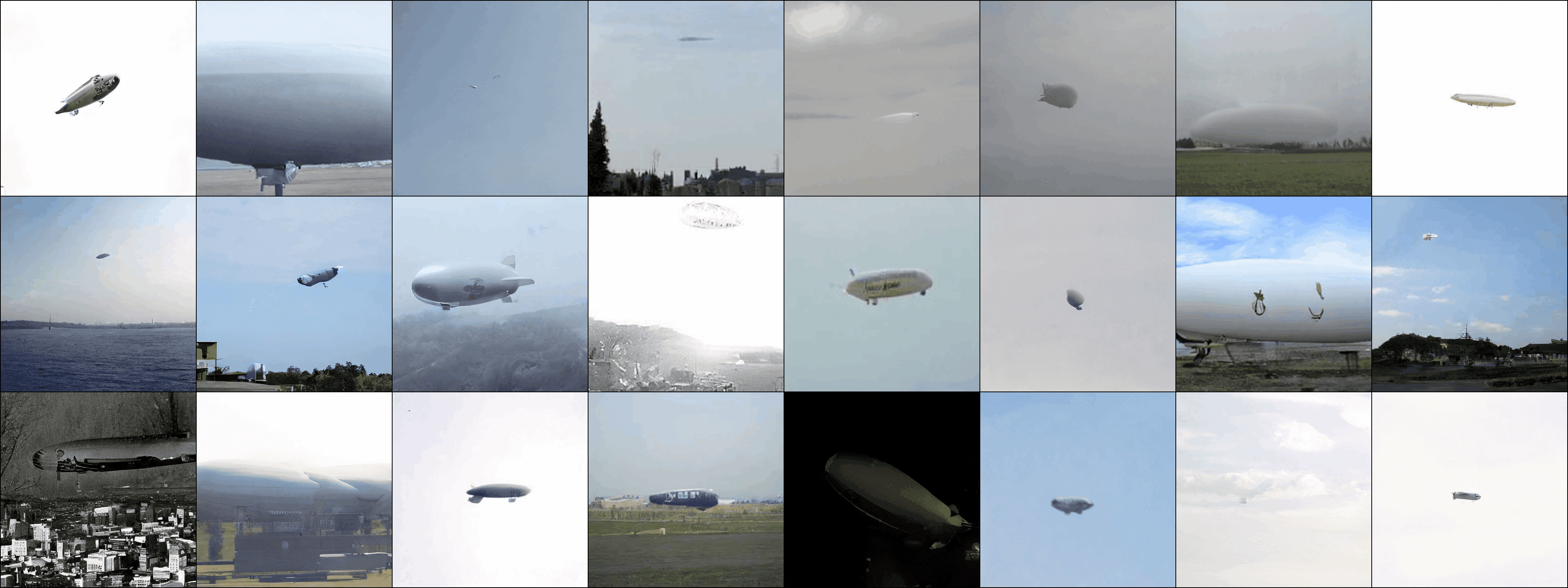}
\vspace{0.5cm}
\includegraphics[width=0.9\linewidth]{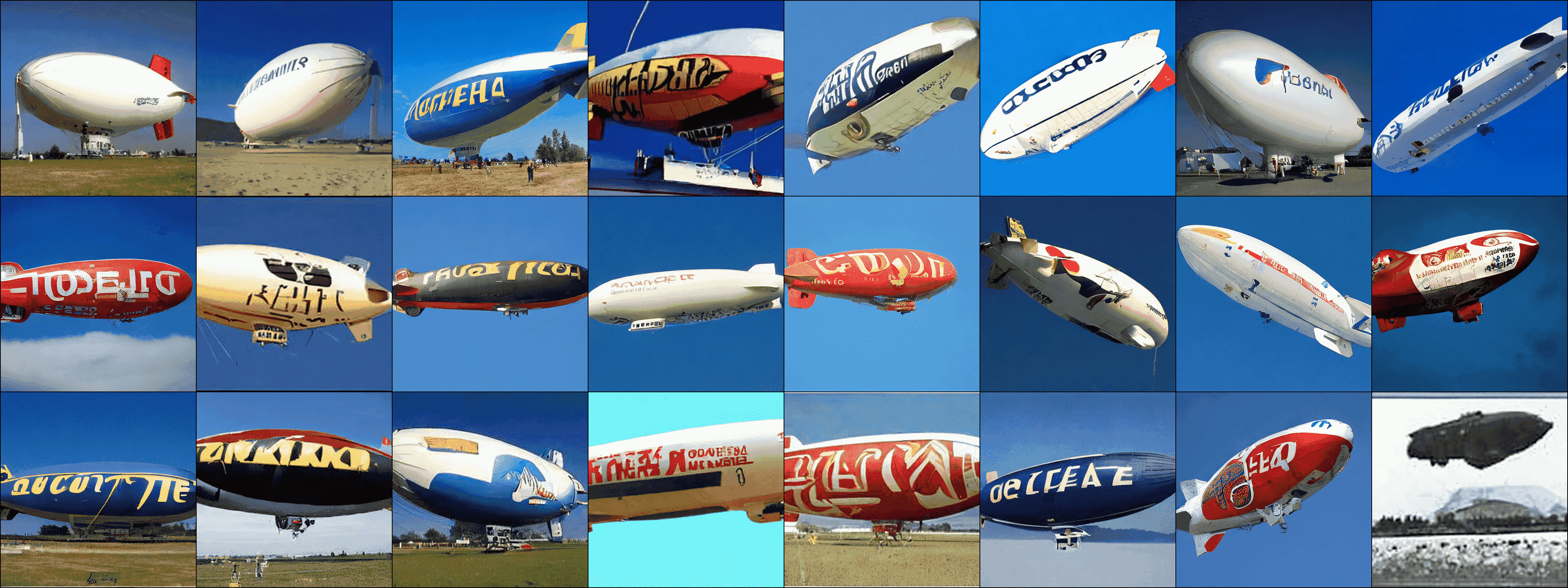}
\vspace{-0.5cm}
\caption{Qualitative comparison of Airship. The top row contains samples with the lowest accumulated score difference (ASD) \(\mathcal{E}_{T}(c)\), where the airships appear faint, occupy minimal space in the image, or are indistinguishable from the sky background—suggesting uncertainty in semantic focus. In contrast, the bottom row showcases high-ASD \(\mathcal{E}_{T}(c)\) samples, where airships are prominently rendered with clear visual traits such as elongated fuselages, tail fins, and gondolas suspended beneath the envelope. This comparison highlights how high-ASD images better align with the semantic concept of the “Airship” class, confirming the effectiveness of ASD as a density-aware signal for filtering.} 
\label{405_img}
\end{figure}

\begin{figure}[htbp]
\centering
\includegraphics[width=0.9\linewidth]{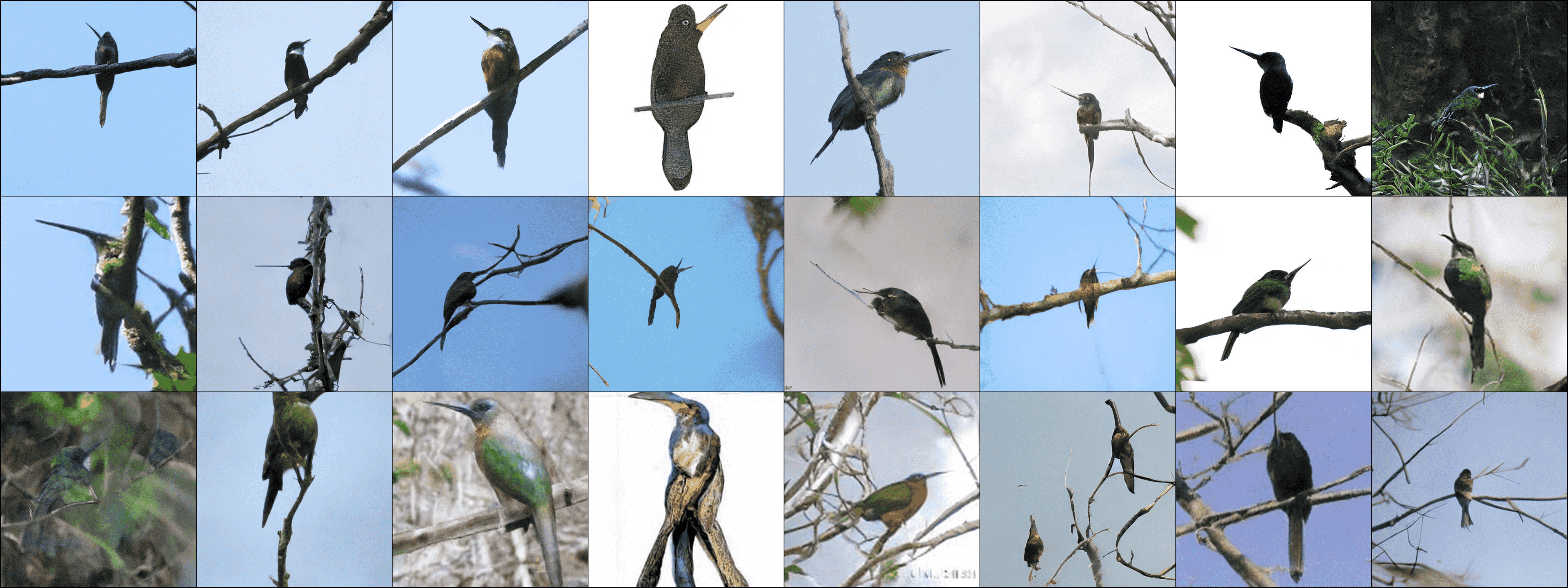}
\vspace{0.5cm}
\includegraphics[width=0.9\linewidth]{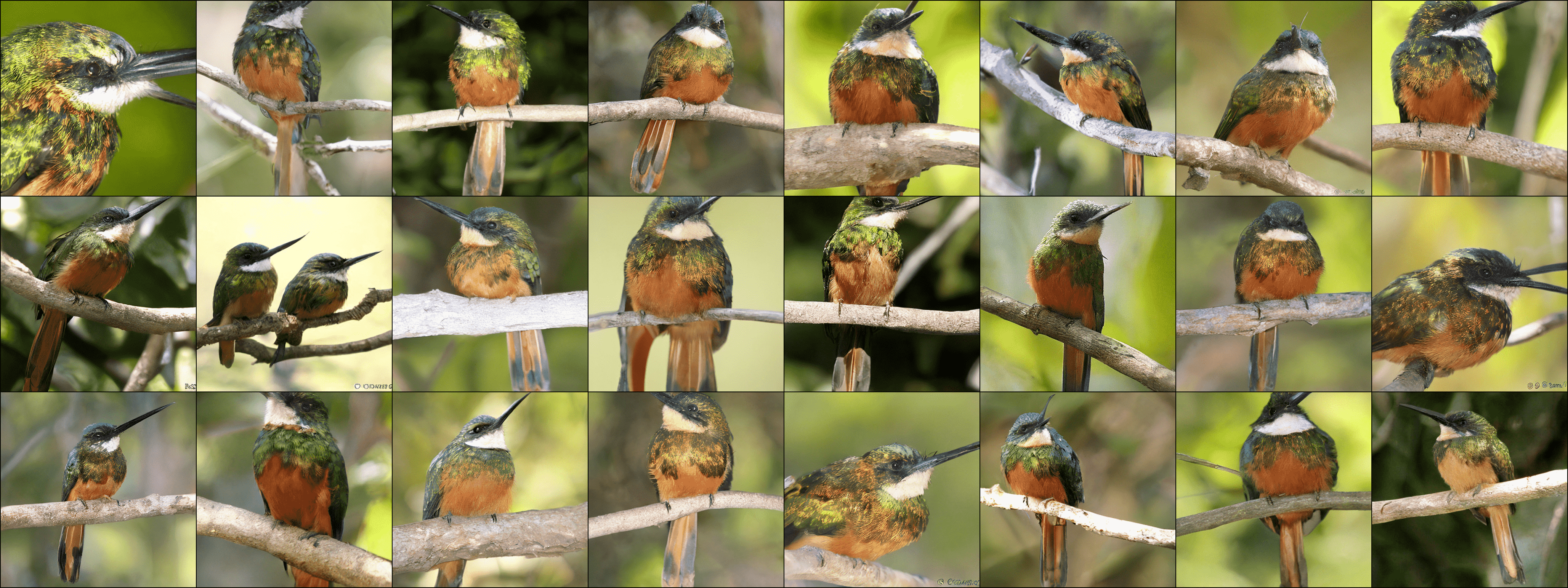}
\label{95_img}
\vspace{-0.5cm}
\caption{Qualitative comparison of Hummingbird.  Top-row samples often depict cluttered tree branches or expansive skies where hummingbirds are visually ambiguous or absent. In contrast, bottom-row images center clearly identifiable hummingbirds, typically showing compact bodies and iridescent feathers.} 
\end{figure}

\begin{figure}[htbp]
\centering
\includegraphics[width=0.9\linewidth]{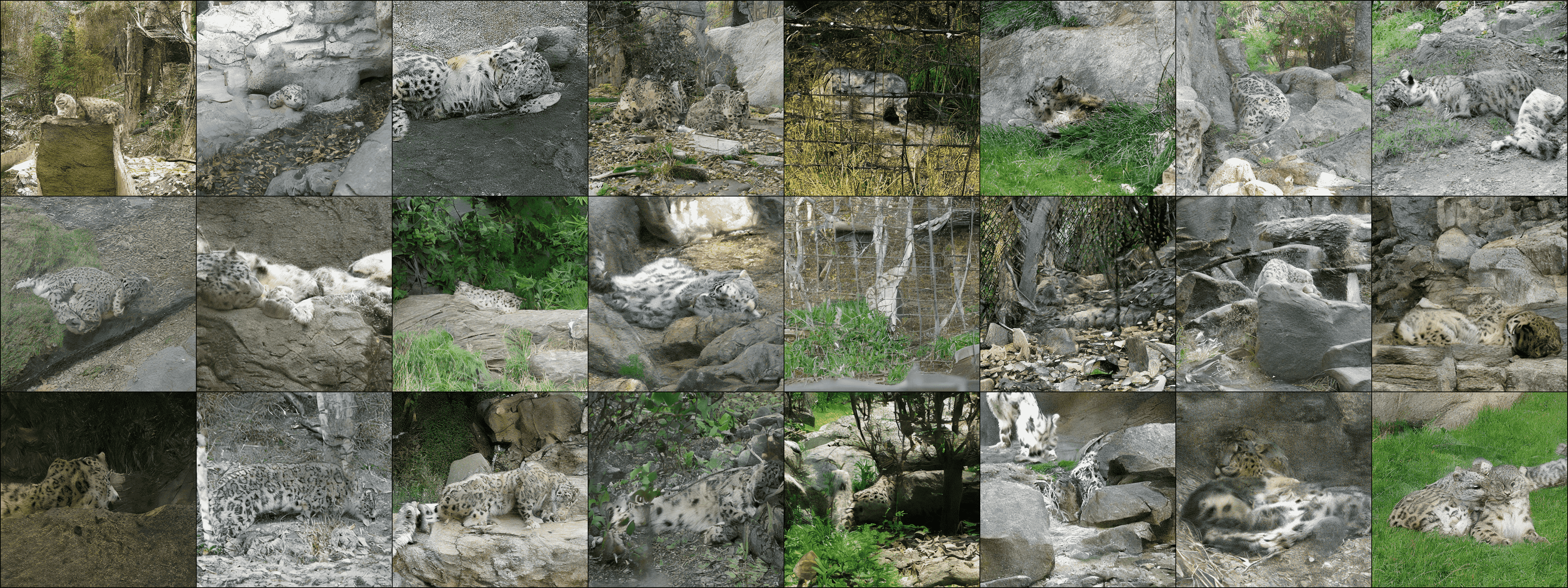}
\vspace{0.5cm}
\includegraphics[width=0.9\linewidth]{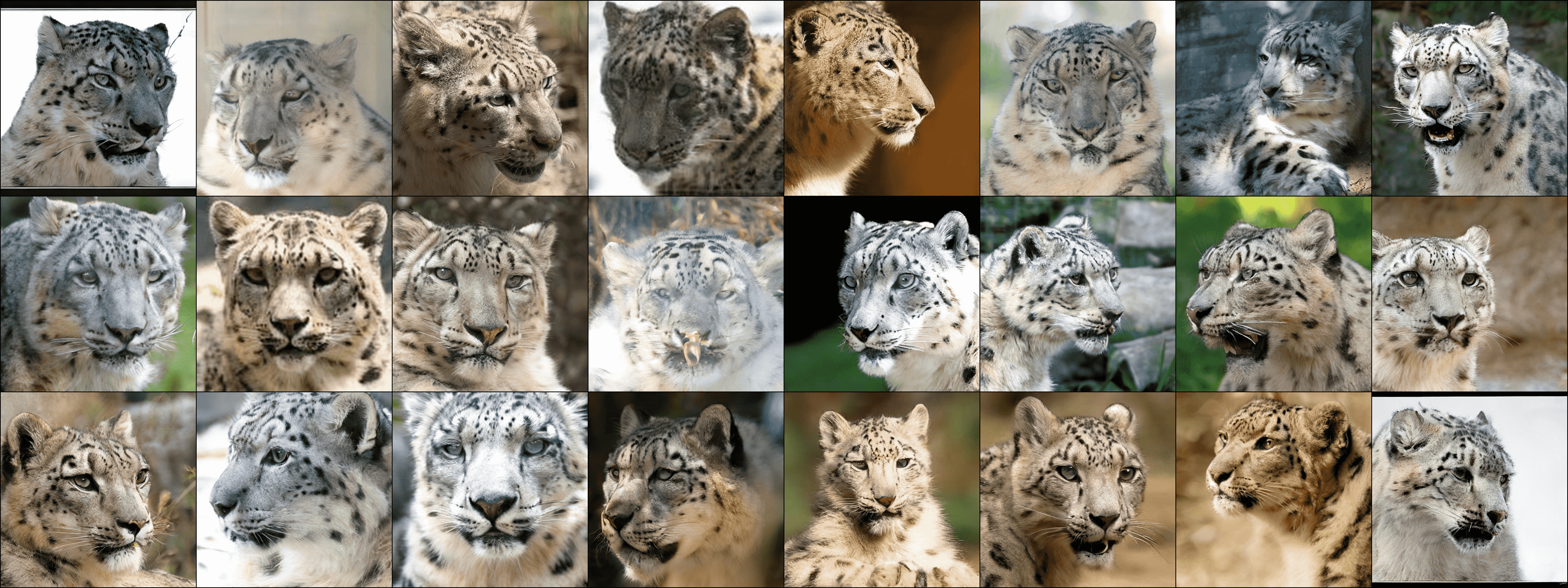}
\label{289_img}
\vspace{-0.5cm}
\caption{Qualitative comparison of Snow Leopard. Top-row samples depict distorted or obscured snow leopards blending into rocky streams, making recognition difficult. Bottom-row images feature clearly centered snow leopards with frontal faces, showing distinct fur patterns and facial structure.} 
\end{figure}

\begin{figure}[htbp]
\centering
\includegraphics[width=0.9\linewidth]{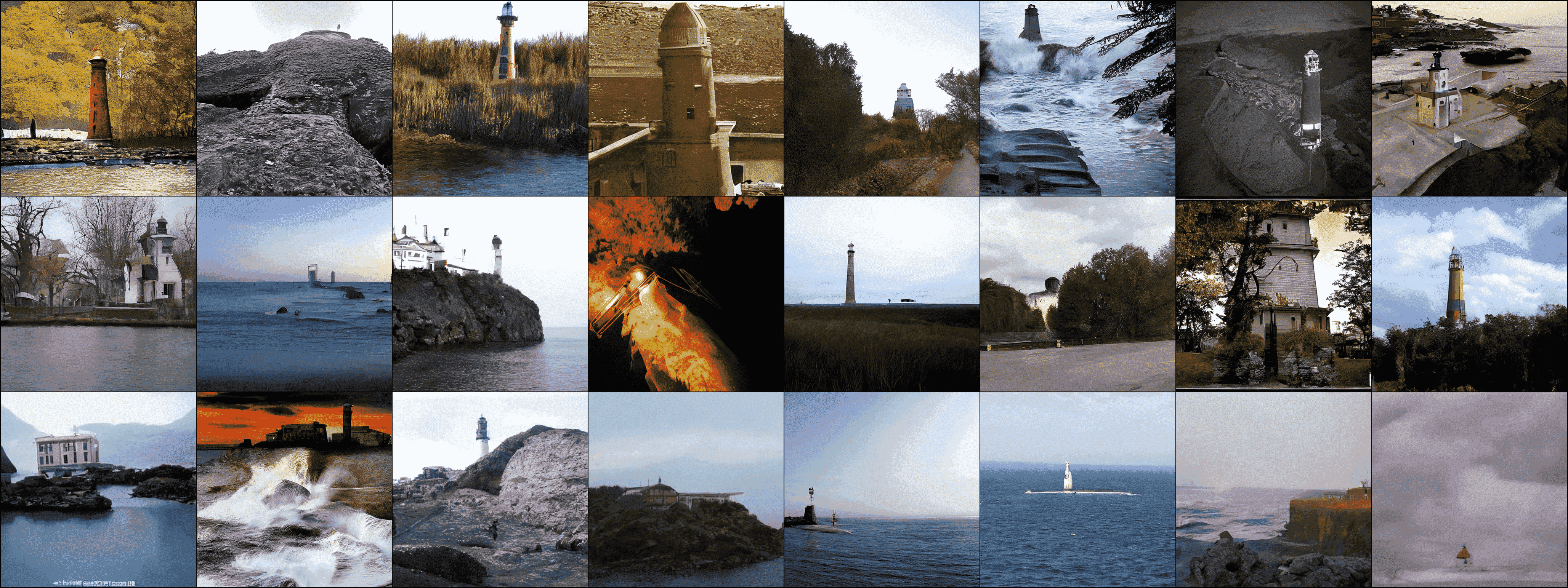}
\vspace{0.5cm}
\includegraphics[width=0.9\linewidth]{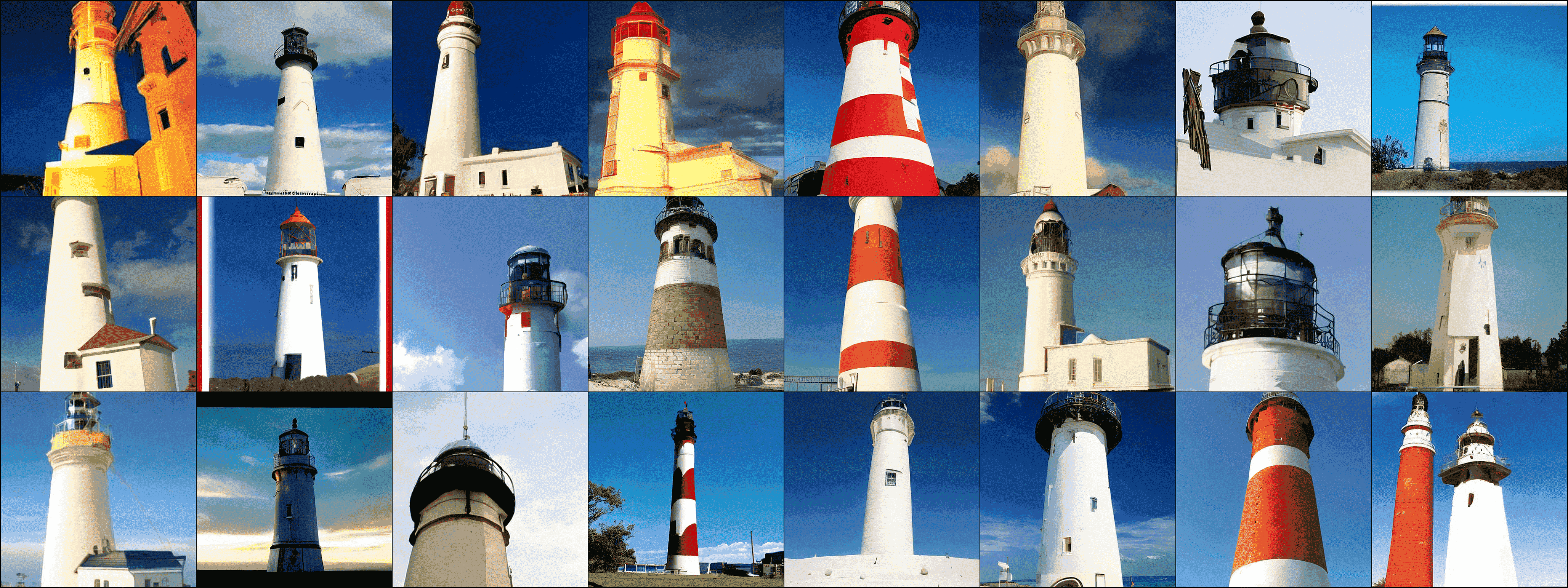}
\label{437_img}
\vspace{-0.5cm}
\caption{Qualitative comparison of Beacon. Top-row images show tiny beacons embedded in varied backgrounds like seascapes and rocky terrain, often difficult to identify. In contrast, bottom-row samples feature prominent, clearly defined beacons with distinctive tower structures.}
\end{figure}

\begin{figure}[htbp]
\centering
\includegraphics[width=0.9\linewidth]{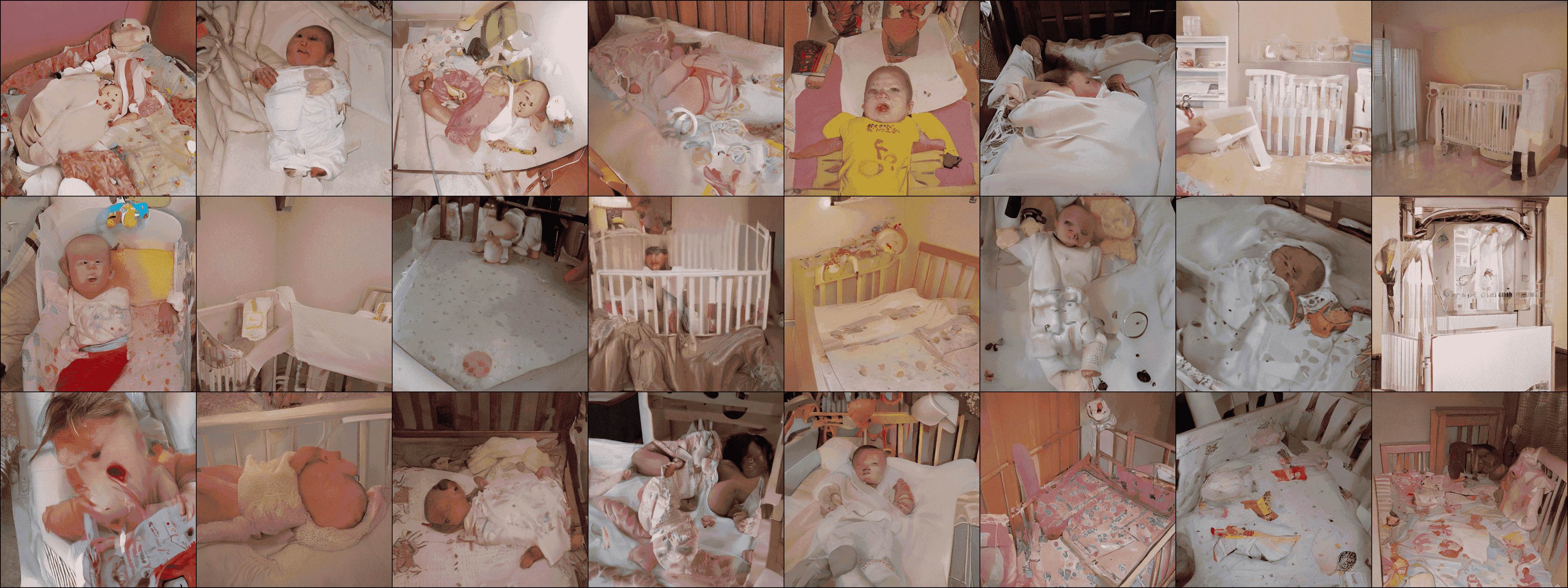}
\vspace{0.5cm}
\includegraphics[width=0.9\linewidth]{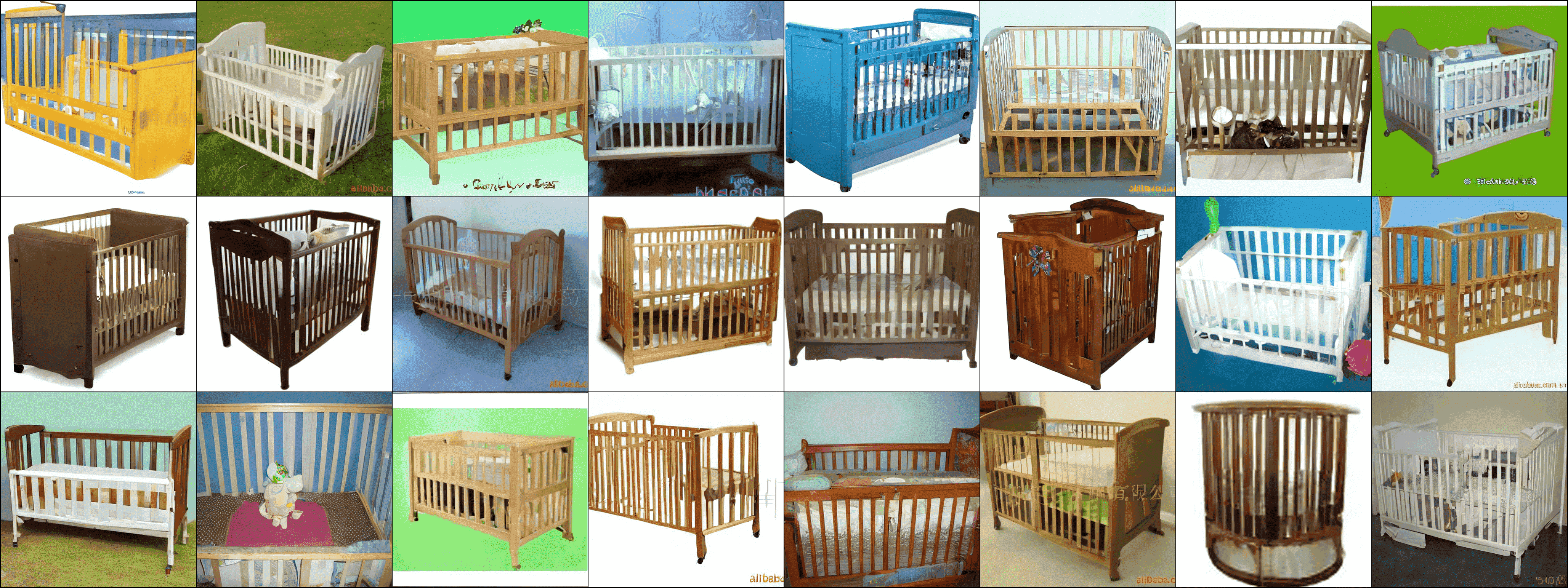}
\label{520_img}
\vspace{-0.5cm}
\caption{Qualitative comparison of Crib. Top-row images depict distorted infants in cluttered bedding, lacking structural clarity. Bottom-row samples present recognizable cribs with wooden railings and coherent furniture structure.} 
\end{figure}

\begin{figure}[htbp]
\centering
\includegraphics[width=0.9\linewidth]{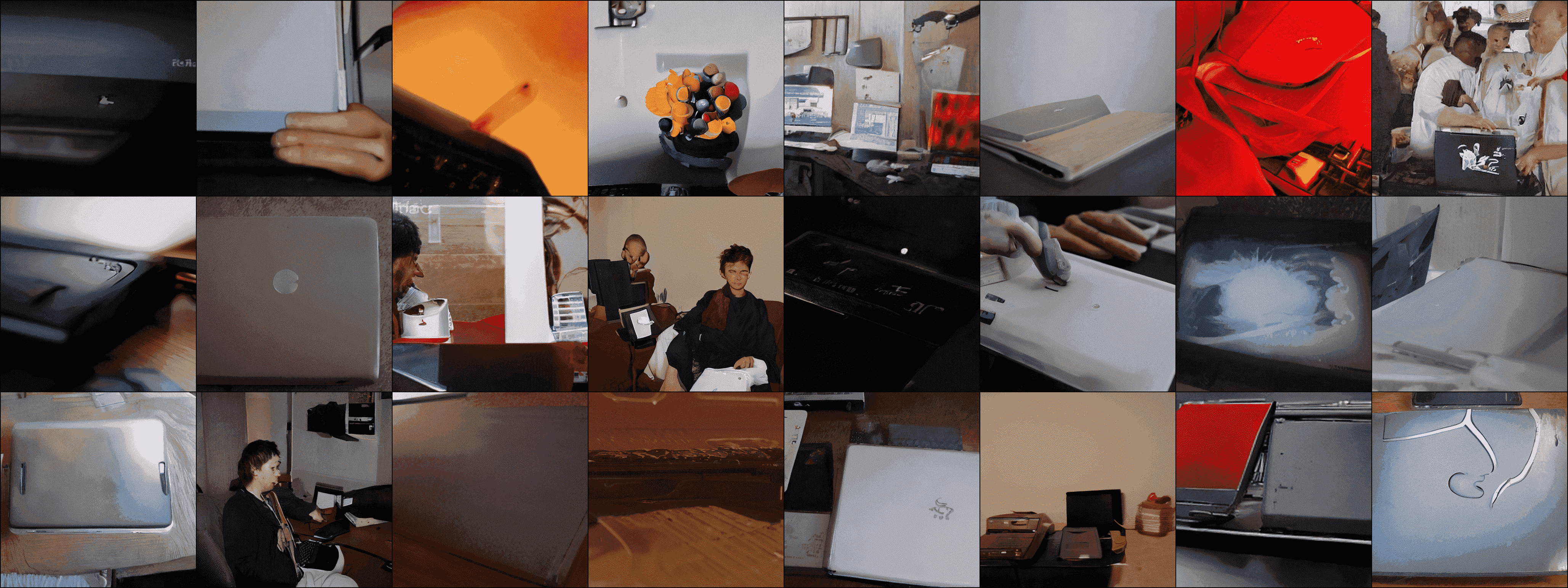}
\vspace{0.5cm}
\includegraphics[width=0.9\linewidth]{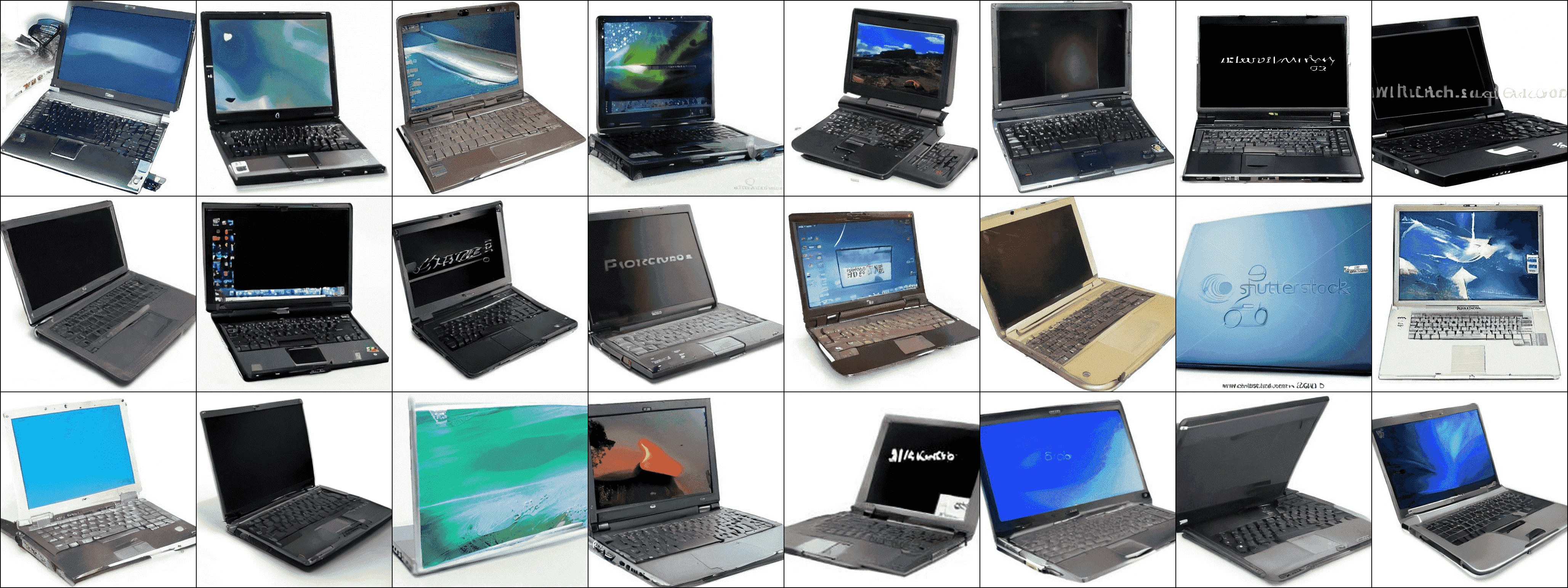}
\label{681_img}
\vspace{-0.5cm}
\caption{Qualitative comparison of Notebook. Top-row images depict distorted Notebook in complex backgrounds, resulting in ambiguous and unclear features. Bottom-row images show clearly presented Notebook in open-close states with well-defined, distinctive structural details.} 
\end{figure}

\begin{figure}[htbp]
\centering
\includegraphics[width=0.9\linewidth]{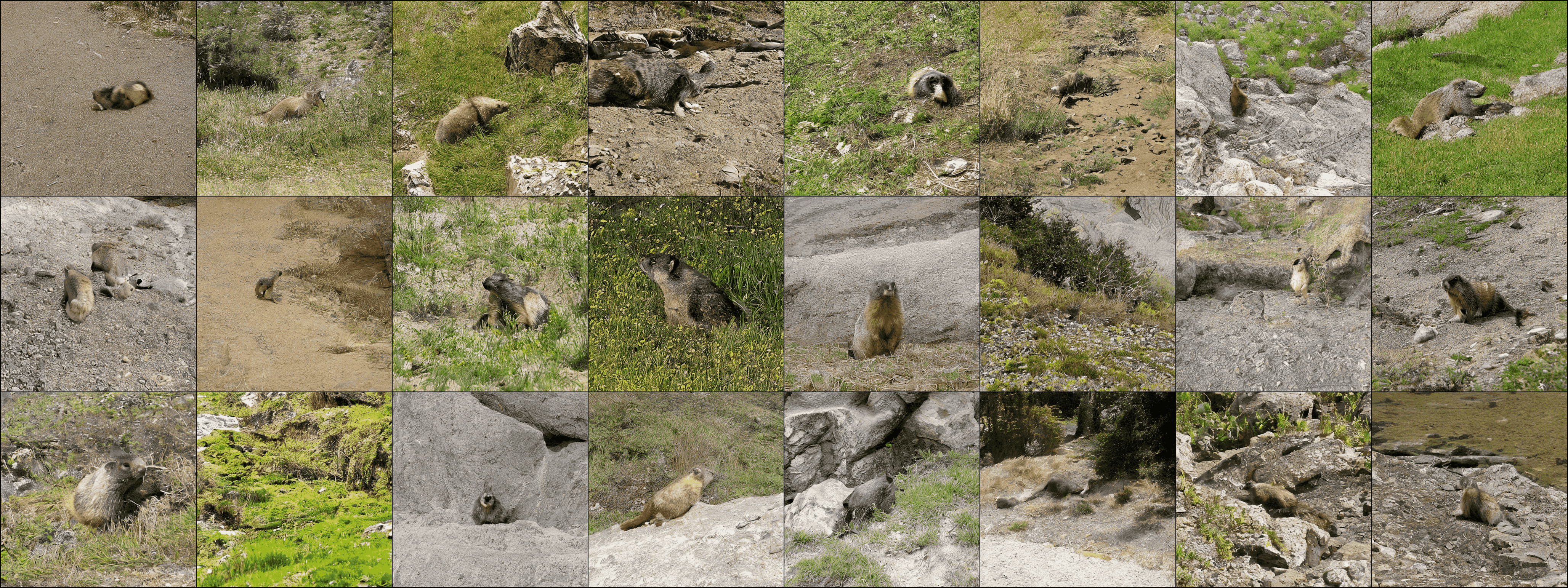}
\vspace{0.5cm}
\includegraphics[width=0.9\linewidth]{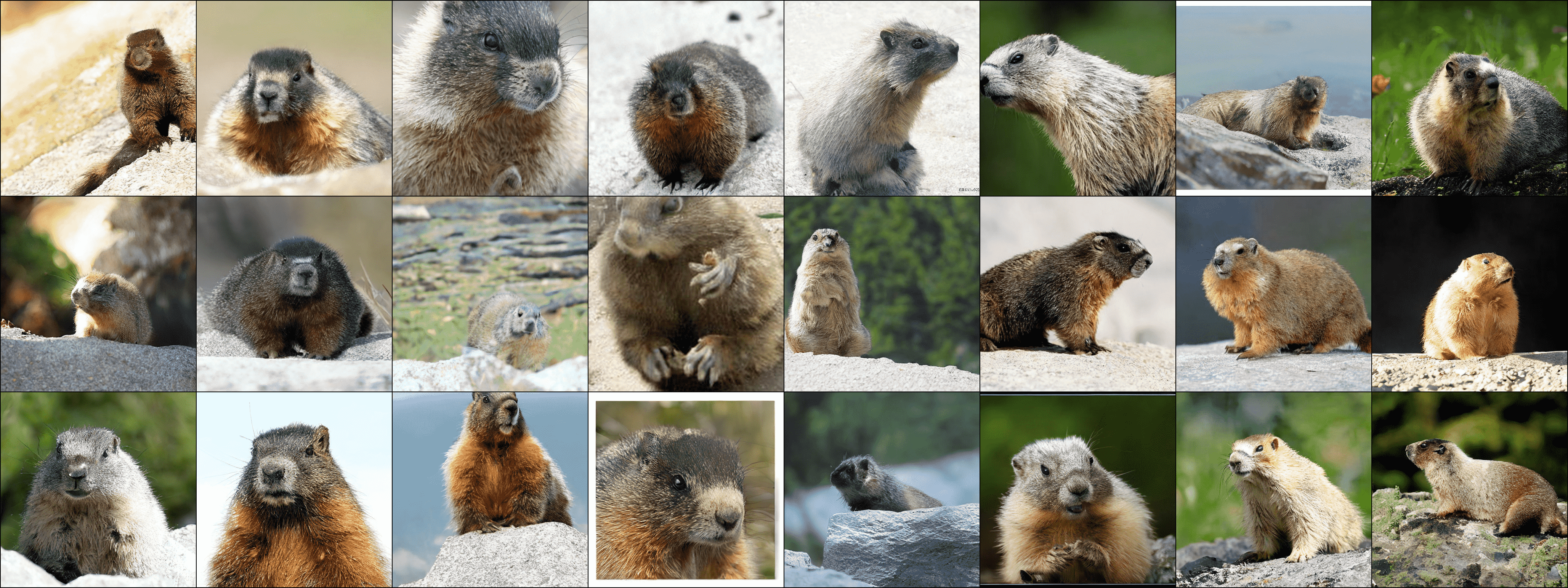}
\vspace{-0.5cm}
\caption{Qualitative comparison of Marmot. Top-row samples depict marmots blending into expansive grassland or rocky backgrounds, often distorted or partially obscured, making recognition difficult. Bottom-row images feature clearly centered marmots showing distinct fur texture, facial features, and characteristic posture.} 
\label{336_img}
\end{figure}

Figures~\ref{405_img} to~\ref{336_img} present additional qualitative results, showcasing samples with the highest and lowest accumulated score differences \(\mathcal{E}_{T}(c)\) across various labels. As illustrated, samples with low \(\mathcal{E}_{T}(c)\) often exhibit severe artifacts or semantic inconsistencies, while those with high \(\mathcal{E}_{T}(c)\) maintain high visual fidelity. These results further highlight the effectiveness of our method in filtering out degenerate generations while retaining high-quality outputs.

\subsection{Quantitative comparison}
\label{Ima_baseline}
To mitigate selection overfitting, we perform a cross-validation study comparing CFG-Rejection with Best-of-N strategy under identical time constraints. Specifically, for Best-of-N, we use Aesthetic Score for selection, then evaluate performance using PickScore and HPSv2 on the selected outputs. The comparison metric is the average score of 1,000 selected samples.

To control inference time, we vary the number of initial candidates in CFG-Rejection from 6,000 to 2,000 (in steps of 100), and adjust Best-of-N accordingly to match the time budget. The x-axis in Figure~\ref{compare_baseline} reflects this normalized computational cost.

\begin{figure}[H]
\centering  
\subfigure[Pick Score comparison]{   
\begin{minipage}{0.4\textwidth}
\centering   
\includegraphics[width=\linewidth]{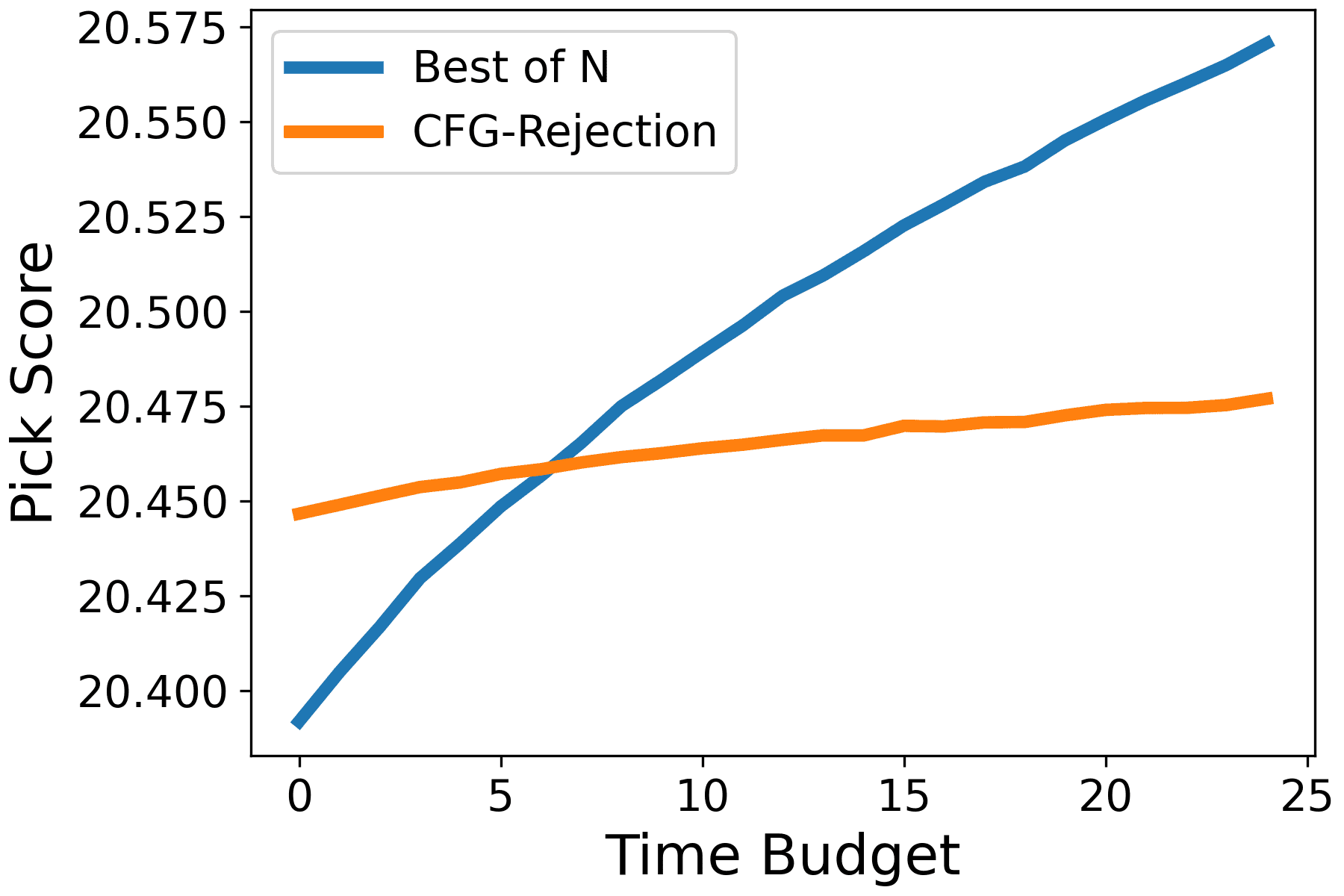}  
\label{pick_app}
\end{minipage}
}
\subfigure[HPSv2 Score comparison]{
\begin{minipage}{0.4\textwidth}
\centering   
\includegraphics[width=\linewidth]{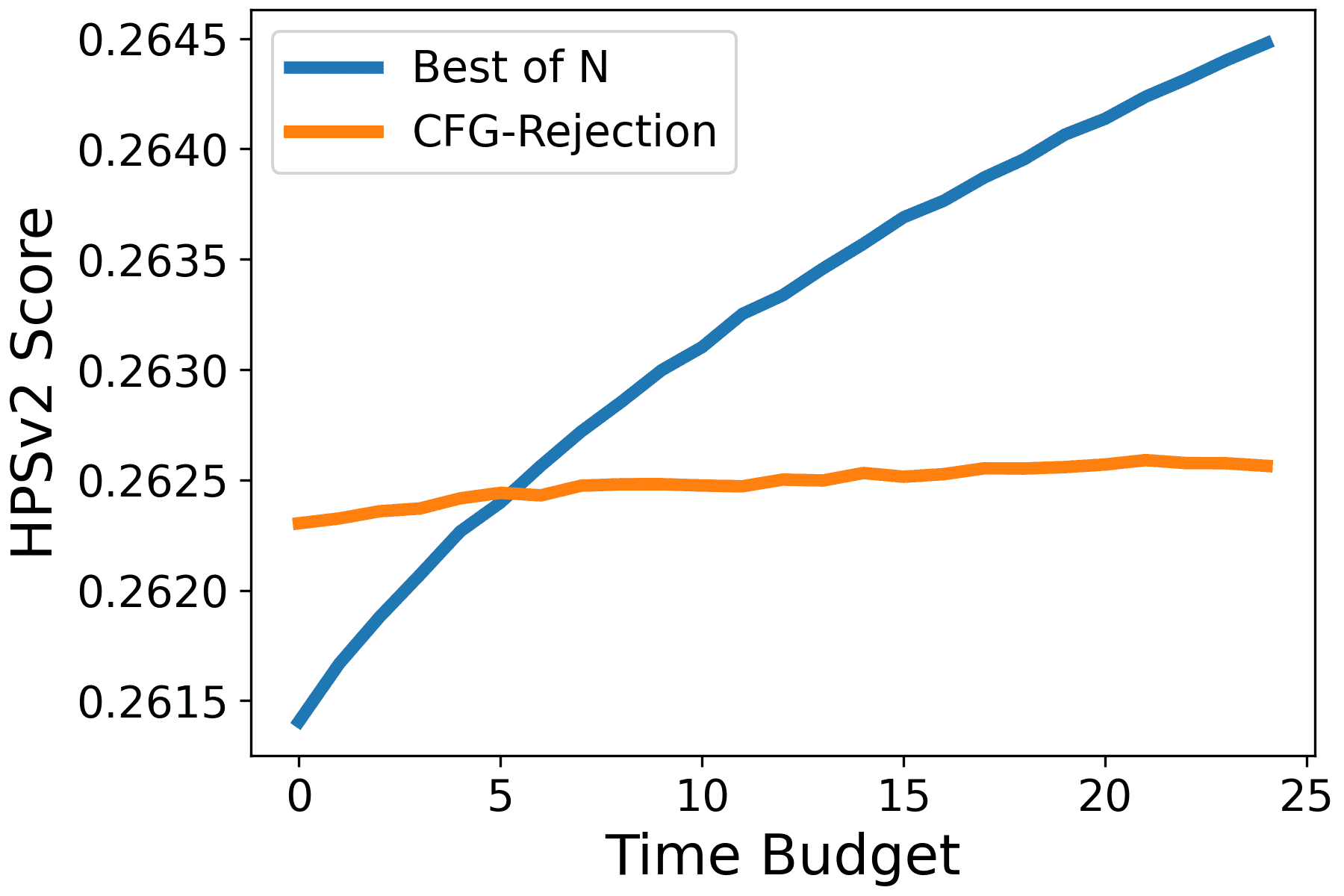}
\label{hpv2_app}
\end{minipage}
}
\caption{Performance comparison under limited inference budget. CFG-Rejection outperforms Best-of-N method under constrained computational budgets in practical usage.}
\label{compare_baseline2}
\end{figure}
\vspace{-0.3cm}

We further extend the cross-validation setup using PickScore and HPSv2 as selection criteria for Best-of-N. The results, shown in Figure~\ref{compare_baseline2}, exhibit the same trend as before. Although Best-of-N improves with more time, it initially underperforms due to the need to fully denoise all candidates before selection. In contrast, CFG-Rejection benefits from early-stage filtering, achieving higher scores in low-time regimes. These findings reinforce the effectiveness and efficiency of our method under realistic computational constraints.

\section{More results on Geneval and DPG}
\label{bench_app}
In this section, we detail the experimental settings for tracking accumulated score differences using SDv1.5 and SDXL, along with quantitative results for SDXL. All image generations are performed on 5 NVIDIA H100 GPUs, while benchmark evaluations are conducted on a single RTX 4090 GPU. For each experiment, filtering is applied with three distinct initial seeds, and the reported results correspond to the average scores.

For SDv1.5, we employ the default PNDM scheduler at a resolution of \(512{\times}512\). The latent noise dimension is \(4{\times}64{\times}64\). To track the score difference \(\mathcal{G}_t(c)\), we compute the direct difference between the outputs of the conditional and unconditional models, flatten the resulting tensors into a vector, and measure its \(\ell_2\)-norm. 

For SDXL, we utilize the default Flow Match Euler Discrete scheduler with a resolution of \(1024{\times}1024\). The latent noise dimension is \(4{\times}128{\times}128\). The procedure for tracking \(\mathcal{G}_t(c)\) follows that of SDv1.5, with the key distinction being a scaling by \(\sigma_t\) at each step \(t\) due to the way noise estimation is incorporated in different sampling schedulers. 

\subsection{GenEval experiment}
\label{gen_sdxl}
The quantitative results on GenEval using SDXL are presented in Table~\ref{sdxl_gen}. As discussed in Section~\ref{paper_gen}, the performance gain of CFG-Rejection with SDXL is less pronounced compared to SDv1.5. We attribute this discrepancy to SDXL’s inherent tendency to generate samples concentrated in high-density regions of the latent space, which limits the effectiveness of density-based filtering strategies. Notably, when the guidance scale is set to \(\omega=9\), little improvement is observed. In this case, the excessive guidance leads to further concentration in high-density areas, rendering CFG-Rejection largely ineffective.

Furthermore, we observe that the two filtering configurations, retaining 4 out of 20 and 4 out of 50 candidates, yield nearly identical performance. This observation is consistent with the trend shown in Figures~\ref{compare_baseline} and~\ref{compare_baseline2}, where performance improves only marginally with increasing sampling budget.

These results suggest that CFG-Rejection is particularly well suited for two scenarios: (1) when using high-diversity models where generated images are distributed across regions of varying density, and (2) when operating under constrained inference budgets, where evaluating all candidates for Best-of-N selection is computationally prohibitive.

\begin{table}[htbp]
  \centering
  \small
  \caption{The quantitative results on GenEval. Model: SDXL}
  \vspace{1em} 
  \label{sdxl_gen}
  \begin{tabular}{@{} l l *{6}{c} c @{}}
    \toprule
    & Method& Single Obj. & Two Obj. & Counting & Colors & Position & Color Attri. & Overall\(\uparrow\)\\
    \midrule
    \multirow{7}{*}{guidance = 5} 
    & random & 0.9714 & 0.6987 & 0.4094 & 0.8493 & \textbf{0.1121} & 0.1942 & 0.5393 \\
    & 4 from 20 & \textbf{0.9813} & 0.7639 & 0.3703 & 0.8405 & 0.1013 & 0.225 & 0.5470 \\
    & 4 from 50 & 0.9781 & \textbf{0.7677} & 0.3250 & \textbf{0.8617} & 0.0875 & 0.2225 & 0.5404 \\
    & \(\tau=10\)  & 0.9766 & 0.7197 & \textbf{0.425} & 0.8484 & 0.09  & 0.2125 & 0.5454 \\
    & \(\tau=20\)  & 0.9781 & 0.7589 & 0.3844 & 0.8444 & 0.0975 & 0.2313 & \textbf{0.5491} \\
    & \(\tau=30\) & \textbf{0.9813} & 0.7652 & 0.3813 & 0.8378 & 0.10 & \textbf{0.2238}  & 0.5482 \\  
    & \(\tau=40\) & \textbf{0.9813} & 0.7665 & 0.3703 & 0.8405 & 0.1025 & 0.225 & 0.5477 \\ 
    \midrule
    \multirow{7}{*}{guidance = 9} 
    & random & 0.9828 & 0.7399 & \textbf{0.4349} & \textbf{0.8759} & \textbf{0.120} & 0.2213 & 0.5625 \\
    & 4 from 20 & 0.9891 & 0.7942 & 0.3672 & 0.8631 & 0.10 & 0.2538 & 0.5612 \\
    & 4 from 50 & \textbf{0.9938} & 0.7828 & 0.3281 & 0.8590 & 0.09 & 0.2525 & 0.5510 \\
    & \(\tau=10\)  & 0.9906 & 0.7652 & 0.4297 & 0.8684 & 0.0925 & 0.2588 & 0.5675 \\
    & \(\tau=20\) & 0.9907 & 0.7867 & 0.4031 & 0.8644 & 0.1013 & \textbf{0.2613} & \textbf{0.5679} \\
    & \(\tau=30\) & 0.9907 & \textbf{0.7993} & 0.3828 & 0.8590 & 0.1013 & 0.2538 & 0.5645 \\  
    & \(\tau=40\) & 0.9891 & 0.7967 & 0.3734 & 0.8604 & 0.1013 & 0.255 & 0.5626 \\ 
    \bottomrule
  \end{tabular}
\end{table}

\begin{table}[H]
  \centering
  \small
  \caption{The quantitative results on DPG-bench. Model: SDXL}
  \vspace{1em} 
  \label{sdxl_dpg}
  \begin{tabular}{@{} l l *{5}{c} c @{}} 
    \toprule
    & Method& Global & Entity & Attribute & Relation & Other. & Overall\(\uparrow\)\\
    \midrule
    \multirow{6}{*}{guidance = 5} 
    & random & 85.46 & 80.86 & 79.50 & \textbf{86.34} &62.6& 73.54  \\
    & 4 from 20 & 83.44 & 81.95 & 79.98 & 85.87 & \textbf{64.8} & 74.58  \\
    & 4 from 50 & 84.50 & 81.98 & 79.64 & 85.58 & 64.4 & 74.52  \\
    & \(\tau=10\)  & \textbf{85.56} & 81.13 & 79.65 & 85.83 & 62.6 & 73.61  \\
    & \(\tau=20\)  & 83.43 &81.64 & 79.85 & 86.08 & 63.2 & 74.3  \\
    & \(\tau=30\) & 83.43 & 81.82 &79.78 &85.83 & 63.8 & 74.42  \\  
    & \(\tau=40\) & 83.28 &\textbf{81.99} & \textbf{79.99} & 85.79 & 64.4  & \textbf{74.66}  \\ 
    \midrule
    \multirow{6}{*}{guidance = 9} 
    & random & \textbf{85.46} & 82.27 & 80.4 & 87.01 &65.67& 75.16  \\
    & 4 from 20 & 83.89 & 82.84& \textbf{81.42} & 87.1 & 65.8 & \textbf{75.98}  \\
    & 4 from 50 & 82.07 & \textbf{82.92} & 81.22 & 86.46  & 65.60 & 75.71  \\
    & \(\tau=10\)  & 84.35 & 82.05 & 80.87 & 86.64 & 64.2 & 75.06  \\
    & \(\tau=20\)  & 83.59 & 82.61 & 81.15 &86.97 & 65.8 & 75.65\\
    & \(\tau=30\) & 83.59 & 82.75 & 81.26 & \textbf{87.26} & \textbf{66.2} & 75.85 \\  
    & \(\tau=40\) & 83.89 & 82.80 & 81.28 & 87.14 & 65.6  & 75.90  \\ 
    \bottomrule
  \end{tabular}
\end{table}

\subsection{DPG-Bench experiment}
\label{dpg_sdxl}
The quantitative results on DPG-Bench using SDXL are shown in Table~\ref{sdxl_dpg}. Similar trends to those observed on the GenEval benchmark can be found here. First, compared to SDv1.5, the performance gain of CFG-Rejection on SDXL is relatively modest. When varying the guidance scale, a greater improvement is observed at \(\omega=5\) compared to \(\omega=9\), suggesting that our method is more effective in high-diversity generation settings.

Second, the performance under two filtering configurations—selecting 4 out of 20 and 4 out of 50 samples—remains nearly identical, indicating that CFG-Rejection is particularly advantageous in scenarios with limited inference budgets, where full evaluation of all candidates is impractical.

Third, we observe that different generation categories respond differently to changes in the filtering threshold \(\tau\). For instance, the Attribute category shows consistent improvement as \(\tau\) increases from 10 to 40, while the Relation category exhibits a non-monotonic trend: performance first improves and then degrades. This intriguing behavior suggests that the semantic signal captured by the ASD metric may vary depending on the specific demands of the generation task. Exploring this direction further could enable more fine-grained control of generation dynamics using intrinsic signals during inference.

\section{Text rendering results with Flux}
\label{flux_app}
We evaluate our method on the visual text rendering task using FLUX.1-dev. The inference is performed with \(T=28\) steps under the default Flow Match Euler Discrete scheduler and a guidance scale of \(\omega=6\). The procedure for tracking \(\mathcal{G}_t(c)\) follows that of SDXL: we compute the difference between the conditional and unconditional model outputs at each step, flatten the resulting tensor into a vector, and compute its \(\ell_2\)-norm. We further scale this value by \(\sigma_t\) at each step \(t\), consistent with the noise estimation mechanism in the sampling scheduler.

As FLUX is trained with guidance distillation, practical deployment requires an additional model evaluation with \(\omega=1\) to obtain the hidden outputs of the conditional and unconditional branches. The score difference is then tracked using these outputs. All generations and filtering processes are run on a single RTX 4090 GPU.

\begin{figure}[htbp]
\centering  
{
    \begin{minipage}[b]{\linewidth}
        \centering
        \includegraphics[width=\linewidth]{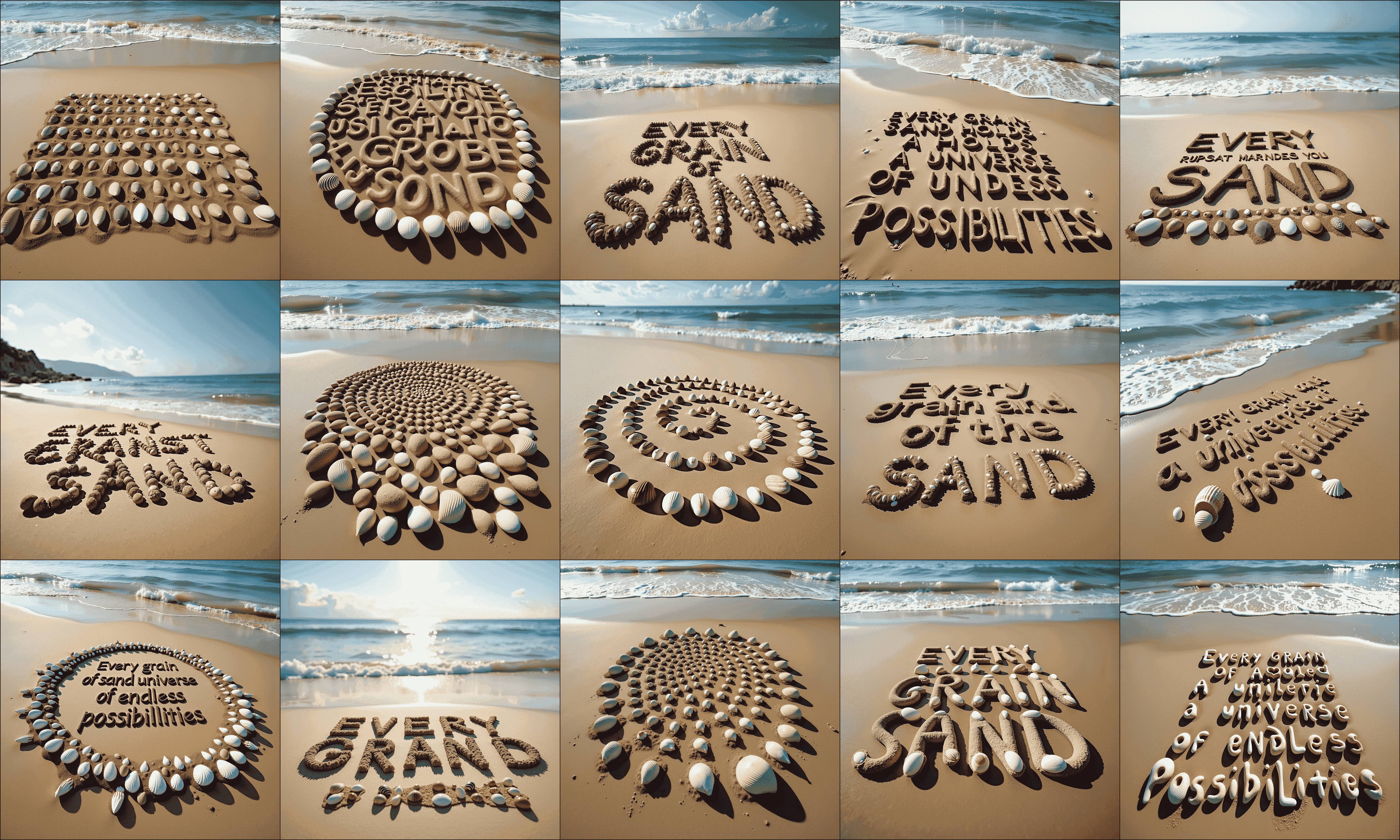}
    \end{minipage}
}

\vspace{0.1cm}

{
    \begin{minipage}[b]{\linewidth}
        \centering
        \includegraphics[width=\linewidth]{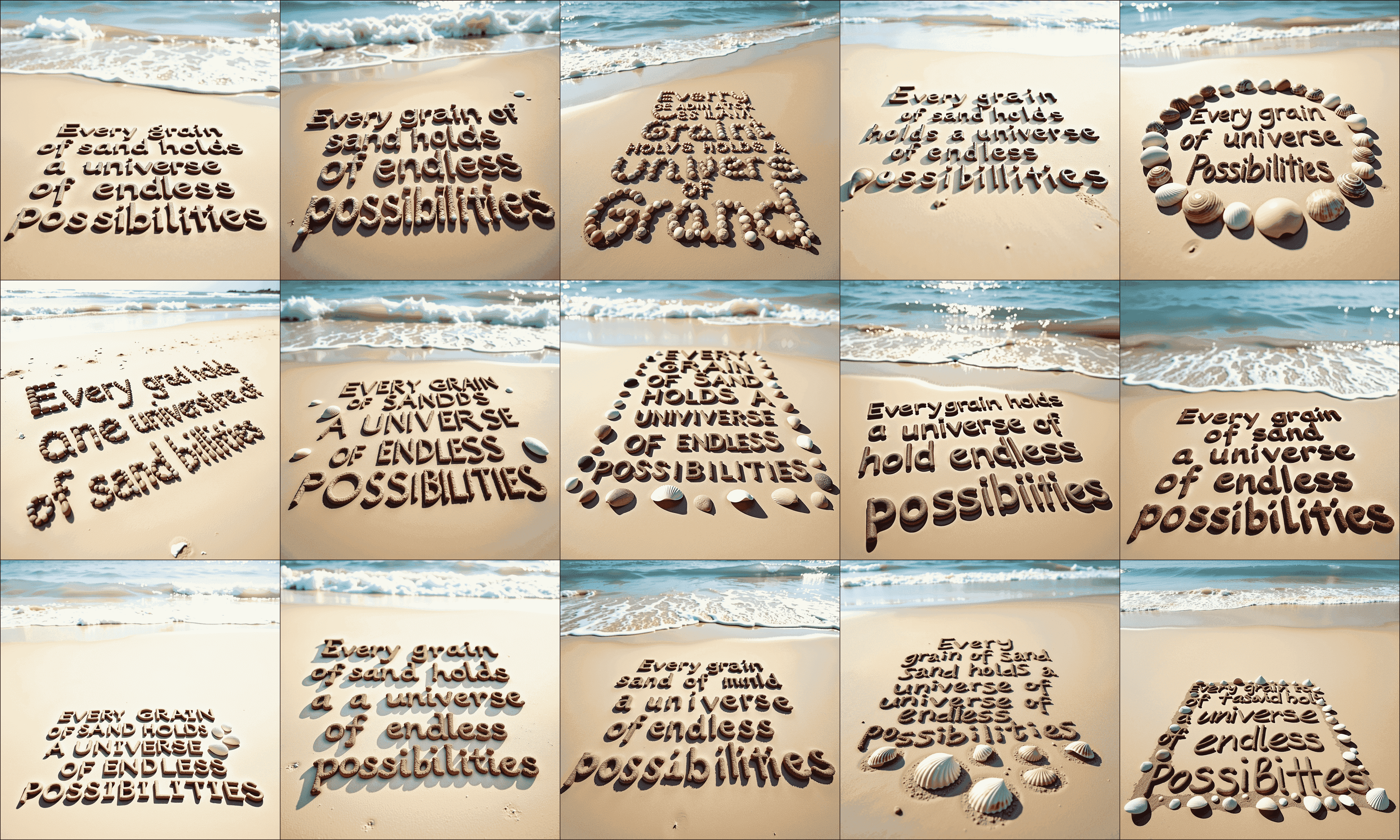}
    \end{minipage}
}%
\vspace{-0.4cm}
  \caption{Visual text rendering for the prompt "A beach with shells organized to form the words 'Every grain of sand holds a universe of endless possibilities'". The top three rows showcase generations with the lowest accumulated score difference (ASD), where many images either omit parts of the phrase or display illegible, fragmented text. In contrast, the bottom three rows depict high-ASD samples, which more reliably render the full phrase with higher clarity and semantic correctness (e.g., well-formed and complete phrases such as "Every grain of sand holds a universe of endless possibilities"). This example further supports ASD’s role in enhancing alignment with complex visual text prompts.}
  \label{flux_app1}
\end{figure}

\begin{figure}[htbp]
\centering  
{
    \begin{minipage}[b]{\linewidth}
        \centering
        \includegraphics[width=\linewidth]{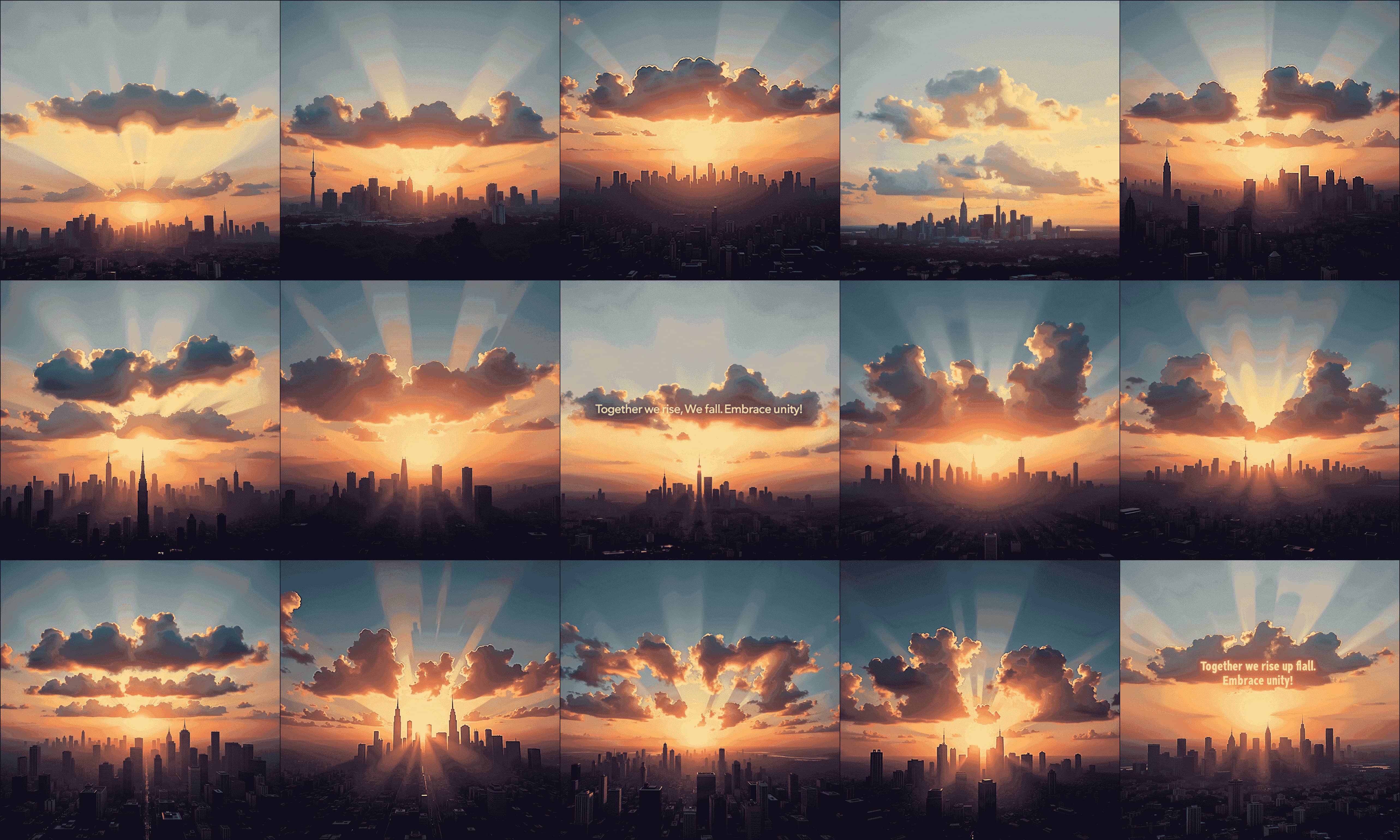}
    \end{minipage}
}

\vspace{0.1cm}

{
    \begin{minipage}[b]{\linewidth}
        \centering
        \includegraphics[width=\linewidth]{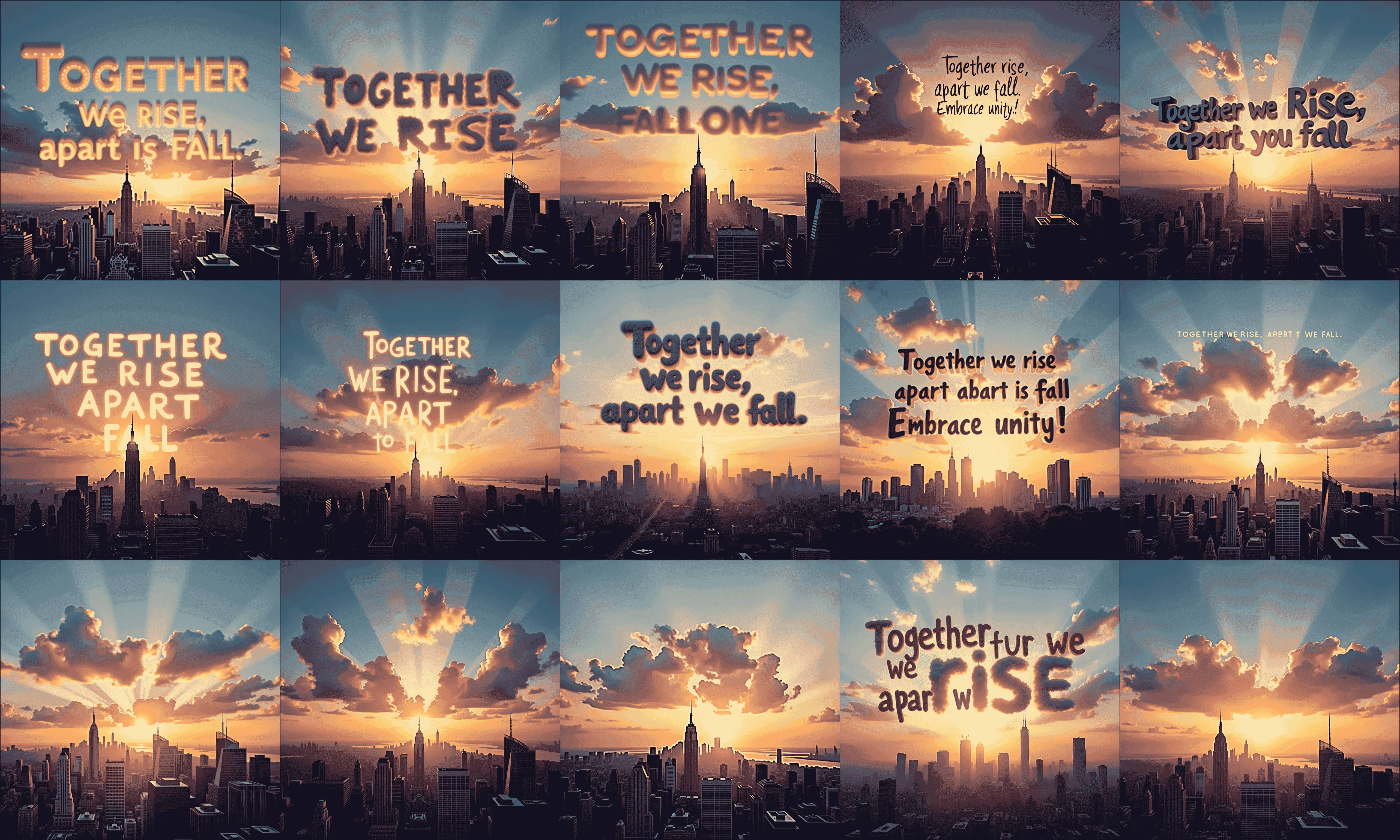}
    \end{minipage}
}%
\vspace{-0.4cm}
  \caption{Visual text rendering for the prompt “A city skyline at sunset with clouds forming the words 'Together we rise, apart we fall. Embrace unity'". The top three rows display images with the lowest accumulated score difference (ASD), while the bottom three rows present those with the highest ASD. High-ASD generations exhibit clearer and more complete textual formations embedded in the clouds, faithfully rendering the intended message (e.g., "TOGETHER WE RISE, apart is FALL"), while low-ASD samples often lack visible or coherent text. This again highlights ASD’s predictive utility in selecting faithful visual text renderings.}
  \label{flux_app2}
\end{figure}

\begin{figure}[htbp]
\centering  
{
    \begin{minipage}[b]{\linewidth}
        \centering
        \includegraphics[width=\linewidth]{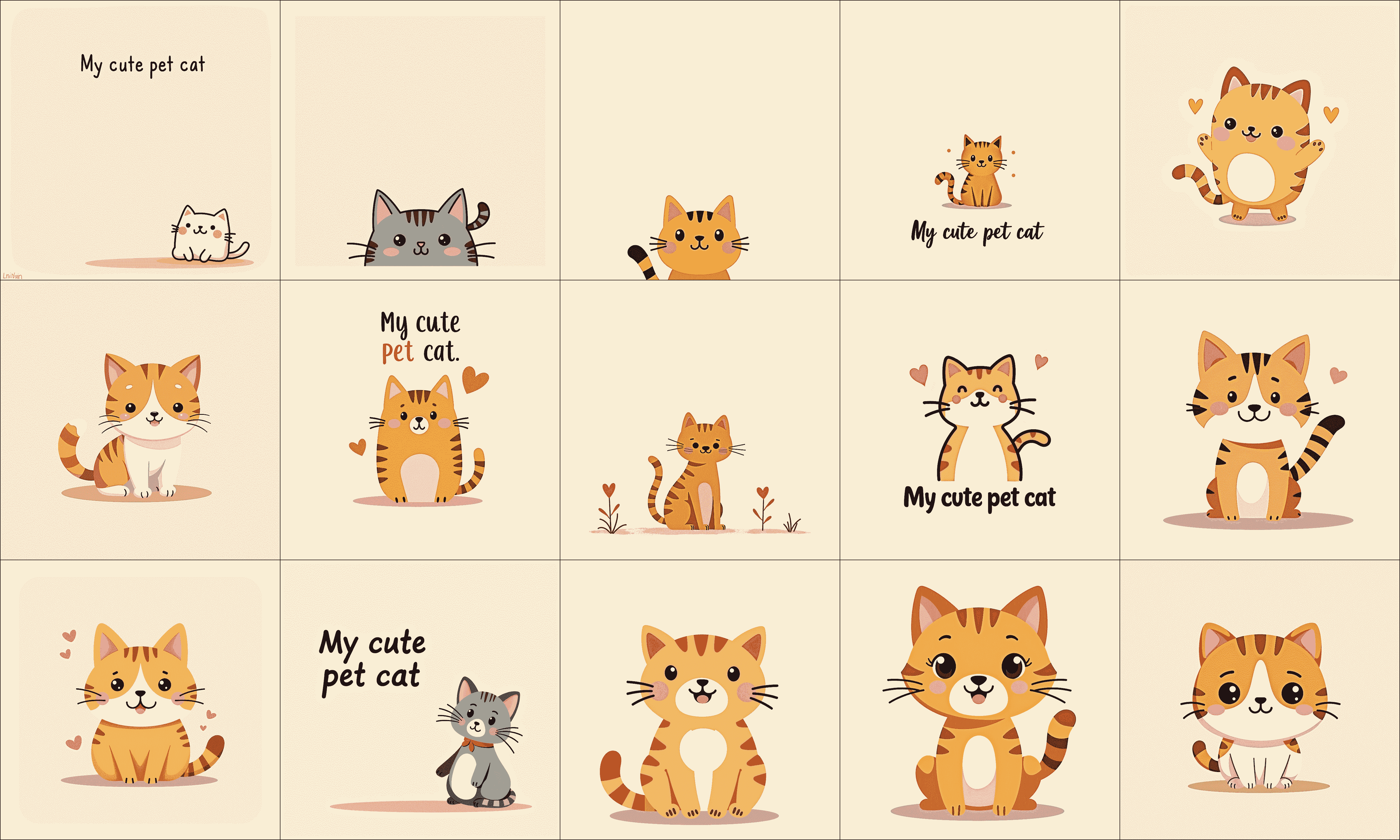}
    \end{minipage}
}

\vspace{-0.1cm}

{
    \begin{minipage}[b]{\linewidth}
        \centering
        \includegraphics[width=\linewidth]{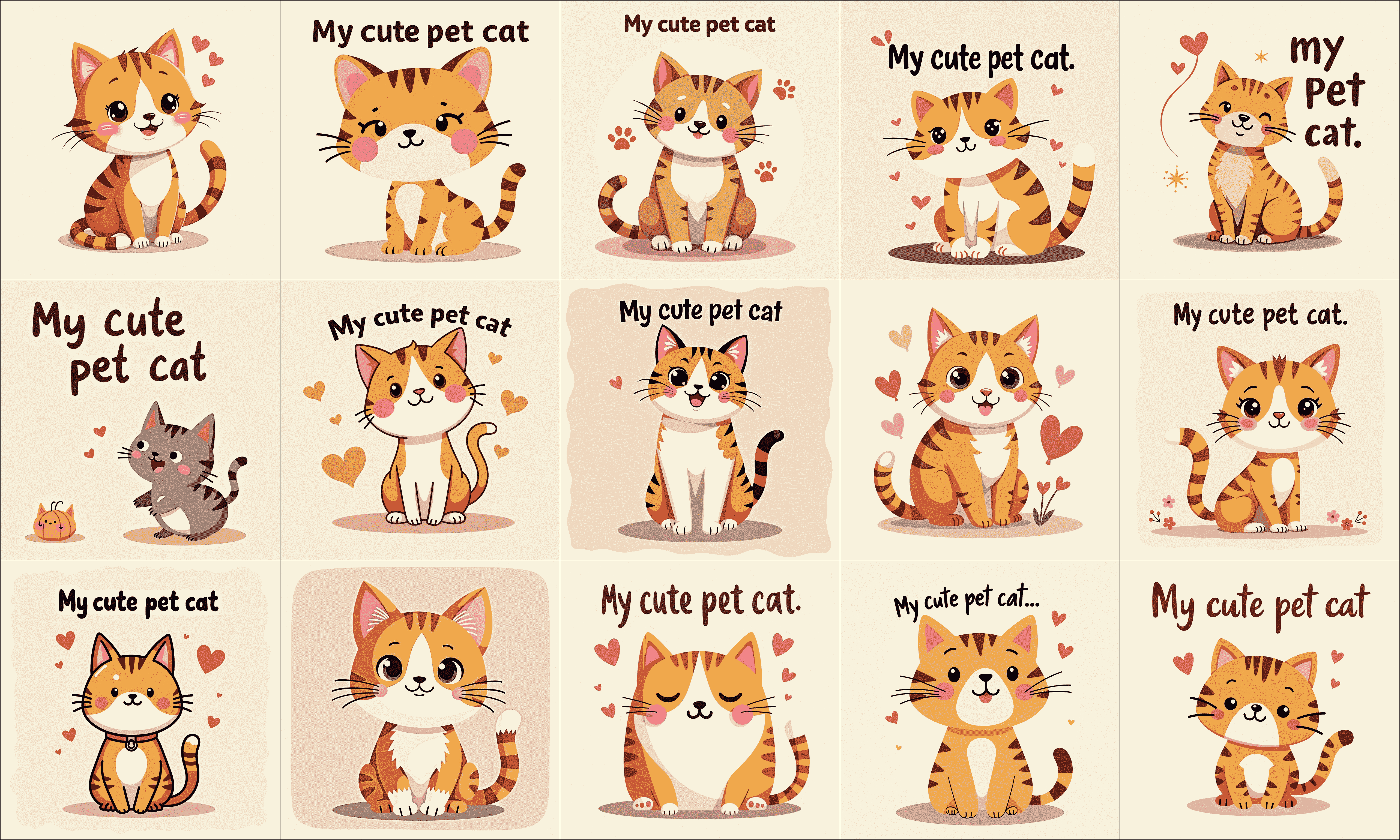}
    \end{minipage}
}%
\vspace{-0.4cm}
  \caption{Visual poster generation for the prompt "A poster with a title 'My cute pet cat'". The top half shows samples with low ASD, where the title is often missing, incomplete, or misaligned (e.g., only “My cute” or “pet cat”). The bottom half presents generations with high ASD, which consistently exhibit well-positioned and legible titles alongside semantically aligned visuals. This reinforces the effectiveness of ASD-based filtering in guiding structured text generation in stylized or design-oriented tasks.}
  \label{flux_app3}
\end{figure}

As illustrated in Figures~\ref{flux_app1} to~\ref{flux_app3}, our method improves the success rate of text rendering. Samples with higher ASD values are more likely to successfully render the intended visual text, whereas low-ASD samples often result in incomplete or entirely missing text. This indicates that the ASD metric provides a meaningful signal for selecting high-quality generations in this task.

\end{document}